\documentclass[letterpaper, 9 pt, technote]{IEEEtran}  

\IEEEoverridecommandlockouts                              

\usepackage{graphics} 
\usepackage{epsfig} 
\usepackage{mathptmx} 
\usepackage{times} 
\usepackage{amsmath} 
\usepackage{amssymb}  
\usepackage{epstopdf}
\usepackage{algpseudocode}
\usepackage{algorithm}
\usepackage[normalem]{ulem}
\usepackage{soul}

\newcommand{\spr}[1]{{\color{red} #1}}

\title{\LARGE \bf
	Automated Construction of Metric Maps using a Stochastic Robotic Swarm Leveraging Received Signal Strength
}

\author{Ragesh K. Ramachandran$^{1}$ and Spring Berman$^{2}$
	\thanks{*This work was supported by the Arizona State University Global Security Initiative.}
	\thanks{$^{1}$Ragesh K. Ramachandran is with the Department of Computer Science, University of Southern California, Los Angeles, CA 90089, USA {\tt\small  rageshku@usc.edu}}
    \thanks{$^{2}$Spring Berman is with the School for Engineering of Matter, Transport and Energy, Arizona State University, Tempe, AZ 85287, USA  {\tt\small  spring.berman@asu.edu}}%
}

\let\oldtitle\title
\renewcommand\title[1]{%
    \begingroup
        \providecommand{\ttlit}{}%
        \renewcommand{\ttlit}[1]{}%
        \providecommand{\titlenote}{}%
        \renewcommand{\titlenote}[1]{}%
        \hypersetup{pdftitle={#1}}%
        \def\thetitle{#1}%
        \pdfbookmark[0]{#1}{title}
    \endgroup
    \oldtitle{#1}%
}

\usepackage{calc}

\usepackage[shortcuts]{extdash}

\usepackage{graphicx}

\usepackage{booktabs}

\usepackage{verbatim}

\usepackage{amsmath}
\usepackage{psfrag}

\makeatletter
\def\citet{\@ifstar{\citetstar}{\citetnostar}}
\def\Citet{\@ifstar{\Citetstar}{\Citetnostar}}
\def\citetnostar{\@ifnextchar[{\squarecitet}{\simplecitet}}
\def\squarecitet[#1]{\@ifnextchar[{\twocitet[#1]}{\onecitet[#1]}}
\def\Citetnostar{\@ifnextchar[{\squareCitet}{\simpleCitet}}
\def\squareCitet[#1]{\@ifnextchar[{\twoCitet[#1]}{\oneCitet[#1]}}
\def\citetstar{\@ifnextchar[{\squarecitetstar}{\simplecitetstar}}
\def\squarecitetstar[#1]{\@ifnextchar[{\twocitetstar[#1]}{\onecitetstar[#1]}}
\def\Citetstar{\@ifnextchar[{\squareCitetstar}{\simpleCitetstar}}
\def\squareCitetstar[#1]{\@ifnextchar[{\twoCitetstar[#1]}{\oneCitetstar[#1]}}
\makeatother
\def\simplecitet#1{\citeauthor{#1}~\citep{#1}}
\def\onecitet[#1]#2{\citeauthor{#2}~\citep[#1]{#2}}
\def\twocitet[#1][#2]#3{\citeauthor{#3}~\citep[#1][#2]{#3}}
\def\simplecitetstar#1{\citeauthor*{#1}~\citep{#1}}
\def\onecitetstar[#1]#2{\citeauthor*{#2}~\citep[#1]{#2}}
\def\twocitetstar[#1][#2]#3{\citeauthor*{#3}~\citep[#1][#2]{#3}}
\def\simpleCitet#1{\Citeauthor{#1}~\citep{#1}}
\def\oneCitet[#1]#2{\Citeauthor{#2}~\citep[#1]{#2}}
\def\twoCitet[#1][#2]#3{\Citeauthor{#3}~\citep[#1][#2]{#3}}
\def\simpleCitetstar#1{\Citeauthor*{#1}~\citep{#1}}
\def\oneCitetstar[#1]#2{\Citeauthor*{#2}~\citep[#1]{#2}}
\def\twoCitetstar[#1][#2]#3{\Citeauthor*{#3}~\citep[#1][#2]{#3}}

\usepackage[cmex10]{mathtools}             
\usepackage{amsfonts,amssymb}
\usepackage{mathrsfs}                      

\usepackage{varioref}
\labelformat{subfigure}{\thefigure\textup{(#1)}}
\labelformat{equation}{\textup{(#1)}}

\usepackage[%
        pagebackref=false,
        bookmarks=true,bookmarksopen=true,
        bookmarksnumbered=true,
        breaklinks=true,
        colorlinks=true,
        anchorcolor=blue,citecolor=blue,
        urlcolor=blue,linkcolor=blue,filecolor=blue,
        menucolor=blue,
    ]{hyperref}
\usepackage{doi}

\usepackage{subfigure}

\makeatletter
\@ifpackageloaded{hyperref}{%
\@ifpackageloaded{amsmath}{%
\newcommand{\AMShreffix}[1]{%
        \expandafter\let\csname AMShreffix#1\expandafter\endcsname%
                \csname #1\endcsname%
        \expandafter\renewcommand\csname #1\endcsname{%
                \@hyper@itemfalse\csname AMShreffix#1\endcsname}}
\AtBeginDocument{%
\AMShreffix{equation}
\AMShreffix{align}
\AMShreffix{alignat}
\AMShreffix{flalign}
\AMShreffix{gather}
\AMShreffix{multline}
}}{}}{}
\makeatother

\begin{document}
	\bstctlcite{IEEEexample:BSTcontrol}

	\maketitle
	\thispagestyle{empty}
	\pagestyle{empty}

	\begin{abstract}
		
		In this work, we present a novel automated procedure for constructing a metric map of an unknown domain with obstacles using uncertain position data collected by a swarm of resource-constrained robots. The robots obtain this data during random exploration of the domain by combining onboard odometry information with noisy measurements of signals received from transmitters located outside the domain. This data is processed offline to compute a density function of the free space over a discretization of the domain. We use persistent homology techniques from topological data analysis to estimate a value for thresholding the density function, thereby segmenting the obstacle-occupied region in the unknown domain. Our approach is substantiated with theoretical results to prove its completeness and to analyze its time complexity. The effectiveness of the procedure is illustrated with numerical simulations conducted on six different domains, each with two signal transmitters.
		
	\end{abstract}

	\section{INTRODUCTION}
	\label{sec:intro}
	
	\IEEEPARstart{S}{warms} of autonomous robots can potentially be used for many applications in remote or hazardous locations, including exploration, environmental monitoring, disaster response, and search-and-rescue operations. These applications may require the swarm to generate a map of the environment without having access to GPS measurements or reliable inter-robot communication. Due to size and cost constraints, each 
    robot may be highly resource-limited and thus unable to use existing mapping techniques such as those in \cite{Robertson:2009:SLM:1620545.1620560}. 
	
	For the most part, existing works on mapping using robotic swarms have focused on generating a {\it topological map}, e.g. \cite{Caccavale2017,DirafzoonBL15}. There has been some work on multi-robot mapping \cite{Thrun2005} that requires inter-robot communication, but to the best of our knowledge, there is no other work which addresses {\it metric} mapping of unknown domains with multiple obstacles using a swarm of robots with the resource limitations that we consider in this paper. 
	A topological map is a sparse representation of an environment that encodes all of its topological features, such as holes that signify the presence of obstacles, and provides a collision-free path through the environment in the form of a roadmap. 
	A metric map, or simply {\it map}, of a domain gives metric information about the subset of a domain that is unoccupied by features such as obstacles.  This representation includes the precise locations and geometries of features in the domain. The combination of metric maps with filtering techniques such as particle filtering or Kalman filtering enables robots with more sophisticated capabilities than the ones we consider to accurately localize themselves. The construction of a topological map is less sensitive to sensor and actuator noise in the robots than the construction of a metric map, but it only yields estimated features that are homotopic to the actual features  in a domain. 
	
	
	In this paper, we develop an automated procedure for constructing a metric map of an unknown, GPS-denied environment with obstacles using uncertain localization data acquired by a swarm of robots with local sensing and no inter-robot communication. 
	The procedure is scalable with the number of robots. Each robot generates the localization data by combining its onboard odometry information with the measured strength of signals that are emitted by transmitters located outside the domain. 
	For example, in a disaster response scenario, the robots may be able to detect radio signals only from the area outside the domain from which they were deployed. Our procedure is also applicable to indoor environments; even though signal propagation through such environments has high unpredictability \cite{Ferris2007}, much research has been devoted to the use of received signal strength intensity (RSSI) for indoor localization of robots \cite{roehrig16}.
	In \cite{Prorok:2014}, a technique is presented for multi-robot localization that could be used for mapping environments without global position information. Similar to our approach, this technique uses robot measurements of external signals; however, unlike our approach, it requires robots to distinguish neighboring robots from obstacles and communicate explicitly with them. We prove that our procedure will generate a metric map under specified assumptions on the coverage of the domain by the robots.
	

	Our previous papers \cite{Ramachandran_Wilson_Berman16}  and \cite{Ramachandran_Wilson_Berman_RAL_17} presented procedures for estimating the number of obstacles in an unknown domain and extracting a topological map of the domain, respectively. The methodology presented in \cite{Ramachandran_Wilson_Berman_RAL_17} 
	generates a topological map in the form of a Voronoi diagram by applying clustering and wave propagation algorithms to a probabilistic map and does not incorporate RSSI measurements.
	We have also developed an optimal control method for mapping GPS-denied environments using a swarm of robots with both advective and diffusive motion \cite{Ramachandran_Elamvazhuthi_Berman}. Although this method only requires measurements of encounter times with obstacles, it relies on an accurate partial differential equation model of the swarm dynamics, and it is ineffective on domains with multiple obstacles.

	Our procedure is fundamentally an {\it occupancy grid mapping method}, which represents the unknown domain using a set of evenly-spaced binary random variables that each indicate the presence or absence of an obstacle at that location in the domain. 
    Such methods have been studied extensively, 
  both in single-robot \cite{Thrun:1996:IGT,Meyer-Delius:2012:OGM} and multi-robot \cite{saeedi2015occupancy,Birk1677951} settings. Our contribution over existing occupancy grid mapping strategies is a guarantee of the probabilistic completeness of our mapping procedure, given in \hyperref[the:threshold]{Theorem~\ref{the:threshold}}. 
  We prove that our approach  results in the map of the 
  unknown environment with probability one, as long as the assumptions associated with \hyperref[the:threshold]{Theorem~\ref{the:threshold}} are satisfied. 
This proof of probabilistic completeness
was absent in our earlier work \cite{Ramachandran_Wilson_Berman_RAL_17}. Our  result in \hyperref[lem: estimation error]{Lemma~\ref{lem: estimation error}}, which is needed to prove \hyperref[the:threshold]{Theorem~\ref{the:threshold}},  cannot be proved for the system considered in \cite{Ramachandran_Wilson_Berman_RAL_17} since it is unobservable. This provides insight into why our approach in \cite{Ramachandran_Wilson_Berman_RAL_17} cannot be used for metric mapping.
We note that although our strategy uses an \textit{extended Kalman filter} as in most landmark-based mapping techniques, such as  simultaneous localization and mapping (SLAM) \cite{Thrun:2005:PR:1121596}, 
  unlike such techniques, our approach does not require  sophisticated sensors or processing to characterize obstacles and distinguish them from other robots.
  
  
  
	The first step in our procedure, namely, data collection by a swarm of robots during exploration of the domain, is a decentralized process. In the subsequent step, the collected data is processed offline to compute a probability of occupancy on the grid cells. The computations from this step onward are executed by a central computer that generates the domain map from the computed density function. 
	This is the only centralized component of our mapping procedure, and it is
	scalable with the number of robots since the map computation can be parallelized. 
	Tools from topological data analysis (TDA) are used to compute a threshold density value in order to identify the obstacle-filled region in the occupancy grid. This computation is performed by constructing a probability-based filtration on the free space in the domain. We direct the reader to Section II of \cite{Ramachandran_Wilson_Berman_RAL_17} for the necessary background on the topological concepts that are used in this paper.  
	
	
	

	The reason for computing the map offline is twofold. First, the robots localize in the domain with uncertainty that increases over time due to noise in their actuators, sensors, and RSSI measurements. Even though we prove in \hyperref[lem: estimation error]{Lemma~\ref{lem: estimation error}} that this uncertainty is bounded, the bound could be large for a particular robot depending on the random path that it follows, which would make its localization data unreliable. Hence, each individual robot can only generate an uncertain map of the region that it explores. However, our approach constructs an accurate estimate of the map of a region by fusing data offline from multiple robots that explore the region.
	Second, in order for each robot to construct the map of the domain online, it should individually cover the entire domain and have sufficient computational capabilities to perform all the map generation calculations onboard. In our strategy, this is infeasible due to the low computational resources of the robots that we consider. As an alternative, the robots could construct local maps, communicate these maps to other robots that they encounter during the course of exploration, and merge the maps that they receive from the other robots. However, this would require the robots	to have communication capabilities, which we do not assume in our scenario.

	\section{PROBLEM STATEMENT}
	\label{sec:problem statement}
	
	We consider the problem of estimating the metric map of a closed, bounded, path connected, GPS-denied domain $D \subset \mathbb{R}^d$ with obstacles using uncertain localization data acquired by a swarm of $N$ robots while exploring the domain. We restrict our analysis to domains with boundaries having regularity of at least Lipschitz continuity. Although here we only consider the case $d = 2$, it is straightforward to extend our procedure to the case where $d > 2$. We exclude scenarios where an obstacle is located very close to the domain boundary, since it is highly unlikely that the robots will enter the gap between the boundary and obstacle. We assume that such gaps are at least twice a robot's sensing diameter. 
	
	Each robot is equipped with a compass, wheel encoders, and a received signal strength indicator (RSSI) device such as Atheros \cite{Xie2015}, and it can detect obstacles and other robots within its local sensing radius and perform collision avoidance maneuvers. Two radio transmitters are assumed to be located outside the domain, and the robots' RSSI devices can measure their signals anywhere inside the domain. The assumption on the transmitters' location is included to avoid complexities in the analysis required to prove \hyperref[the:threshold]{Theorem~\ref{the:threshold}} that arise from singularities in the signal strength attenuation model, i.e., points where this model is undefined. However, since these points are isolated, and thus comprise a set of measure zero, in practice our approach should work even if the transmitters are located inside the domain. As the proof of  \hyperref[the:threshold]{Theorem~\ref{the:threshold}} shows, our strategy requires at least two transmitters to map a two-dimensional domain. We assume that the robots have sufficient memory to store the data that they collect during exploration.
	We also assume that after a sufficiently large time $T$, the robots have covered the domain according to the coverage definition given in \autoref{subsec:Completeness of the Approach}.  After time $T$, the robots move to a common location for extraction of the stored data.
	
	
	
	The robots move with a constant speed  $v$ and a heading $\theta(t)$ at time $t$ with respect to a fixed global frame. 
	The position and velocity state vectors of a robot in this frame are defined as $\mathbf{X}(t) =  [x(t),\ y(t)]^T$ and $\mathbf{V}(t) =  [v_x(t),\ v_y(t)]^T = [v\ \cos(\theta(t)), \ v\ \sin(\theta(t))]^T$, respectively. At the initial time $t=0$, the start of the exploration phase, a precise estimate of $\mathbf{X}(0)$ and $\theta(0)$ is provided to each robot. During the deployment, each robot generates a uniform random number $U \in [0, 1]$ at the start of every time step $\Delta t$. If $U \leq p_{th}$, where $p_{th}$ is a specified value, then the robot randomly chooses a new heading $\theta(t) \in [-\pi,\pi]$. We define $\mathbf{W}_x(t) \in \mathbb{R}^2$ and $\mathbf{W}_v(t) \in \mathbb{R}^2$ as vectors of independent, zero-mean normal random variables that are generated at time $t$. These vectors model randomness in the robots' motion due to wheel actuation noise. We define the vector $\mathbf{W}(t) \in \mathbb{R}^{4}$ as $\mathbf{W}(t) = [\mathbf{W}_x(t) ~ \mathbf{W}_v(t)]$  and note that $\mathbf{W}(t) \sim \mathcal{N}(\mathbf{0}, \mathbf{Q}),\ \mathbf{Q} \in \mathbb{R}^{4 \times 4}$. 
	
	Using this notation, we model each robot as a point mass that follows the standard linear odometry motion model \cite{Thrun:2005:PR:1121596}, whose state space form can be written as:
	\begin{align}
	\label{eqn:State_space_motion_model}
	\begin{bmatrix}
	\mathbf{X}(t + \Delta t)
	\\ 
	\mathbf{V}(t + \Delta t)
	\end{bmatrix}
	= 
	\begin{bmatrix}
	\mathbf{I} & \Delta t \mathbf{I}\\  
	\mathbf{0}& \mathbf{I}
	\end{bmatrix}
	\begin{bmatrix}
	\mathbf{X}(t)
	\\ 
	\mathbf{V}(t)
	\end{bmatrix}
	+
	\begin{bmatrix}
	\mathbf{W}_x(t)
	\\ 
	\mathbf{W}_v(t)
	\end{bmatrix},
	\end{align}
	where $\mathbf{I}$ is the identity matrix.
	We denote the system matrix of \autoref{eqn:State_space_motion_model} by $\mathbf{A}$.
	
	While performing this correlated random walk through the domain, each robot uses an extended Kalman filter \cite{Thrun:2005:PR:1121596} to estimate its global position and the associated covariance matrix from its onboard odometry and RSSI measurements of the signals emitted by the two transmitters. The robot records this estimated position and covariance matrix at fixed time intervals. 
	Although exploration through random walking gives only weak guarantees on complete coverage of the domain, it is a simple motion strategy that can be implemented on robots with the limitations that we consider. It should be noted that any exploration strategy that accommodates these limitations can be substituted for random walking. We specify that the line joining the two transmitters lies outside the domain (see  \hyperref[the:threshold]{Theorem~\ref{the:threshold}}).  The signal strength attenuation of a radio signal from a transmitter $i$ is a function of distance from the transmitter location $\mathbf{X}_i$ \cite{Seidel1992}. We adopt the model $S_i(\mathbf{X}(t)) = K_i Pow_i ||\mathbf{X}(t) - \mathbf{X}_i||^{-\alpha}$ presented in \cite{Nguyen2005}, 
	where $\alpha \in [0.1, 2]$, $Pow_i$ is the transmitted signal voltage of transmitter $i$, and $K_i$ is the corresponding proportionality constant. We set $\alpha=2$, as is commonly done in the literature \cite{Seidel1992}. We define $\mathbf{S}(\mathbf{X}(t)) = [S_1(\mathbf{X}(t)), ~...,~ S_l(\mathbf{X}(t))]^T$, where $l$ is the number of transmitters (here, we set $l=2$).
	We also define $\mathbf{N}_S(t) \in \mathbb{R}^l$ and $\mathbf{N}_V(t) \in \mathbb{R}^2$ as vectors of independent, zero-mean normal random variables that are generated at time $t$. These vectors model noise in the robots' RSSI devices and wheel encoders, respectively. 
	Let $\mathbf{Z}(t)$ denote the vector of sensor measurements received by a robot at time $t$. Then the output equation of the system can be written as, 
	\begin{align}
	\label{eqn:output_eqn}
	\mathbf{Z}(t) = \begin{bmatrix}
	\mathbf{S}(\mathbf{X}(t))\\ 
	\mathbf{V}(t)
	\end{bmatrix}
	+
	\begin{bmatrix}
	\mathbf{N}_S(t)\\ 
	\mathbf{N}_V(t)
	\end{bmatrix}.
	\end{align}


	\section{MAP GENERATION PROCEDURE}
	\label{sec: procedure}
	This section describes a procedure for extracting a metric map of the domain as an occupancy grid map using the noisy localization data collected by the swarm of robots. 
	
	\subsection{Computation of the Density Function of Free Space on a Discretization of the Domain}
	\label{subsec:Computation of Density Function}

	As in other occupancy grid mapping algorithms \cite{Thrun:2005:PR:1121596}, our first step is to discretize the domain into a fine grid of $M$ cells. The objective of this step is to use the robots' recorded data on their estimated positions to compute a density function $p^{f}: m_i \rightarrow [0, 1]$ that encodes the probability of a cell $m_i,\ i \in {1, ..., M}$ being unoccupied by an obstacle, or {\it free}.  We use the notation $p^{f}_i$ instead of $p^{f}(m_i)$ for brevity. Here we summarize our approach to computing $p^{f}_i$. Although it is similar to the approaches in our earlier work \cite{Ramachandran_Wilson_Berman_RAL_17,Ramachandran_Wilson_Berman16}, the probabilistic occupancy grid map computation in this paper uses a different equation for $s_i$ (\autoref{eq:scorecomp}), the score assigned to each grid cell $i$, than the computation in our previous works. 

	While each robot $j \in \{1, ..., N\}$ moves randomly through the unknown environment, it records data at times $t_k \in [0, T]$, $k \in {1, ..., K}$. This data consists of the tuple $d^{j}_{k} = \{\mu^{j}_{k}, \sigma^{j}_{k}\}$, where $\mu^{j}_{k} \in \mathbb{R}^{2}$ and $\sigma^{j}_{k} \in \mathbb{R}^{2 \times 2}$ are the mean and covariance matrix, respectively, of the robot's estimate of its position in Cartesian coordinates at time $t_k$. We define $p_{ijk}$ as the discrete probability that the $j^{th}$ robot occupied the cell $m_i$ at time $t_k$.
	This probability is calculated for all robots, cells, and times $t_k$ by integrating the Gaussian distribution with mean $\mu^{j}_{k}$ and covariance matrix $\sigma^{j}_{k}$ over the part of the domain occupied by cell $m_i$.
	We then filter out probabilities $p_{ijk}$ that are obtained from Gaussian distributions which are centered far from each grid cell $m_i$.
	Toward this end, we define the set $P_i = \{p_{ijk} ~|~ p_{ijk} > \rho\}$, where $\rho > 0$ is a tolerance. 
	In this paper, we set $\rho=0.05$. 
	We compute $p_{i}^{f}$ for each cell $m_i$ using a technique similar to the $\log$ odds computation that is commonly employed in the robot SLAM literature \cite{Thrun:2005:PR:1121596}. 
	A score $s_i \in [0, \infty)$ is assigned to each grid cell $m_i$ according to the equation
	\begin{align}
	s_i = \frac{1}{\left | P_i \right |}\sum _{p_{ijk} \in P_i} \log\left ( \frac{1}{1 - p_{ijk}} \right ). \label{eq:scorecomp}
	\end{align}
	We then compute the probability that cell $m_i$ is free using the formula $p_{i}^{f} ~=~ 1 - (\exp(s_i))^{-1}$. 
	
	
	
	
	
	Next, we apply a moving average linear filter, a common technique in image processing, to the probabilities $p_{i}^{f}$. This ensures that the automated thresholding step, described in the next section, is effective even if the robots fail to cover a few free grid cells in the domain. For each grid cell $m_i$, we replace $p_{i}^f$ with the mean of $p_i^f$ and the $p_j^f$ of its neighboring grid cells $m_j$.  This eliminates any $p_{i}^{f}$ value that is unrepresentative of its neighborhood. The simulations in \autoref{sec:simulation results} use a $3 \times 3$ square neighborhood for  filtering. Strictly speaking, this step can be skipped if the assumption on the coverage of the domain by the swarm is satisfied.

	\subsection{Thresholding the Density Function to Generate the Map}
	\label{subsec:Thresholding the Density Function to Generate the Map}
	In this step, we threshold each $p_{i}^{f}$ to classify the corresponding grid cell $m_i$ as a free or obstacle-occupied cell. The existence of a threshold for this classification is proven in \hyperref[the:threshold]{Theorem~\ref{the:threshold}}. 
	We apply persistent homology \cite{edelsbrunner2008persistent}, a topological data analysis (TDA) technique based on algebraic topology \cite{Hatcher2002},
	to automatically find a threshold based on the $p_{i}^{f}$ of each grid cell. An implicit assumption required for this technique is that each obstacle contains at least one grid cell with  $p^f_i = 0$. This TDA-based technique provides an adaptive method for thresholding an occupancy grid map of a domain that contains obstacles at various length scales. In fact, it can be used with other occupancy grid mapping methods to implement automated thresholding. We describe this technique in full in \cite{Ramachandran_Wilson_Berman_RAL_17}. In this paper, the persistent homology computation 
	was done using the MATLAB-based JavaPlex package \cite{Adams2014}.

	\section{ANALYSIS OF THE MAPPING PROCEDURE}
	\label{sec:analysis}
	
	\subsection{Probabilistic Completeness of the Procedure}
	\label{subsec:Completeness of the Approach}
	
	In this section, we analyze the {\it completeness of the approach} in a  probabilistic sense, meaning that the procedure described in \autoref{sec: procedure} will result in a probabilistic occupancy map of the unknown domain that distinguishes between occupied and free grids cells, provided that the inputs to the procedure satisfy certain assumptions with probability one. The approach may fail to produce the desired output if the assumptions do not hold.  The simulation results in \autoref{sec:simulation results} demonstrate the effectiveness of our procedure even when the required assumptions are not fully satisfied.
	
	The most important assumption required for the completeness of our approach is that the domain is {\it completely covered} by the swarm of robots. By this, we mean that 
the recorded localization data includes at least one data tuple per free grid cell whose $\mu$ lies inside the grid cell. We assume that even if some of the robots fail to return after exploring the domain, sufficient data is obtained from the recovered robots to achieve complete coverage of the domain.
	
	
	
	We begin our analysis by proving the existence of a threshold on the density function, which serves as a decision variable to distinguish between free and occupied grid cells. Toward this end, we first state the following lemma, which gives a result that is required to prove \hyperref[the:threshold]{Theorem~\ref{the:threshold}}. The result in \hyperref[lem: estimation error]{Lemma~\ref{lem: estimation error}} follows trivially when both the robot dynamics and measurement models are linear. However, proving this result requires a careful analysis when either the dynamics or measurement models are nonlinear, as in our case. 
	
	\vspace{2mm}
	\newtheorem{lem}{Lemma}
	
	\begin{lem}
		\label{lem: estimation error}
		The error in the robots' position estimates is bounded with probability one, with a common bound for all robots, if each robot follows the motion model \autoref{eqn:State_space_motion_model} and estimates its state vector using an extended Kalman filter based on the outputs in \autoref{eqn:output_eqn}.
	\end{lem}
	
	\vspace{2mm}
	\begin{IEEEproof}
		Let $\mathbf{S}(\mathbf{X})_{\mathbf{X}}$ denote the first-order derivative of $\mathbf{S}(\mathbf{X})$ with respect to $\mathbf{X}$ in \autoref{eqn:output_eqn}. Note that we have dropped the variable $t$ for conciseness. 
		We also define $ \mathbf{\hat{X}}$ as the estimate of $\mathbf{X}$. 
        Assuming that $\mathbf{S}(\mathbf{X})$ is at least twice differentiable in a neighborhood of the estimate $\mathbf{\hat{X}}$, we can write the first-order Taylor series expansion of $\mathbf{S}(\mathbf{X})$ about $\mathbf{\hat{X}}$ as,
		\begin{align}
		\label{eqn:s_x taylor expan}
		\mathbf{S}(\mathbf{X}) = \mathbf{S}(\mathbf{\hat{X}}) + \mathbf{S}(\mathbf{X})_{\mathbf{\hat{X}}}(\mathbf{X} - \mathbf{\hat{X}}) + \mathcal{O}(||\mathbf{X} - \mathbf{\hat{X}}||^2 ).
		\end{align}
	Defining $\mathbf{h} = (\mathbf{X} - \mathbf{\hat{X}})$, the higher-order terms in \autoref{eqn:s_x taylor expan} are represented by $\mathcal{O}(||\mathbf{h}||^2)$. By 
        definition, 
        $\mathcal{O}(||\mathbf{h}||^2)$ is bounded above by $K_\mathbf{S}\left \| \mathbf{h} \right \|^2$  as $\left \| \mathbf{h} \right \| \rightarrow 0$, where $K_\mathbf{S}$ is some positive constant. This implies that there exists an open ball of radius $\epsilon > 0$ around $\mathbf{0}$ such that if $\left \| \mathbf{h} \right \| < \epsilon$, 
        then $ \left \|\mathcal{O}( \left \| \mathbf{h} \right \|^2 )\right \| \leq K_\mathbf{S} \left \|  \mathbf{h} \right \|^2  $.  Thus, the inequality
		\begin{align}
		\label{eqn:s_x bound}
		\left \| \mathbf{S}(\mathbf{X}) - \mathbf{S}(\mathbf{\hat{X}}) - \mathbf{S}(\mathbf{X})_{\mathbf{\hat{X}}}(\mathbf{X} - \mathbf{\hat{X}})\right \| \leq K_\mathbf{S} \left \| \mathbf{X} - \mathbf{\hat{X}} \right \|^2
		\end{align}
		is satisfied in some neighborhood of $\mathbf{\hat{X}}$ if $\mathbf{S}(\mathbf{\hat{X}})$ is at least twice differentiable in that neighborhood.
		
		From Theorem 3.1 in \cite{Reif1999}, we know that the estimation error $\zeta_k = \left \| \mathbf{X}_k - \mathbf{\hat{X}}_k \right \| $ of an extended Kalman filter at the $k^{th}$ time step, where $k \in \{1,2,...,K\}$, is bounded with probability one as long as the following conditions hold:
		
		\vspace{2mm}
		
		\begin{enumerate}
			\item $\zeta_0 \leq \epsilon $ for some $\epsilon \geq 0$.
			\item Define $\mathbf{X}_k^s = [\mathbf{X}_k; \mathbf{V}_k]$ as the state vector in \autoref{eqn:State_space_motion_model}, $f(\mathbf{X}_k^s)$ as the state map, and $\mathbf{A}_k = \frac{\partial f}{\partial \mathbf{X}_k^s}(\mathbf{\hat{X}}_k^s)$. The matrix  $\mathbf{A}_k$ is nonsingular for all $k \geq 0$.  
			\item Let $h(\mathbf{X}_k^s)$ be the output map given in \autoref{eqn:output_eqn} and $\mathbf{H}_k = \frac{\partial h}{\partial \mathbf{X}_k^s}(\mathbf{\hat{X}}_k^s)$. Define the functions $\phi$ and $\chi$ as:
			\begin{eqnarray}
			\hspace{-4mm} \phi (\mathbf{X}_k^s,\mathbf{\hat{X}}_k^s) \hspace{-2mm} &=&  \hspace{-2mm} f(\mathbf{X}_k^s) - f(\mathbf{\hat{X}}_k^s) - \mathbf{A}_k(\mathbf{\hat{X}}_k^s)\left(\mathbf{X}_k^s - \mathbf{\hat{X}}_k^s \right), \label{eqn:phi} \\
			\hspace{-4mm} \chi (\mathbf{X}_k^s,\mathbf{\hat{X}}_k^s)  \hspace{-2mm} &=&  \hspace{-2mm} h(\mathbf{X}_k^s) - h(\mathbf{\hat{X}}_k^s) - \mathbf{H}_k(\mathbf{\hat{X}}_k^s)\left(\mathbf{X}_k^s - \mathbf{\hat{X}}_k^s \right). \label{eqn:chi}
			\end{eqnarray} 
			There exist real numbers $\epsilon_\phi, \epsilon_\chi, K_\phi, K_\chi > 0$ such that
			\begin{eqnarray}
			\hspace{-4mm} \left \| \phi (\mathbf{X}_k^s,\mathbf{\hat{X}}_k^s) \right \| &\leq& K_\phi \left \|\mathbf{X}_k^s - \mathbf{\hat{X}}_k^s \right \|^2 ~~\leq~~ K_\phi  \epsilon_\phi^2, \label{eqn:state error bound condition} \\
			\hspace{-4mm} \left \| \chi (\mathbf{X}_k^s,\mathbf{\hat{X}}_k^s) \right \| &\leq& K_\chi \left \|\mathbf{X}_k^s - \mathbf{\hat{X}}_k^s \right \|^2 ~~\leq~~ K_\chi \epsilon_\chi^2. \label{eqn:output error bound condition}
			\end{eqnarray}
			\item There are positive real numbers $\bar{a}, \bar{h}, \bar{p}, \underline{p} > 0$ such that 
			\begin{eqnarray}
			\left \| \mathbf{A}_k \right \| &\leq& \bar{a} \label{eqn:bound on A_k condition}, \\
			\left \| \mathbf{H}_k  \right \|&\leq& \bar{h} \label{eqn:bound on H_k condition}, \\
			\underline{p}I &\leq& \mathbf{P}_k ~\leq~ \bar{p}\mathbf{I}, \label{eqn:bound on P_k condition}
			\end{eqnarray}
			where $\mathbf{P}_k$ is the covariance matrix at the $k^{th}$ time step.
		\end{enumerate}
		
		\vspace{2mm}
		
		To prove the lemma, we will now show that these four conditions are satisfied. Conditions (1) and (2) are satisfied because $\zeta_0 = 0$ and $\mathbf{A}_k$ is the constant matrix $\mathbf{A}$, which is nonsingular. \autoref{eqn:state error bound condition} is satisfied trivially, since $\phi (\mathbf{X}_k^s,\mathbf{\hat{X}}_k^s)$ is zero when $\mathbf{A}_k$ is a constant matrix. In condition (3), we need to determine whether the bounds described in \autoref{eqn:output error bound condition} are fulfilled with $h(\mathbf{X}_k^s) = [\mathbf{S}(\mathbf{X}(t)); \mathbf{V}(t)] $  in \autoref{eqn:chi}. To verify this, it is enough to show that a condition analogous to \autoref{eqn:chi} is satisfied when the output map is restricted to the signal map. In other words, we need to check whether \autoref{eqn:chi} is satisfied when $ h(\mathbf{X}_k^s) = \mathbf{S}(\mathbf{X})$.  \autoref{eqn:s_x bound} shows that this condition is true locally at every point as long as $\mathbf{S}(\mathbf{X})$ is at least twice differentiable, which is true for all points inside the domain in our case, since the transmitters are located outside the domain.
		
		Examining condition (4), we find that computation of $\mathbf{H}_k$ is required for further analysis. Given the definition $ h(\mathbf{X}_k^s) = [\mathbf{S}(\mathbf{X}(t)); \mathbf{V}(t)]$ from \autoref{eqn:output_eqn}, we can compute the Jacobian of $ h(\mathbf{X}_k^s)$ as, 
		\begin{align}
		\label{eqn:H_k}
		\mathbf{H}_k(\mathbf{X}_k^s) 
		= 
		\begin{bmatrix}
		\mathbf{S}_k(\mathbf{X_k})_{\mathbf{X}}& \mathbf{0}\\ 
		\mathbf{0}& \mathbf{I}
		\end{bmatrix},	
		\end{align}
		where
		\begin{align}
		\label{eqn:S(X) is derivative}
		\mathbf{S}_k(\mathbf{X_k})_{\mathbf{X}} 
		= 
		\begin{bmatrix}
		\frac{-\alpha K_1 Pow_1(x_k - x_{t1})}{\left ( (x_k -x_{t1})^2 + (y_k -y_{t1})^2\right )^{\frac{2+\alpha}{2}}} & \frac{-\alpha K_1 Pow_1(y_k - y_{t1})}{\left ( (x_k -x_{t1})^2 + (y_k -y_{t1})^2\right )^{\frac{2+\alpha}{2}}} \\ 
		\frac{-\alpha K_2 Pow_2(x_k - x_{t2})}{\left ( (x_k -x_{t1})^2 + (y_k -y_{t1})^2\right )^{\frac{2+\alpha}{2}}}&  \frac{-\alpha K_2 Pow_2(y_k - y_{t2})}{\left ( (x_k -x_{t1})^2 + (y_k -y_{t1})^2\right )^{\frac{2+\alpha}{2}}}
		\end{bmatrix}.
		\end{align}
		Here, $(x_{ti}, y_{ti})$ are the Cartesian position coordinates of the $i^{th}$ transmitter. 
		
		\autoref{eqn:bound on A_k condition} and \autoref{eqn:bound on H_k condition} are trivially satisfied. Now it is left to prove that the constraint described using \autoref{eqn:bound on P_k condition} is also in agreement. This inequality is related to the observability of the system. Using Theorem 4.1 in \cite{Reif1999}, we deduce that \autoref{eqn:bound on P_k condition} is satisfied if the linearized system is observable for every $n$; i.e., if the observability matrix of the linearized system $\mathbf{O}_k = [\mathbf{H}_k; ~\mathbf{H}_k\mathbf{A}_k; ~\mathbf{H}_k\mathbf{A}_k^2; ~\mathbf{H}_k\mathbf{A}_k^3]$ has full rank for all $k$. 
		After row transformations of $\mathbf{O}_k$ using Gaussian elimination, we obtain
		\begin{align}
		\label{eqn:reduce obs matrix}
		\mathbf{O}_k 
		= 
		\begin{bmatrix}
		\mathbf{S}_k(\mathbf{X_k})_{\mathbf{X}}& &\mathbf{0}\\ 
		\mathbf{0}& &\mathbf{I}\\ 
		\mathbf{0}& &\Delta t\mathbf{S}_k(\mathbf{X_k})_{\mathbf{X}}\\ 
		& \text{\huge0} &
		\end{bmatrix},
		\end{align}
		where the large $\mathbf{0}$ is a  matrix of zeros.  Since $\Delta t \neq 0$, it is evident from \autoref{eqn:reduce obs matrix} that $\mathbf{O}_k$ is not full rank if and only if $\mathbf{S}_k(\mathbf{X_k})_{\mathbf{X}}$ is not full rank. It can be shown that the points $(x_k, y_k)$ at which $\mathbf{S}_k(\mathbf{X_k})_{\mathbf{X}}$ is not full rank obey the following equation:
		\begin{equation}
		\frac{(x_k - x_{t1})}{(y_k - y_{t1})} = \frac{(x_k - x_{t2})}{(y_k - y_{t2})} = constant. \label{eqn: unobserv cond}
		\end{equation}

        Under these constraints, condition (4) is satisfied, implying that the state estimation error for each robot is bounded. Therefore, the estimation error of the robot positions is also bounded, since it is a part of the state. The maximum of all the robots' position estimation errors serves as a bound with probability one for these errors. 
	\end{IEEEproof}
    \vspace{2mm}
\textit{Remark:} The points $(x_k,y_k)$ that satisfy \autoref{eqn: unobserv cond} comprise the line joining the two transmitters. Therefore, if we ensure that this line does not pass through the domain, then the system is observable. Alternatively, we could  make the system observable by introducing a third transmitter which is non-collinear to the other two transmitters. 
	
	 \vspace{2mm}
	
	Finally, we prove the existence of a threshold value of $p_{i}^{f}$ that distinguishes the free grid cells from the occupied cells. 
	
	\vspace{2mm}
	
	\newtheorem{theorem}{Theorem}
	
	\begin{theorem}
		\label{the:threshold}
		Under the assumption of complete coverage of the domain, a grid cell $m_i$ is free 
		if and only if there exists a threshold $\gamma \in [0, 1]$ for which 
		$p_{i}^{f} > \gamma$ with probability one.
	\end{theorem}
	
	\vspace{2mm}
	
	\begin{IEEEproof}
		We begin by proving the sufficient part of the statement; i.e., that there exists a threshold $\gamma \in [0, 1]$  such that a grid cell $m_i$ is free  if $p_{i}^{f} > \gamma$.
		\hyperref[lem: estimation error]{Lemma~\ref{lem: estimation error}} shows that there exists an estimation error bound with probability one  on the position estimate for all robots.  Thus, the uncertainty associated with position is also bounded. Also, for every free grid cell $m_i$, there is at least one data tuple $d^{j}_{k}$ 
		that is centered inside the grid due to the assumption on coverage. The boundedness of the uncertainty ensures the existence of a two-dimensional symmetric Gaussian distribution function with an associated covariance matrix having a finite norm $\sigma_{max}$. The integral of this function over the grid cell is less than or equal to the integral of the Gaussian function associated with $d^{j}_{k}$. 
		Without loss of generality, we assume for simplicity that the grids cells are squares with area $[-s,s] \times [-s,s]$ that is very small compared to the domain size. After some algebraic manipulation, we can derive that
		\begin{align}
		\label{eqn: lower bound of p_{ijk}}
		\frac{1}{\left | P_i \right |}\sum _{p_{ijk} \in P_i} \log\left( \frac{1}{1 - p_{ijk}} \right) > \frac{1}{\left | P_i \right |} \log\left ( \frac{1}{1 - (p)_i} \right )	
		\end{align}
		where,
		\begin{align}
		\label{eqn:(p)_i definition}
		(p)_i = \frac{1}{2\pi\sigma_{max}^2} \int_{-s}^{s} \int_{-s}^{s} \exp \left(-\frac{1}{2} \left(\frac{x^2}{\sigma_{max}^2}+\frac{y^2}{\sigma_{max}^2} \right) \right) dx dy.
		\end{align}


		
Using the change of variables $x = y = t \sqrt{2}\sigma_{max} $, the above double integral can be simplified to:
		\begin{align}
		\label{eqn:(p)_i definition_simplified}
		(p)_i =  \left ( \frac{1}{\sqrt{\pi}} \int_{-\frac{s}{\sqrt{2}\sigma_{max}}}^{\frac{s}{\sqrt{2}\sigma_{max}}}
		\exp(-t^2)dt  \right )^2.
		\end{align}
		
	From \cite{Abramowitz1974}, $erf(s) \equiv \frac{1}{\sqrt{\pi}} \int_{-s}^{s} \exp(-t^2) dt$ is the {\it error function}. 
This function can also be defined as $erf(s) \equiv 1-erfc(s)$, where $erfc(s) = \frac{2}{\sqrt{\pi}} \int_{s}^{\infty} \exp(-t^2)dt$ is the {\it complementary error function}.
		Therefore, \autoref{eqn:(p)_i definition_simplified} can be expressed as: 
		\begin{align}
		\label{eqn:(p)_i definition_simplified_error}
		(p)_i =  \left ( erf \left(\frac{s}{\sqrt{2}\sigma_{max}} \right) \right )^2.
		\end{align}
		
	 Corollary 1 in \cite{Chang2011} gives an upper bound on $erfc(s)$, which we can use 
to compute the following lower bound on $erf(s)$: 
		\begin{align}
        \label{eqn:erfs_ineq}
		erf(s) \geq 1-\exp(-s^2).
		\end{align}
		
From \autoref{eqn:(p)_i definition_simplified_error} and \autoref{eqn:erfs_ineq}, we obtain a lower bound on $(p)_i$: 
		\begin{align}
		\label{eqn:(p)_i lower bound}
		(p)_i \geq \left ( 1 - \exp\left ( -\frac{s^2}{2\sigma_{max}^2} \right ) \right )^2.
		\end{align}
        
		We now combine \autoref{eqn:(p)_i lower bound} and \autoref{eqn: lower bound of p_{ijk}} and use the formula $p_{i}^{f} ~=~ 1 - (\exp(s_i))^{-1}$ to compute $p_{i}^{f}$, as mentioned in \autoref{subsec:Computation of Density Function}. After some algebraic simplification, we obtain the following inequality:
		\begin{align}
		\label{eqn: lower bound on p^f_i}
		p_{i}^{f} > 1 - \left ( 1 - \left ( 1 - \exp\left ( -\frac{s^2}{2\sigma_{max}^2} \right ) \right )^2 \right )^{\left ( \frac{1}{|P_i|} \right )}.
		\end{align}
		Note that the threshold, given by the right side of the above inequality, is bounded between zero and one, and that it increases as $\sigma_{max}$ decreases and vice versa.
		
		Now we prove the condition of necessity of the statement; i.e., that there exists a threshold $\gamma$ such that $p_{i}^{f} > \gamma$ implies that the grid cell $m_i$ is free, for all grid cells. We use proof by contradiction to establish this. First, suppose that this proposition is false. Then for any chosen $\gamma$,  $p_{i}^{f} > \gamma$ does not imply that the grid cell $m_i$ is free, for all grid cells. In other words, for every $\gamma$ chosen, there exists at least one occupied grid cell  for which $p_{i}^{f} > \gamma$. Let us choose $\gamma$ to be the right-hand side of \hyperref[eqn: lower bound on p^f_i]{Inequality~\ref{eqn: lower bound on p^f_i}}. Now assume that an occupied grid cell $ Obsm_i$ satisfies the condition  $p^{f}(Obsm_i) > \gamma$. This can happen in two possible cases. First, there may exist at least one data point $d^{j}_{k}$ 
		which is centered inside $ Obsm_i$. This cannot occur, since we assume that the robot cannot move over obstacles. Second,  $\sigma_{max}$ may be unbounded, which is also not true according to \hyperref[lem: estimation error]{Lemma~\ref{lem: estimation error}}. Therefore, we have confirmed that there exists a threshold that filters occupied grid cells. In other words, we were wrong to assume that the proposition was false. Thus, the proposition is true.
	\end{IEEEproof}
	
	\vspace{2mm}
    
\textit{Remark:} \hyperref[the:threshold]{Theorem~\ref{the:threshold}} proves that our strategy will never assign the same probability $p_i^f$ to both a free grid cell and an occupied grid cell. This guarantees that the free grid cells and occupied grid cells can be distinguished  by a threshold value of $p_i^f$. In fact, multiple such thresholds exist, and thus it is useful to compute a threshold which is optimal in some sense. Toward this end, the final step of our mapping procedure described in \autoref{subsec:Thresholding the Density Function to Generate the Map} computes the minimum threshold above which the number of computed topological features of the domain (specifically, the number of connected components and holes in the thresholded map) remains constant.


	
    	\vspace{1mm}

	\textit{Remark:} \hyperref[lem: estimation error]{Lemma~\ref{lem: estimation error}} cannot be proven in the strategy presented in our previous paper \cite{Ramachandran_Wilson_Berman_RAL_17}, since the system there is unobservable (for this case, $\mathbf{S}_k(\mathbf{X_k})_{\mathbf{X}} = 0$ in \autoref{eqn:reduce obs matrix}). Thus, the strategy presented in that paper does not guarantee metric map generation of the domain. Instead, it generates only a conservative topological map \cite{Choset01topologicalsimultaneous}. In addition, note that the lower bound on $p_i^f$ from \hyperref[eqn: lower bound on p^f_i]{Inequality~\ref{eqn: lower bound on p^f_i}} increases as $|P_i|$ increases, indicating that as more robots visit a grid cell $i$, its probability of being free increases.
	
        	\vspace{2mm}

	A TDA-based technique is used to estimate the threshold $\gamma$, because the threshold computed using \hyperref[eqn: lower bound on p^f_i]{Inequality~\ref{eqn: lower bound on p^f_i}} works only with complete coverage. If $\gamma_{true}$ and $\gamma_{est}$ are the true threshold and estimated threshold computed using the method described in \autoref{subsec:Thresholding the Density Function to Generate the Map}, respectively, then $\gamma_{est} \geq \gamma_{true}$, since the metric and topological information coincide once the filtration parameter exceeds the value $1 - \gamma_{true}$.
	
	\subsection{Computational Complexity Analysis}
	\label{subsec:Computational Complexity Analysis}
	
	We analyze the computational complexity of the procedure using a similar approach to that in our paper \cite{Ramachandran_Wilson_Berman_RAL_17}. Based on our analysis, the worst-case complexity of our procedure is $ \mathcal{O}(M^{2.372})$. The first step of our procedure is the computation of the density function. This computation varies linearly with the amount of data and the number of grid cells. That is, if $N$ robots each collect $K$ elements of data while exploring a discretized domain containing $M$ grid cells, then the cost of computing the density function is of the order $\mathcal{O}(NKM)$. This step can be parallelized by processing data from each robot in parallel, resulting in a reduced computational cost of the order $\mathcal{O}(KM)$. The thresholding step is the most computationally expensive part of the procedure. This is because it requires the generation of a simplicial complex, whose size is linear in $M$, and a persistent homology  computation that has a worst-case complexity of $\mathcal{O}(M^{2.372})$, although for most practical scenarios it approaches $\mathcal{O}(M)$ \cite{Edelsbrunner_persistenthomology:}.
	
	\begin{figure*}[th]
		
		\begin{tabular}{ccccc}
			\centering	
			\hspace{-3mm}		
			
			\subfigure[PAO=0\% \label{fig:World no obstacle}]{\includegraphics[width=.18\linewidth, height=.16\linewidth]{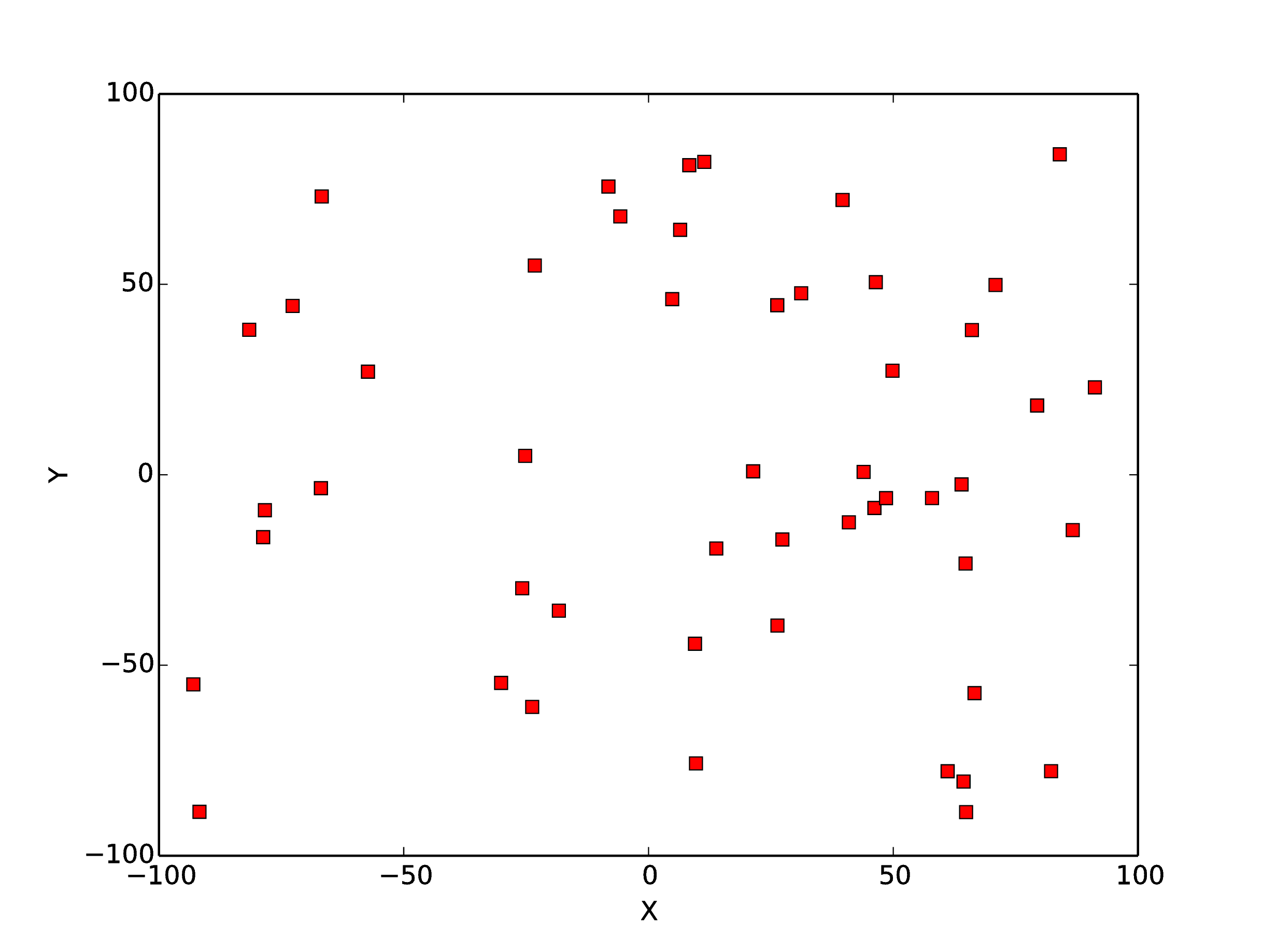}} 
			&
			\subfigure[PAO=6\% \label{fig:World one obstacle}]{\includegraphics[width=.18\linewidth, height=.16\linewidth]{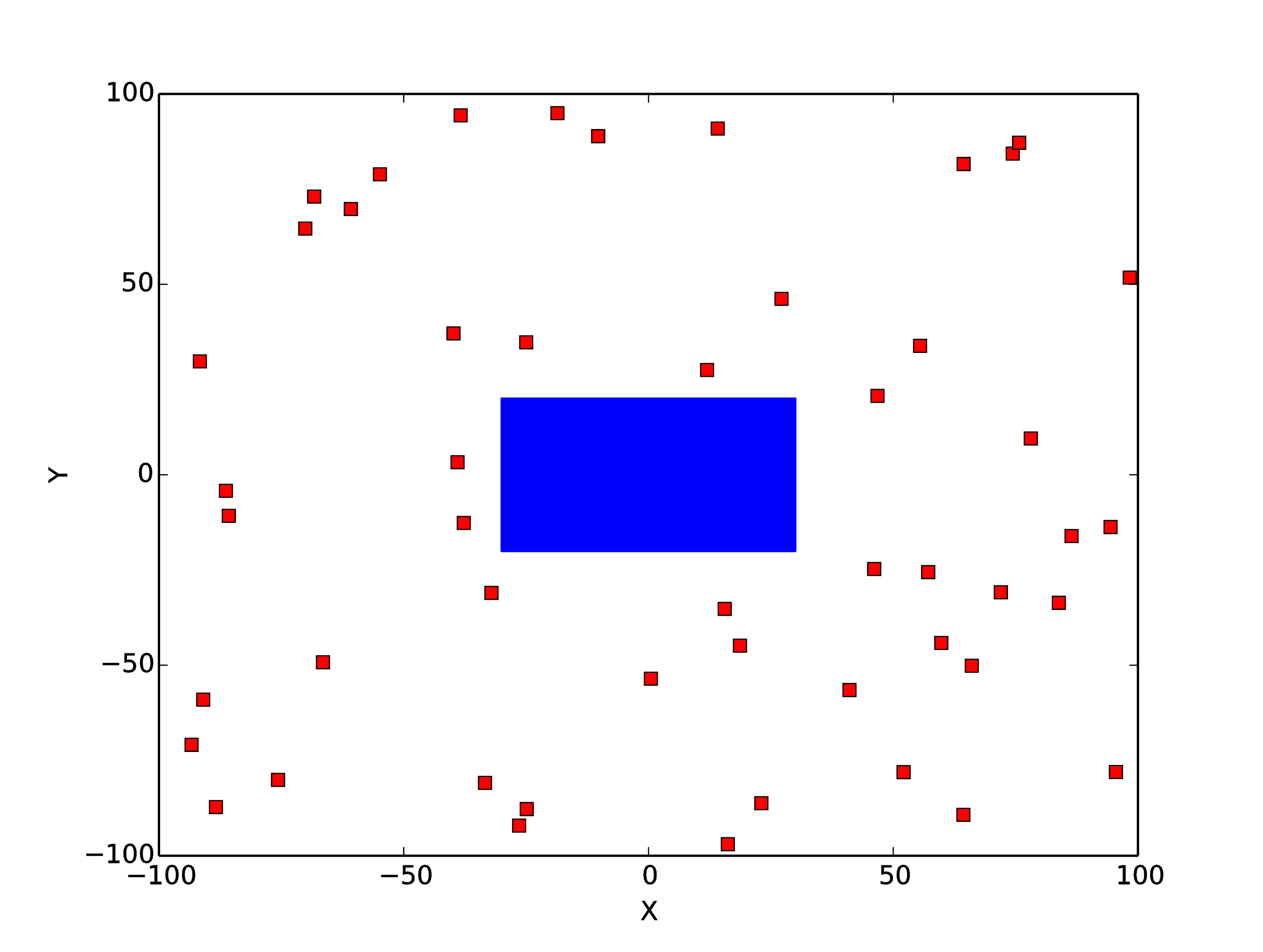}} 
			&

			\subfigure[PAO=9\% \label{fig:World two obstacle}]{\includegraphics[width=.18\linewidth, height=.16\linewidth]{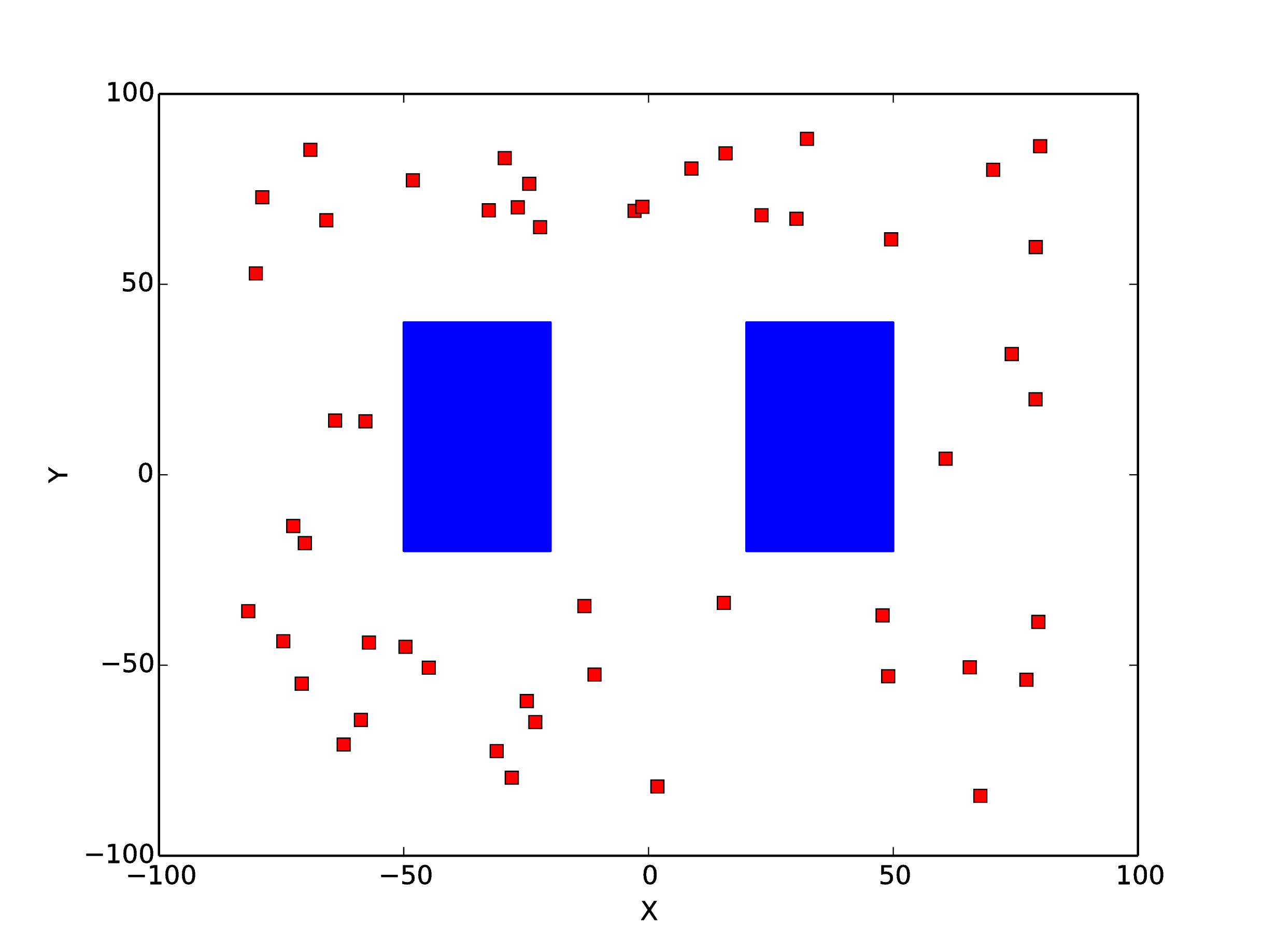}}
			&

			\subfigure[PAO=23.27\% \label{fig:World three obstacle}]{\includegraphics[width=.18\linewidth, height=.16\linewidth]{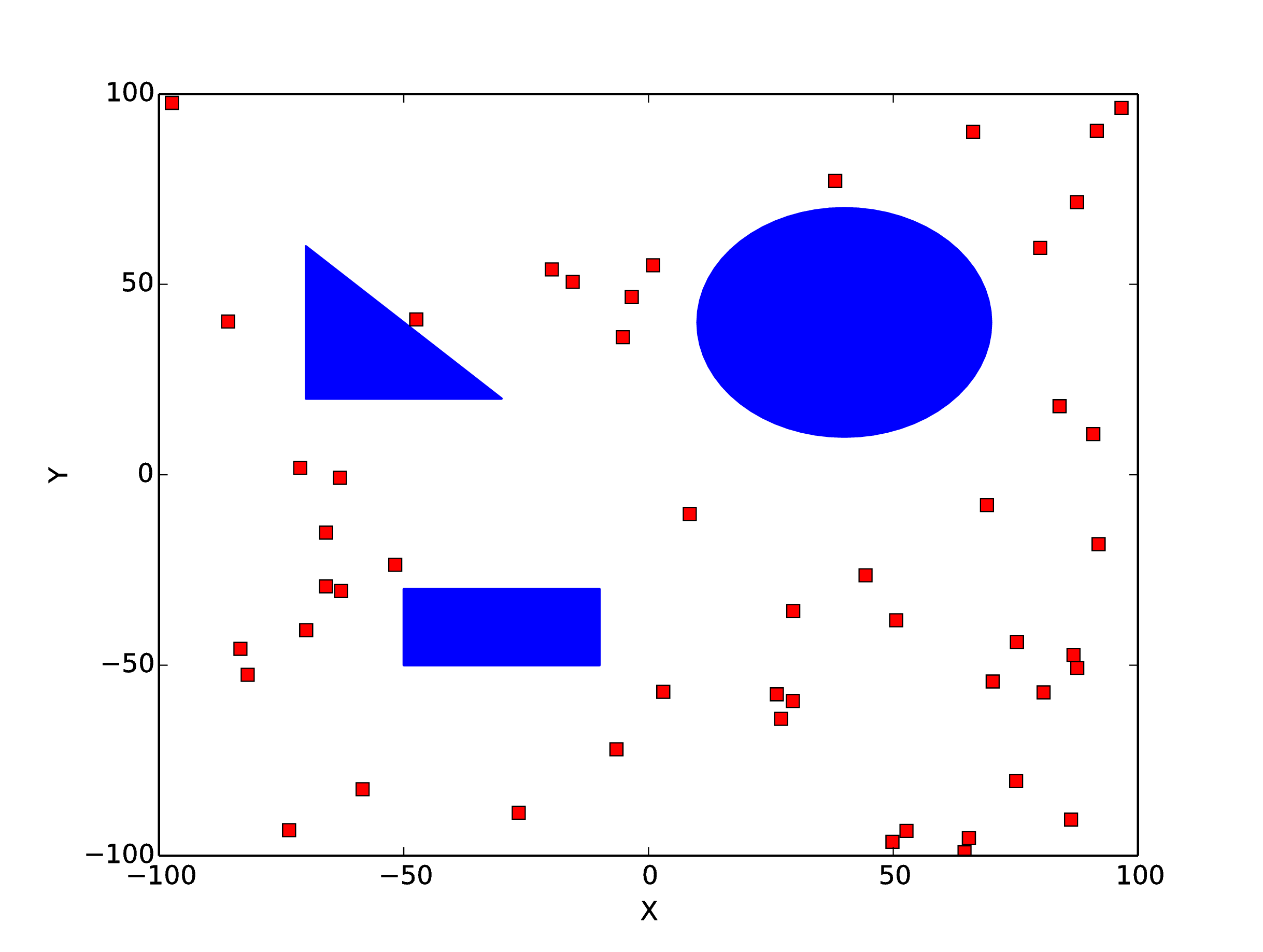}}
			&
			
			\subfigure[PAO=8\%  \label{fig:World four obstacle}]{\includegraphics[width=.18\linewidth, height=.16\linewidth]{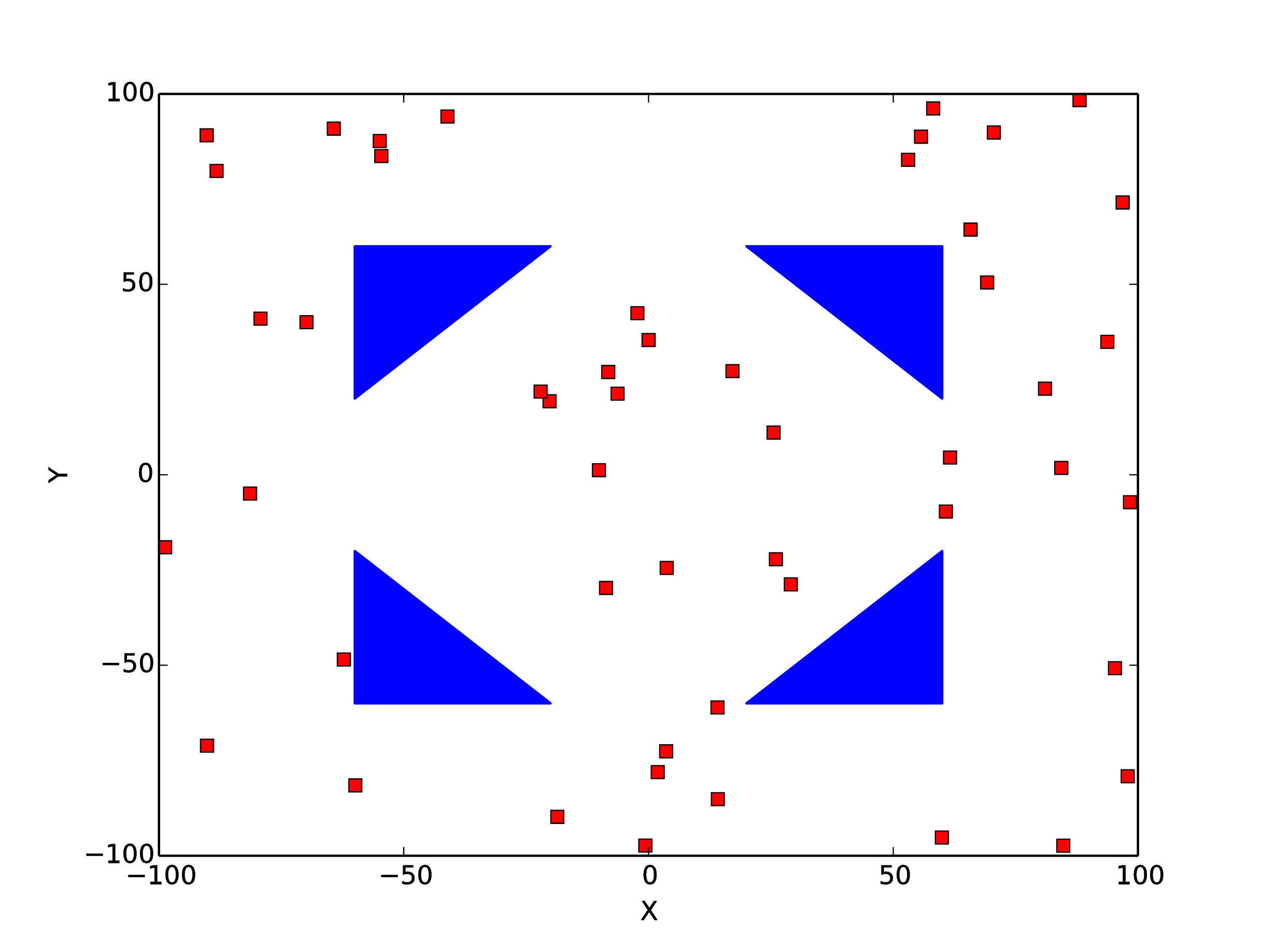}}
			
		\end{tabular}
		\caption{Snapshots of a simulated swarm of robots (red squares) moving through different domains with obstacles (blue shapes).}
		\label{fig:worlds}
	\end{figure*}
	
	
	\begin{figure*}[th]
		
		\begin{tabular}{ccccc}
			\centering	
			\hspace{-3mm}		
			
			\subfigure[PAO=0\% \label{fig:no obstacle pdf computed}]{\includegraphics[width=.18\linewidth, height=.16\linewidth]{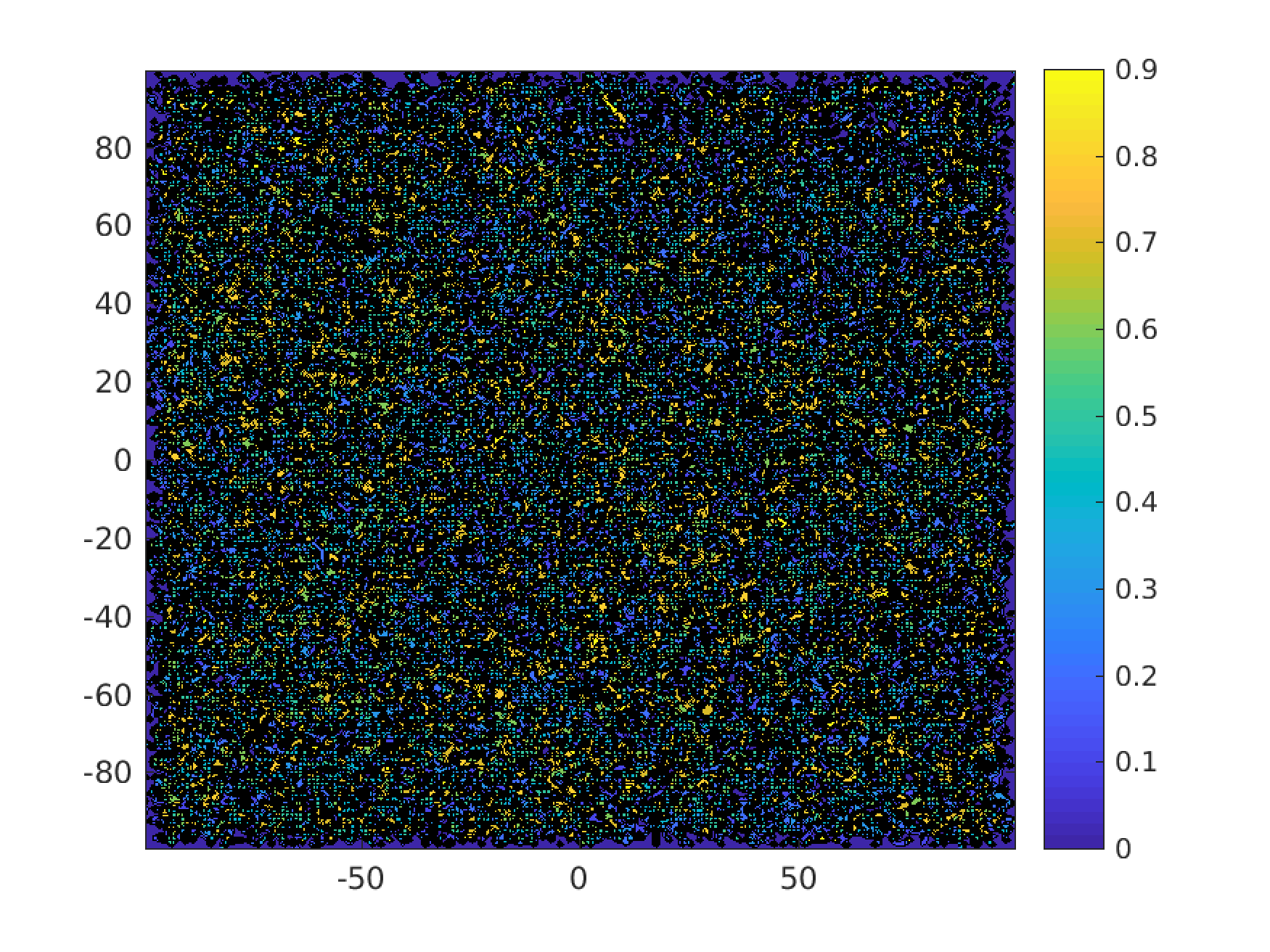}} 
			&
			
			\subfigure[PAO=6\% \label{fig:one obstacle pdf computed}]{\includegraphics[width=.18\linewidth, height=.16\linewidth]{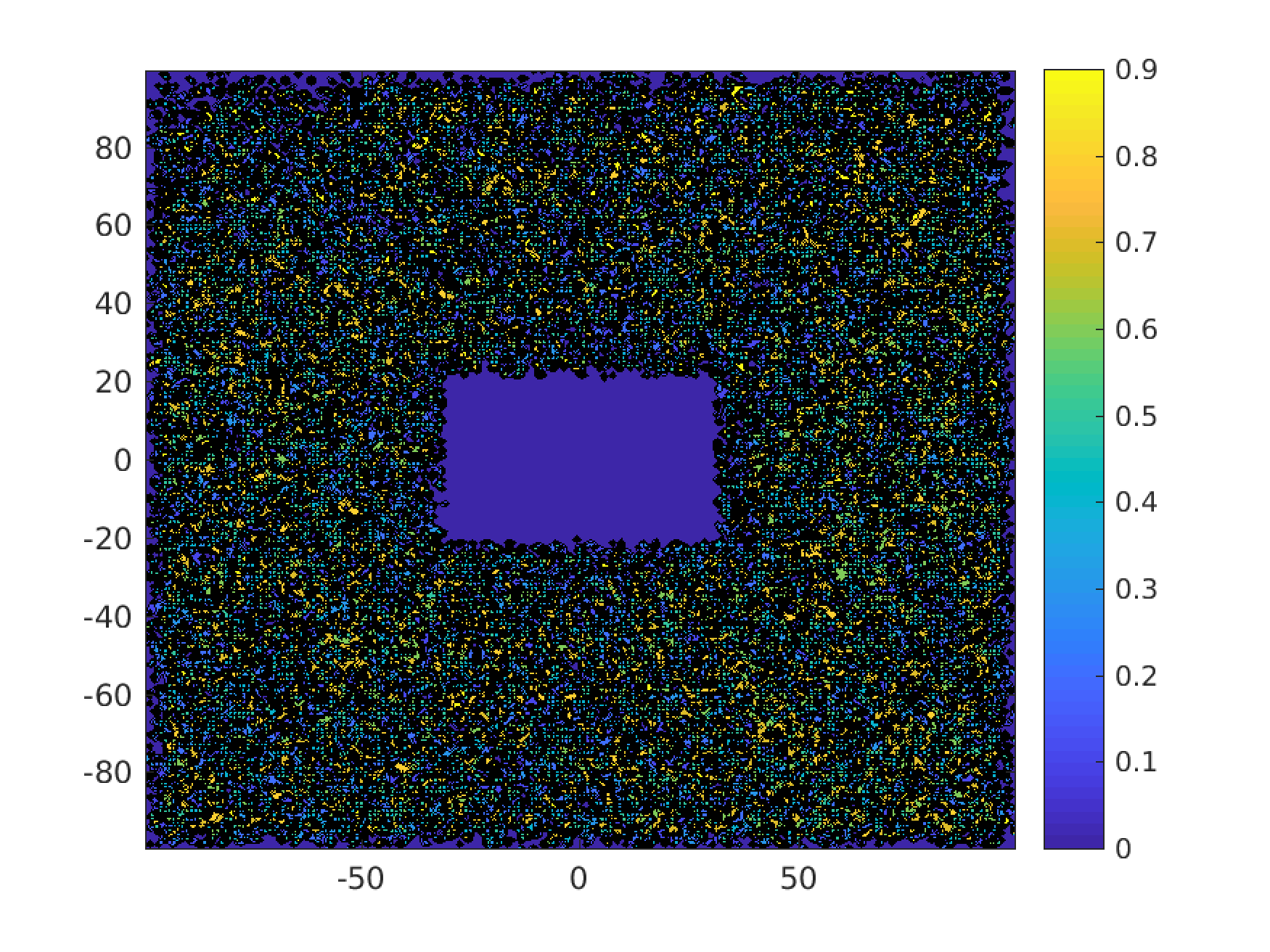}} 
			&

			\subfigure[PAO=9\% \label{fig:two obstacle pdf computed}]{\includegraphics[width=.18\linewidth, height=.16\linewidth]{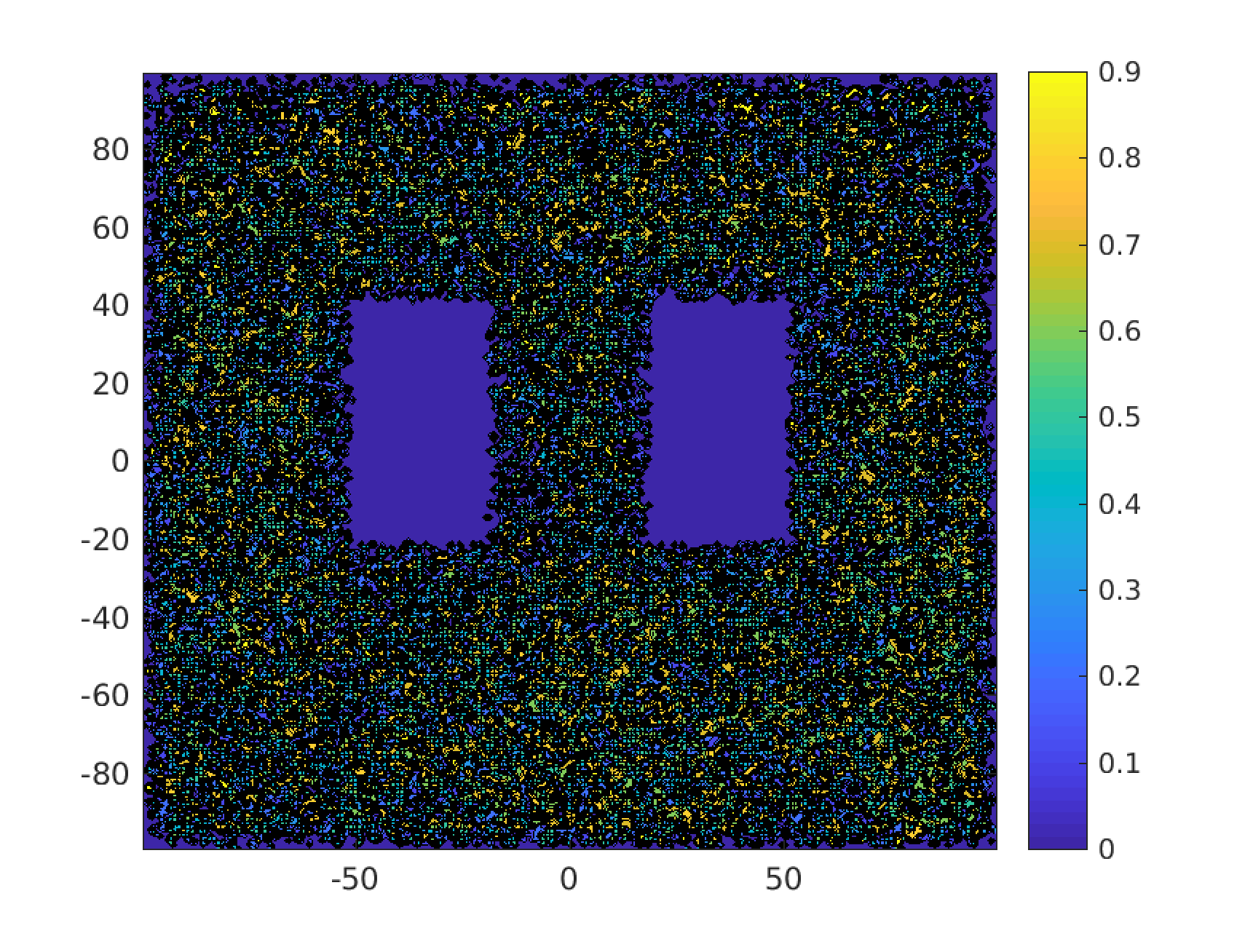}}
			&

			\subfigure[PAO=23.27\% \label{fig:three obstacle pdf computed}]{\includegraphics[width=.18\linewidth, height=.16\linewidth]{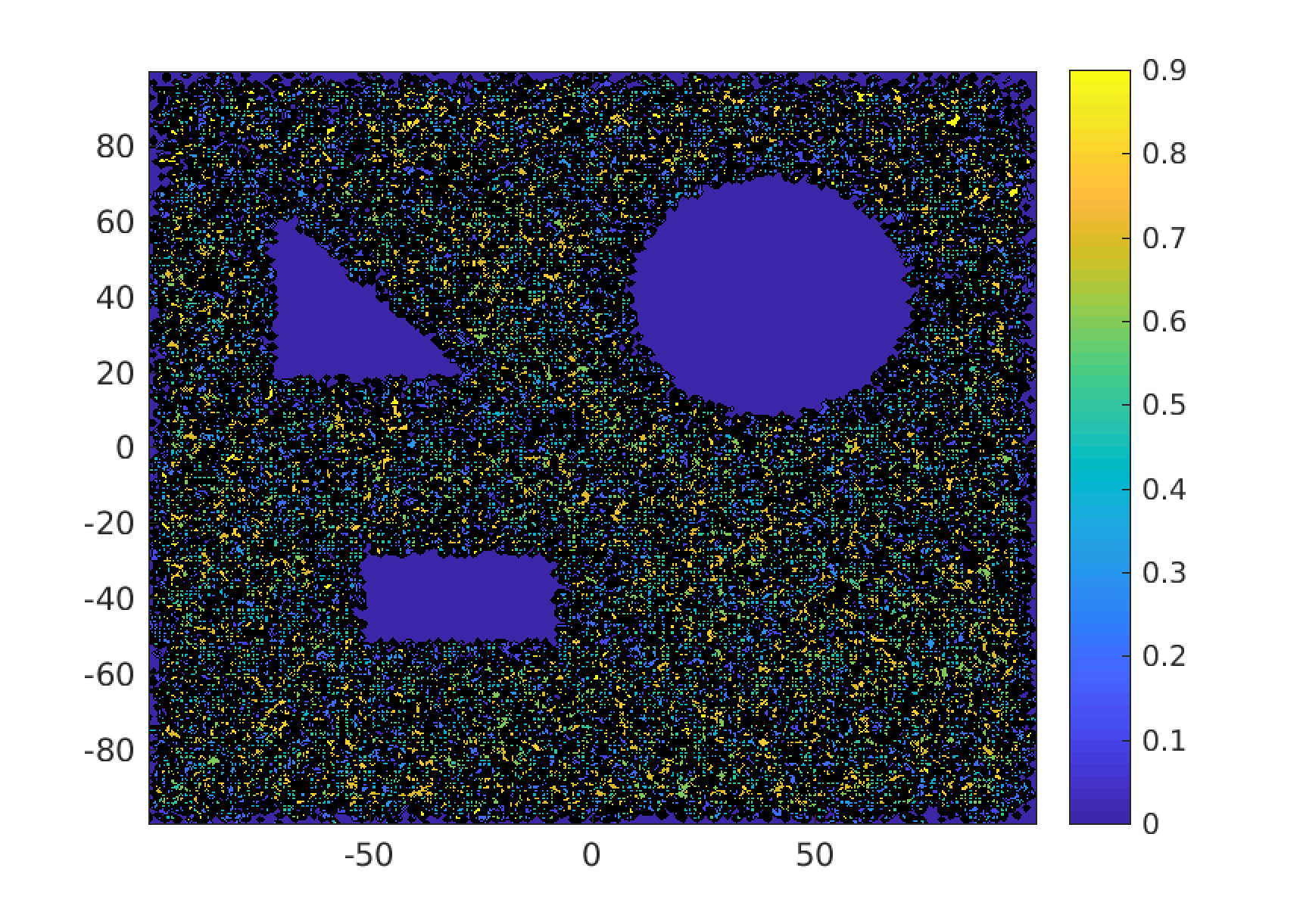}}
			&
			
			\subfigure[PAO=8\% \label{fig:four obstacle pdf computed}]{\includegraphics[width=.18\linewidth, height=.16\linewidth]{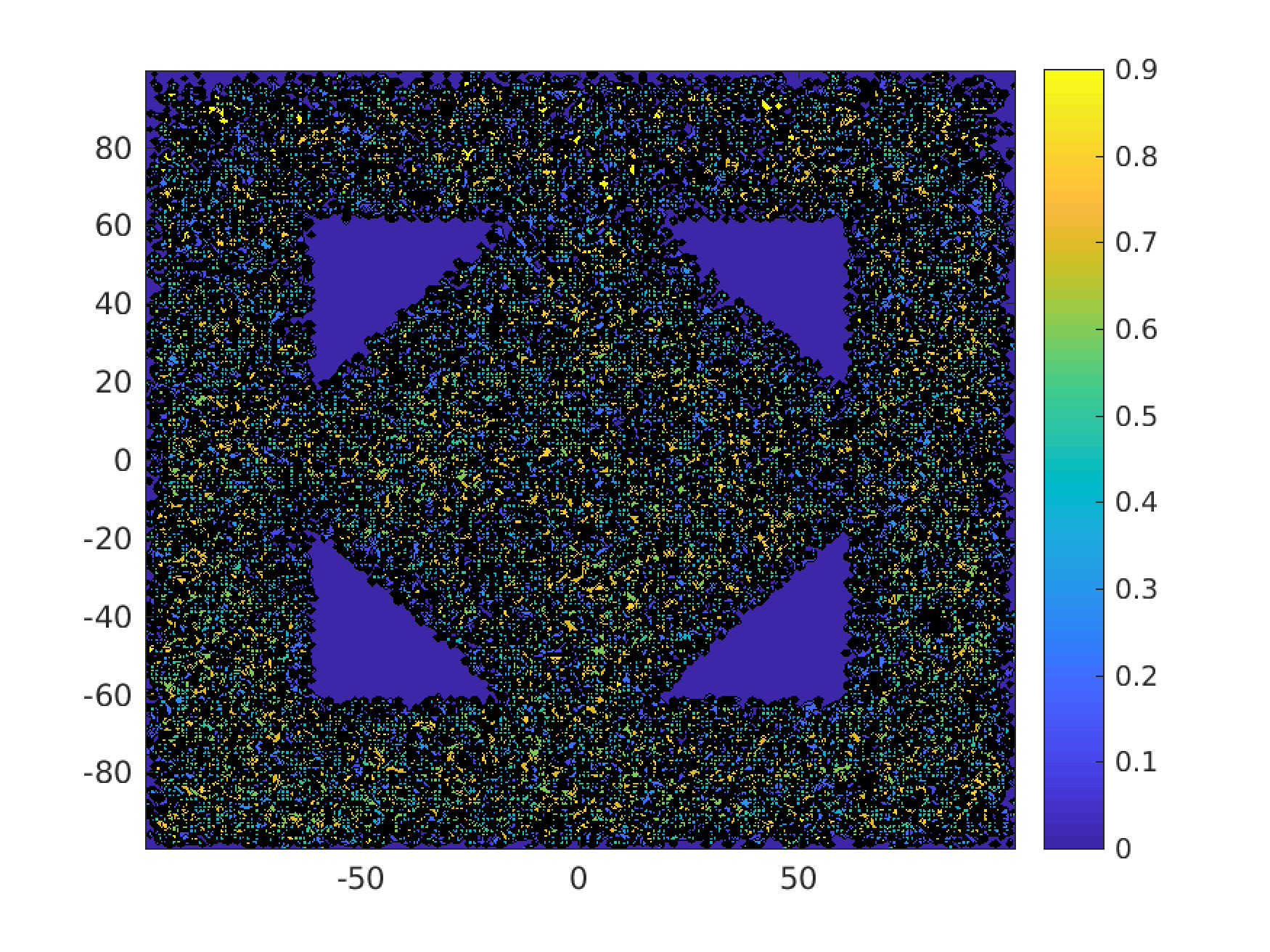}}
			
		\end{tabular}
		\caption{Contour plots of $p^f_i$, the probability that grid cell $m_i$ is free, over all grid cells of the discretized domains generated after the step described in \autoref{subsec:Computation of Density Function}. Colorbar values range from 0 to 0.9.}  
		\label{fig:pdf computed}
	\end{figure*}
	
	
	
	\begin{figure*}[th]
		
		\begin{tabular}{ccccc}
			\centering	
			\hspace{-3mm}		
			
			\subfigure[PAO=0\% \label{fig:no obstacle pdf filtered}]{\includegraphics[width=.18\linewidth, height=.16\linewidth]{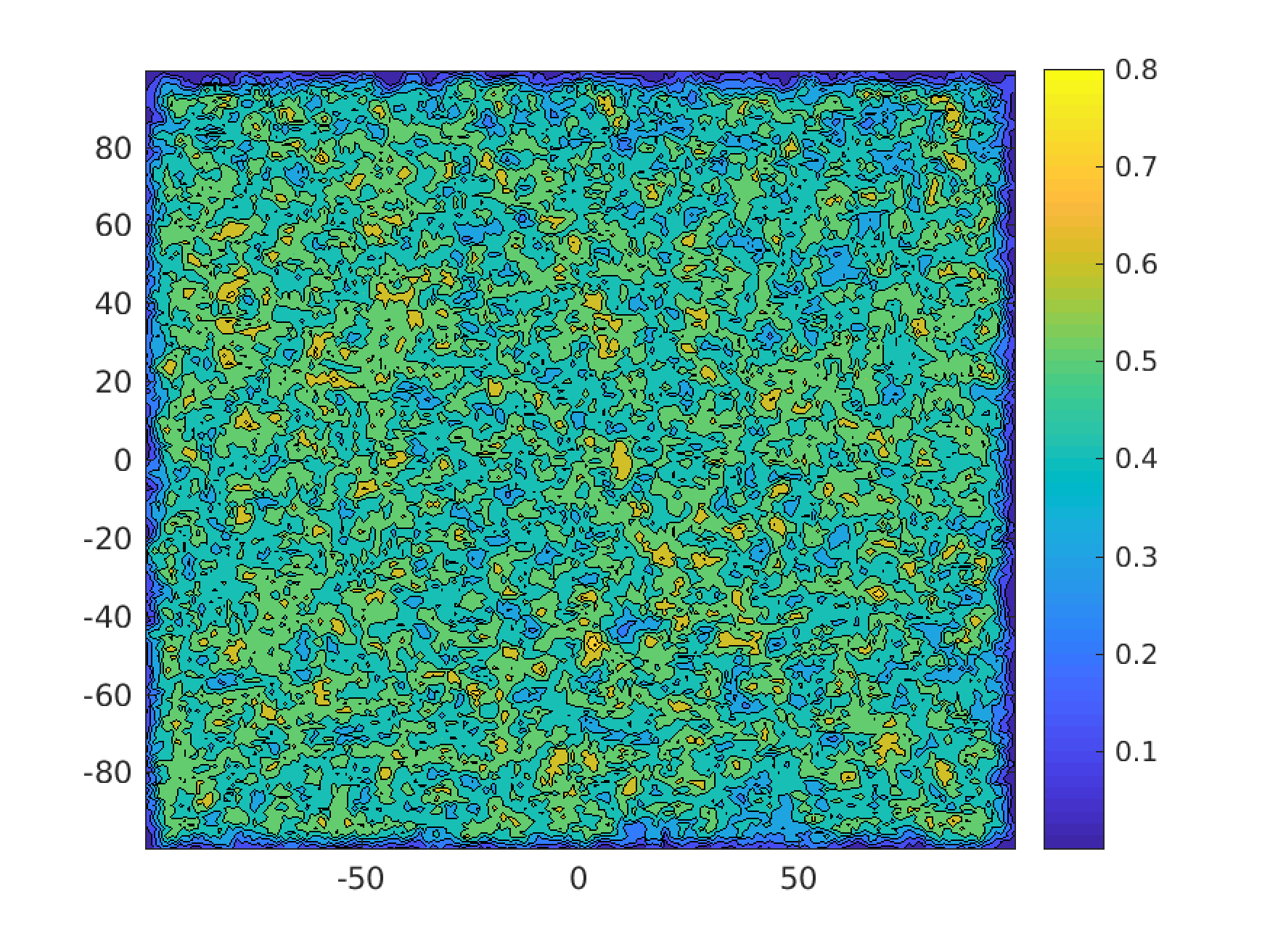}} 
			&
			
			\subfigure[PAO=6\% \label{fig:one obstacle pdf filtered}]{\includegraphics[width=.18\linewidth, height=.16\linewidth]{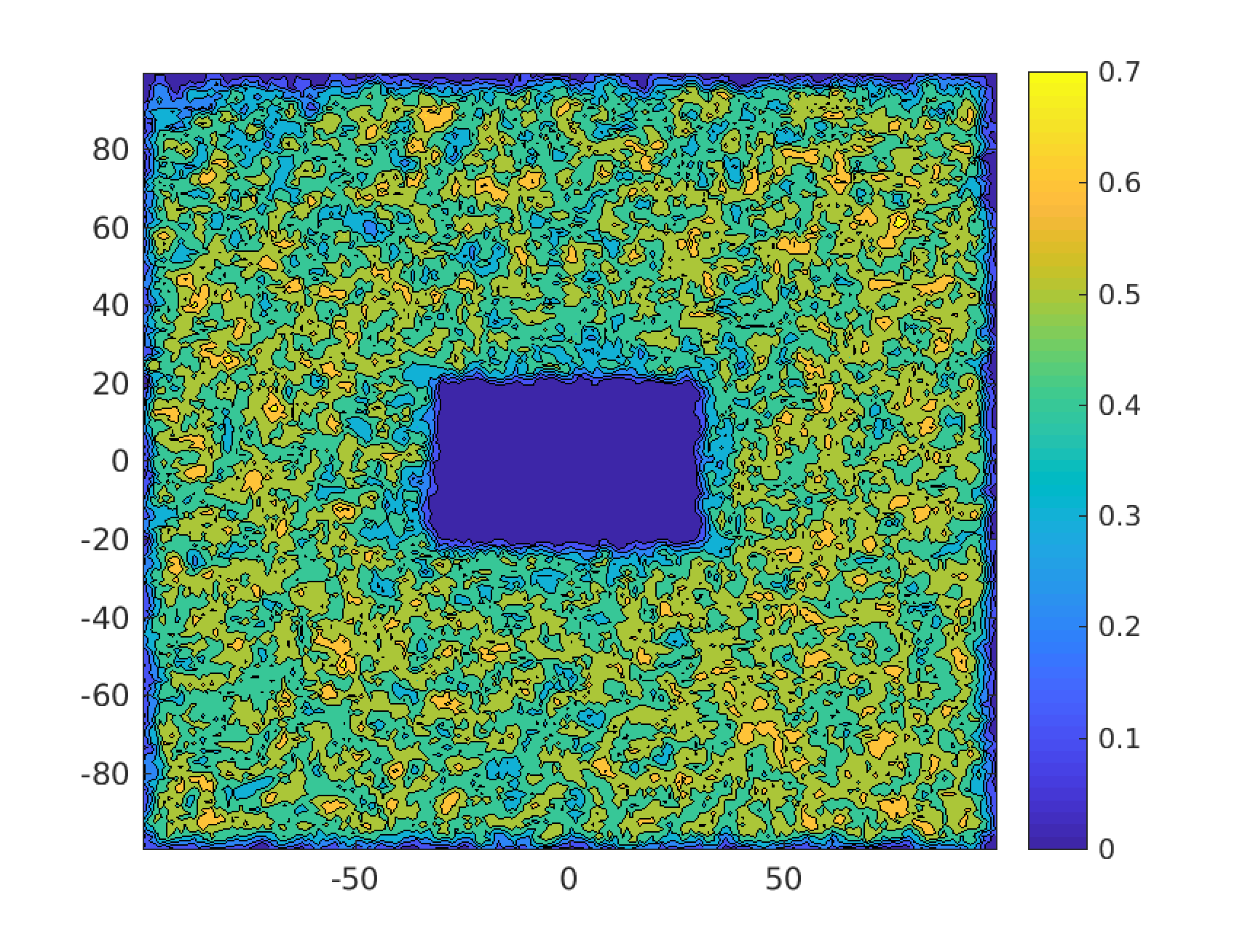}} 
			&

			\subfigure[PAO=9\% \label{fig:two obstacle pdf filtered}]{\includegraphics[width=.18\linewidth, height=.16\linewidth]{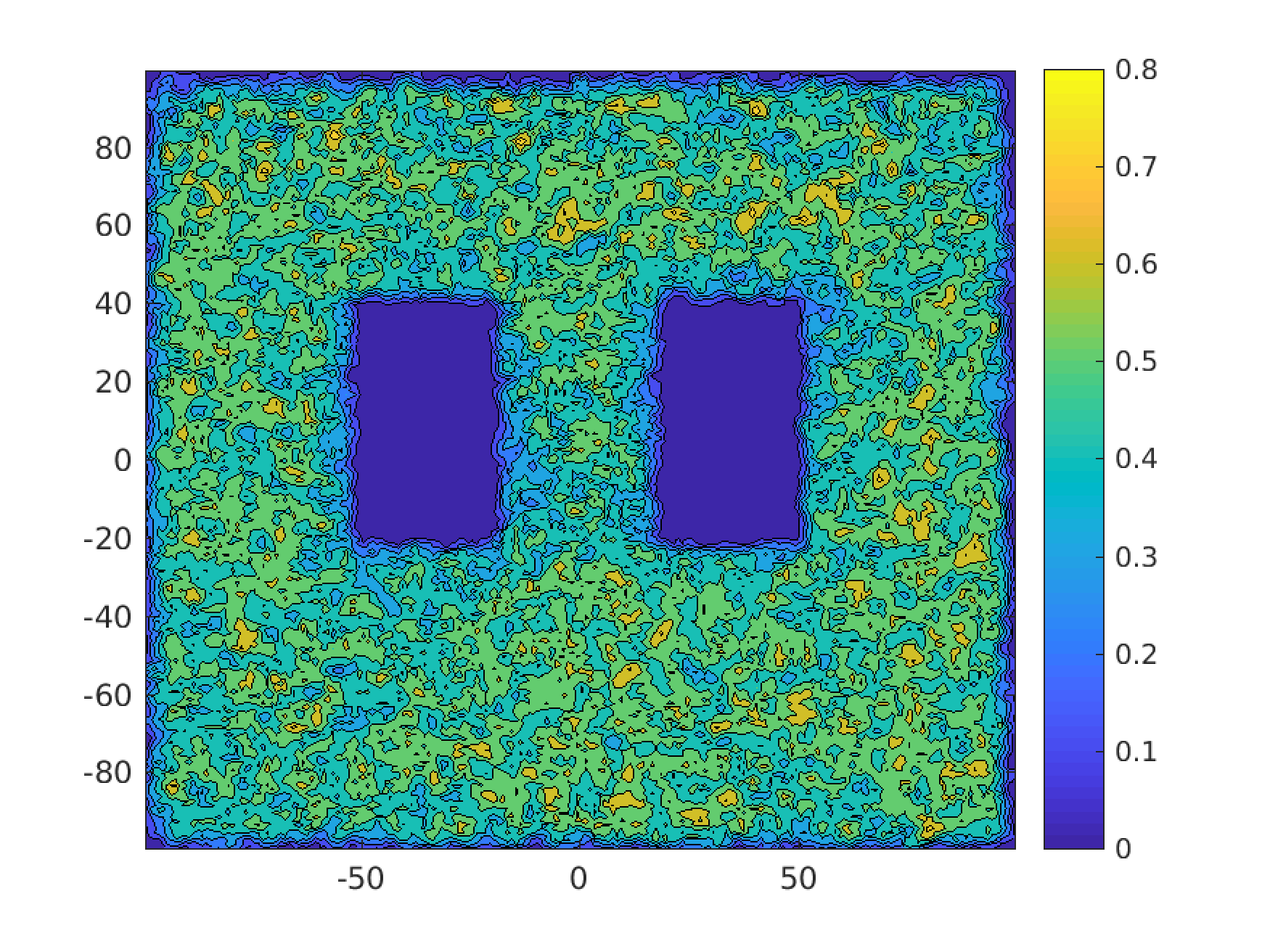}}
			&

			\subfigure[PAO=23.27\% \label{fig:three obstacle pdf filtered}]{\includegraphics[width=.18\linewidth, height=.16\linewidth]{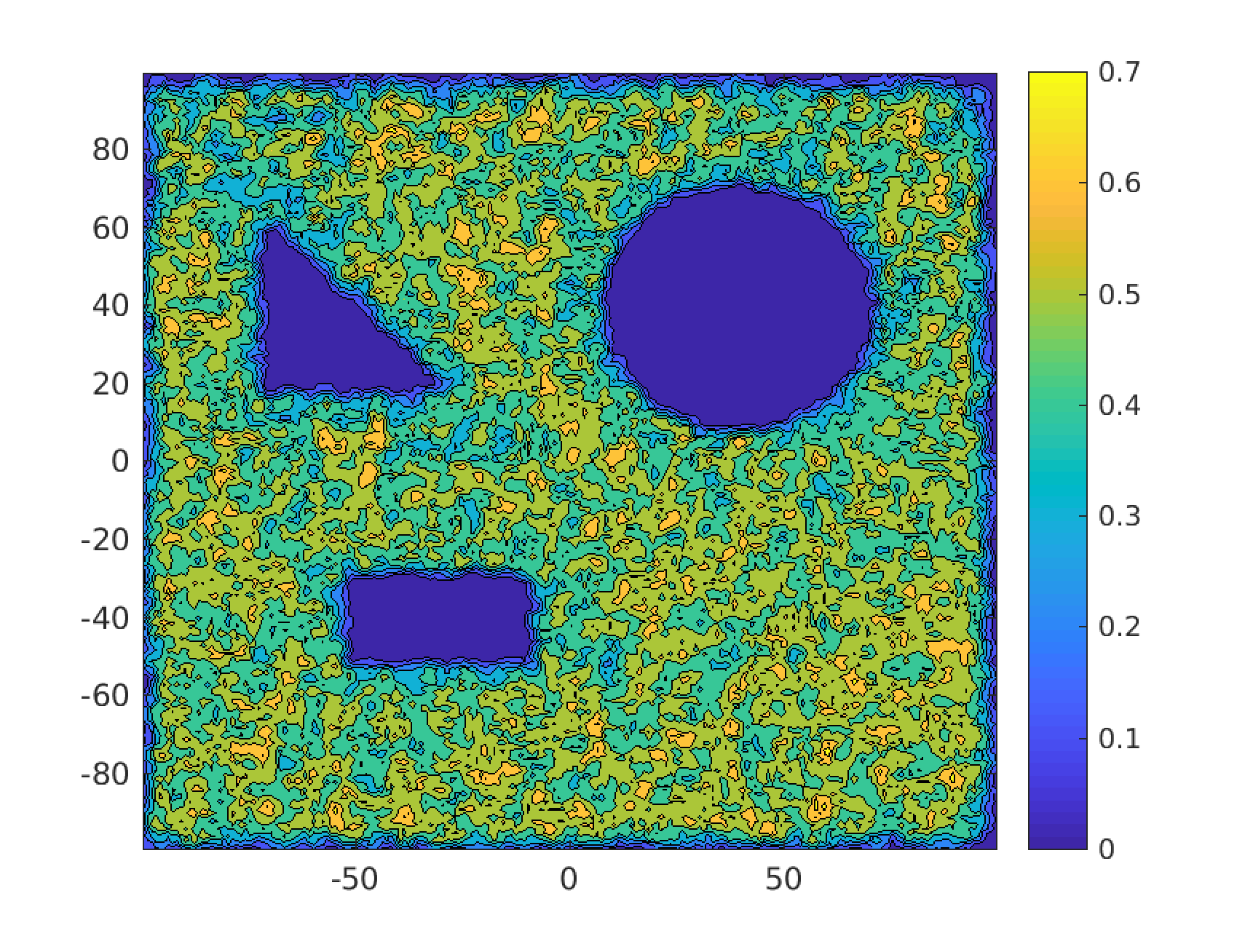}}
			&
			
			\subfigure[PAO=8\% \label{fig:four obstacle pdf filtered}]{\includegraphics[width=.18\linewidth, height=.16\linewidth]{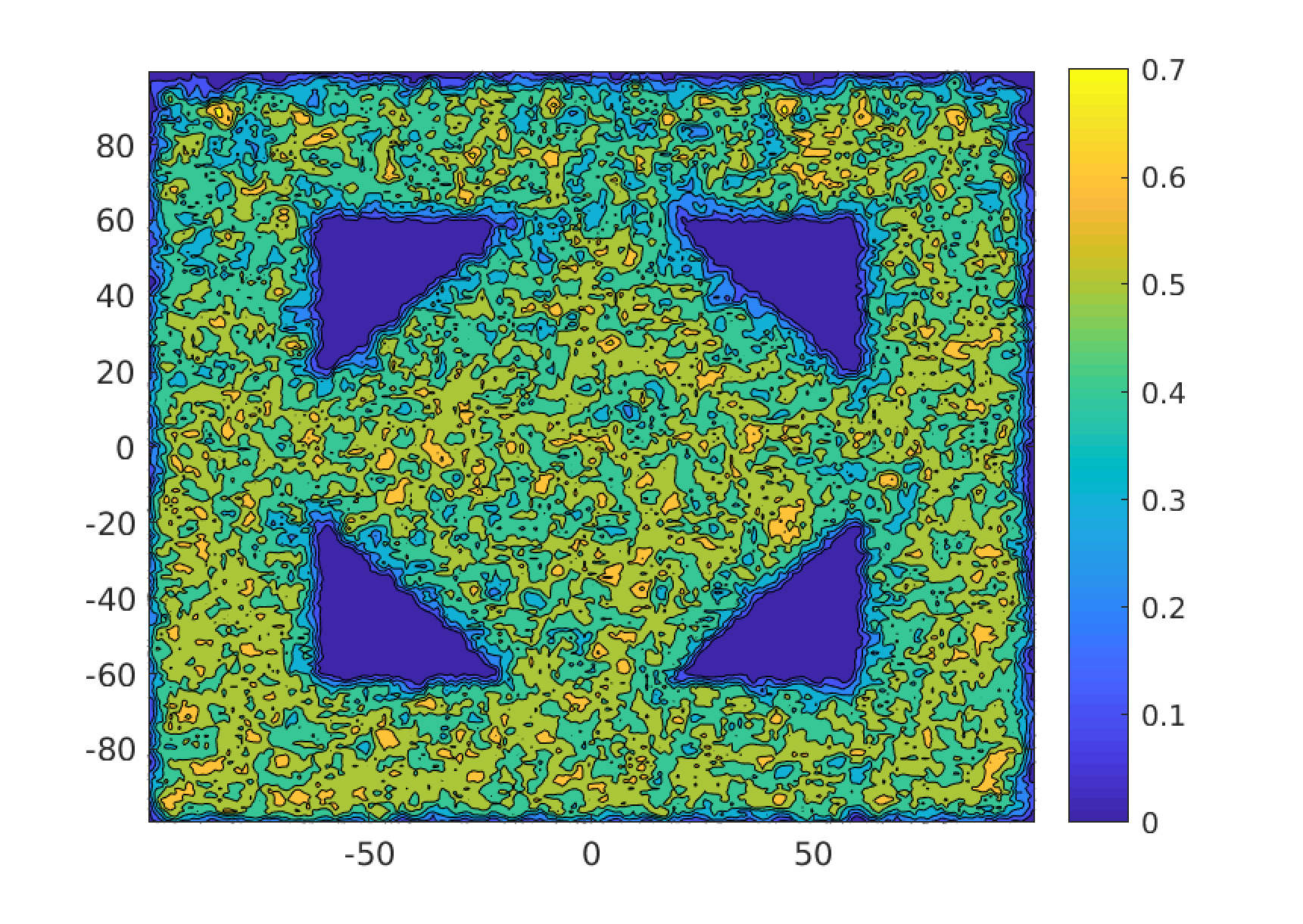}}
			
		\end{tabular}
		\caption{Contour plots of the filtered $p^f_i$ shown in \autoref{fig:pdf computed}, as described in \autoref{subsec:Computation of Density Function}.}  
		\label{fig:pdf Filtered}
	\end{figure*}
	
	

	\begin{figure*}[th]
		
		\begin{tabular}{ccccc}
			\centering	
			\hspace{-3mm}		
			\subfigure[PAO=0\% \label{fig:no obstacle barcode}]{\includegraphics[width=.18\linewidth, height=.2\linewidth]{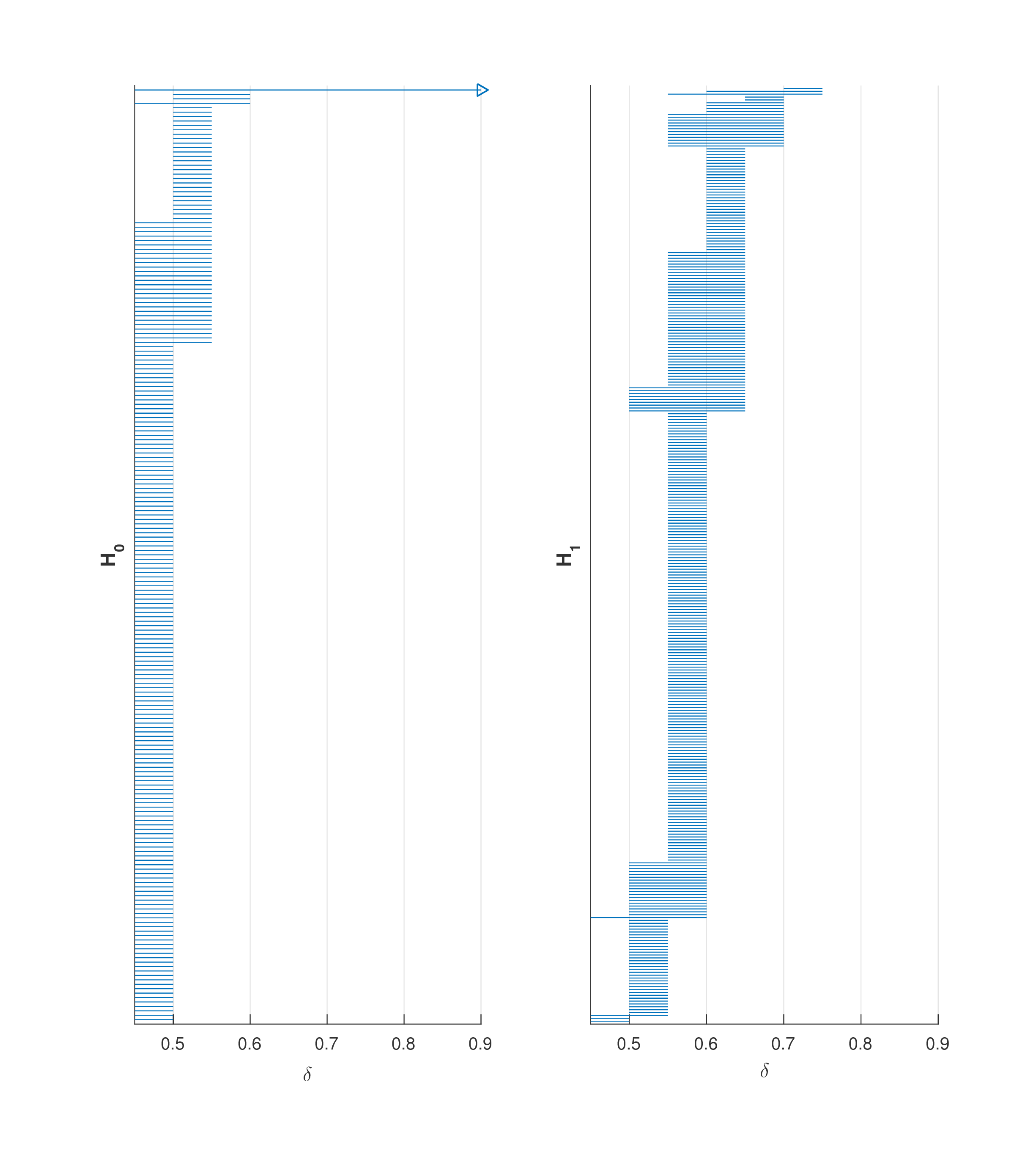}} 
			&
			
			\subfigure[PAO=6\% \label{fig:one obstacle barcode}]{\includegraphics[width=.18\linewidth, height=.2\linewidth]{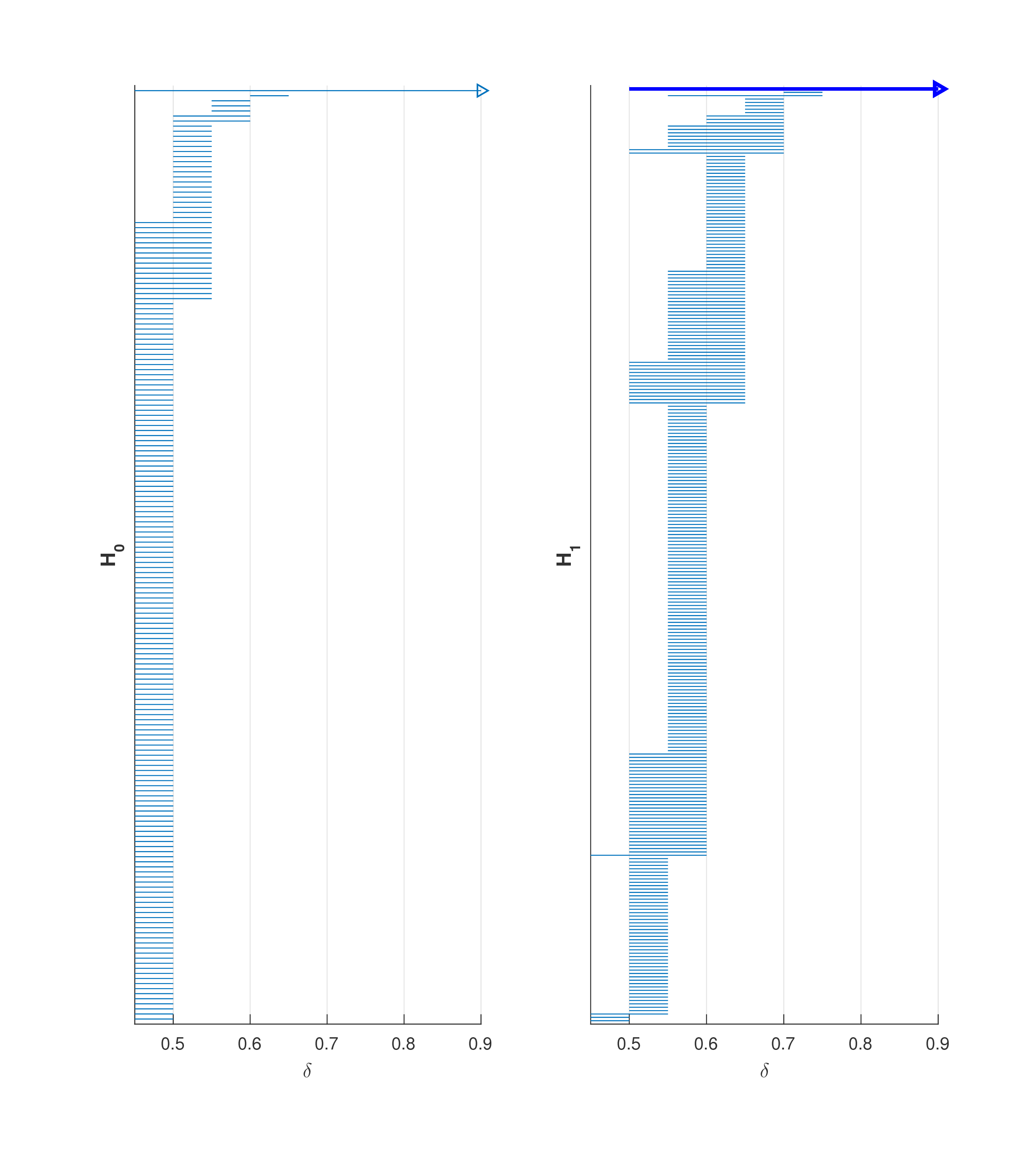}} 
			&

			
			\subfigure[PAO=9\% \label{fig:two obstacle barcode}]{\includegraphics[width=.18\linewidth, height=.2\linewidth]{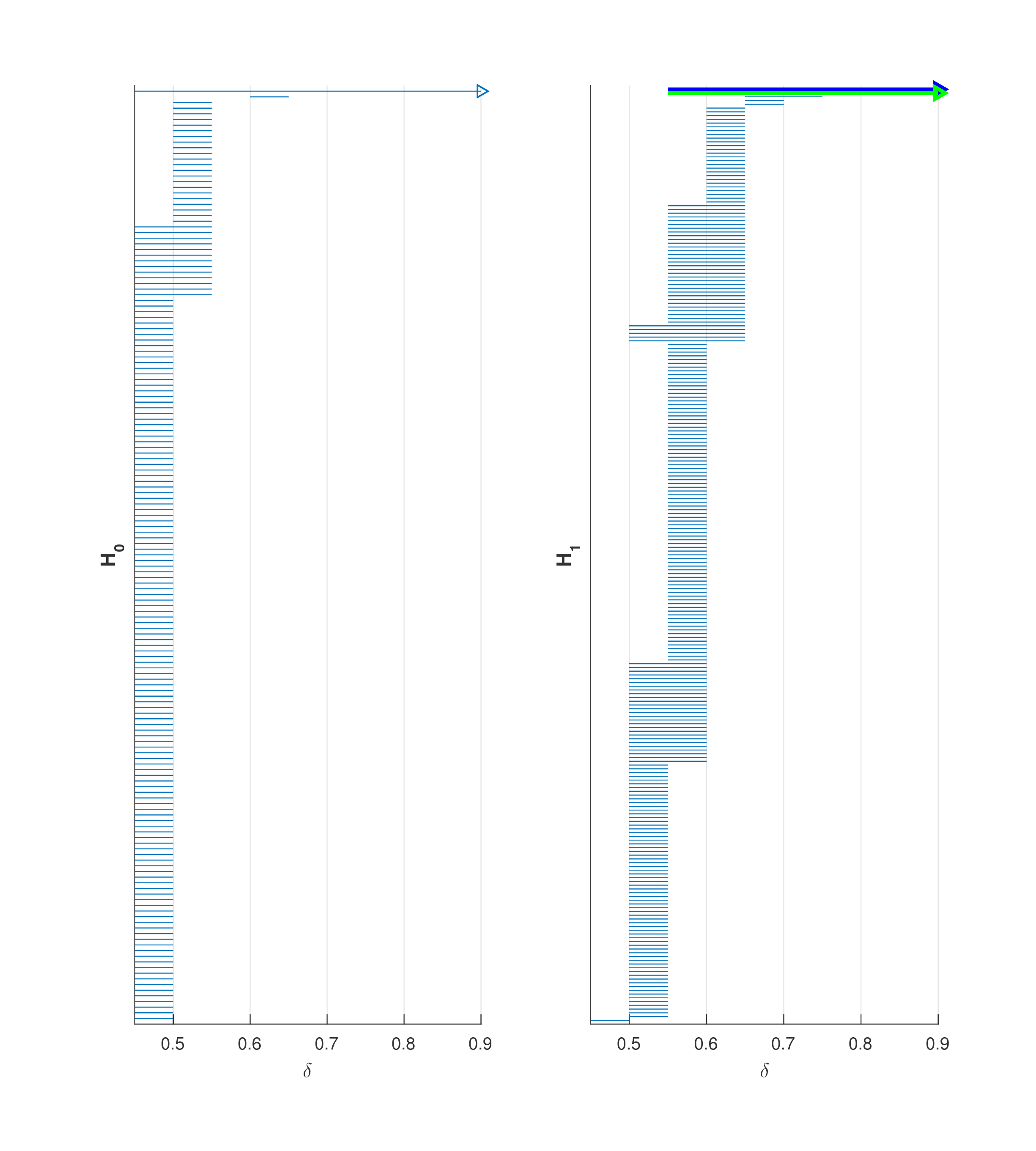}}
			&

			
			\subfigure[PAO=23.27\% \label{fig:three obstacle barcode}]{\includegraphics[width=.18\linewidth, height=.2\linewidth]{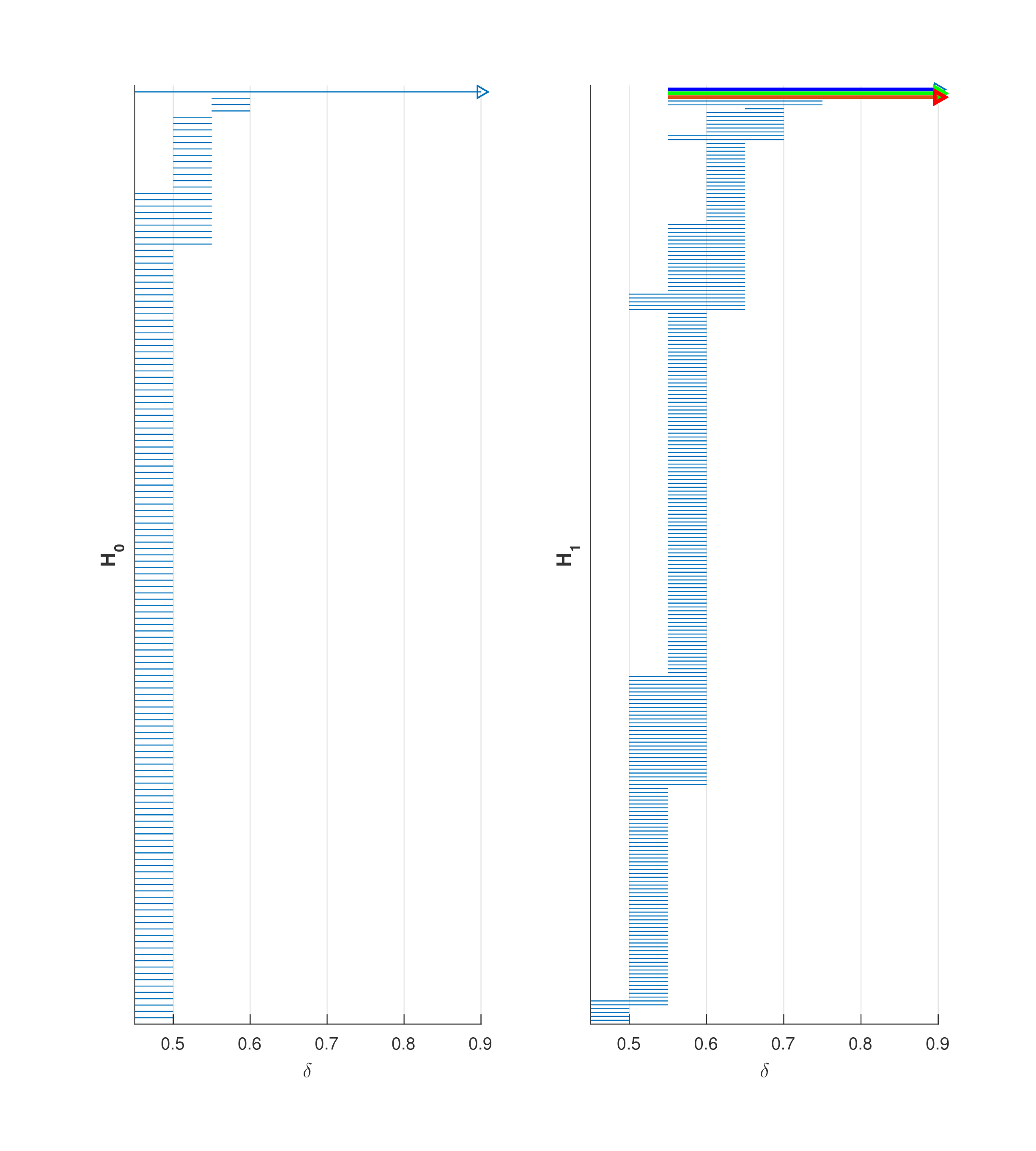}}
			&
			
			
			\subfigure[PAO=8\% \label{fig:four obstacle barcode}]{\includegraphics[width=.18\linewidth, height=.2\linewidth]{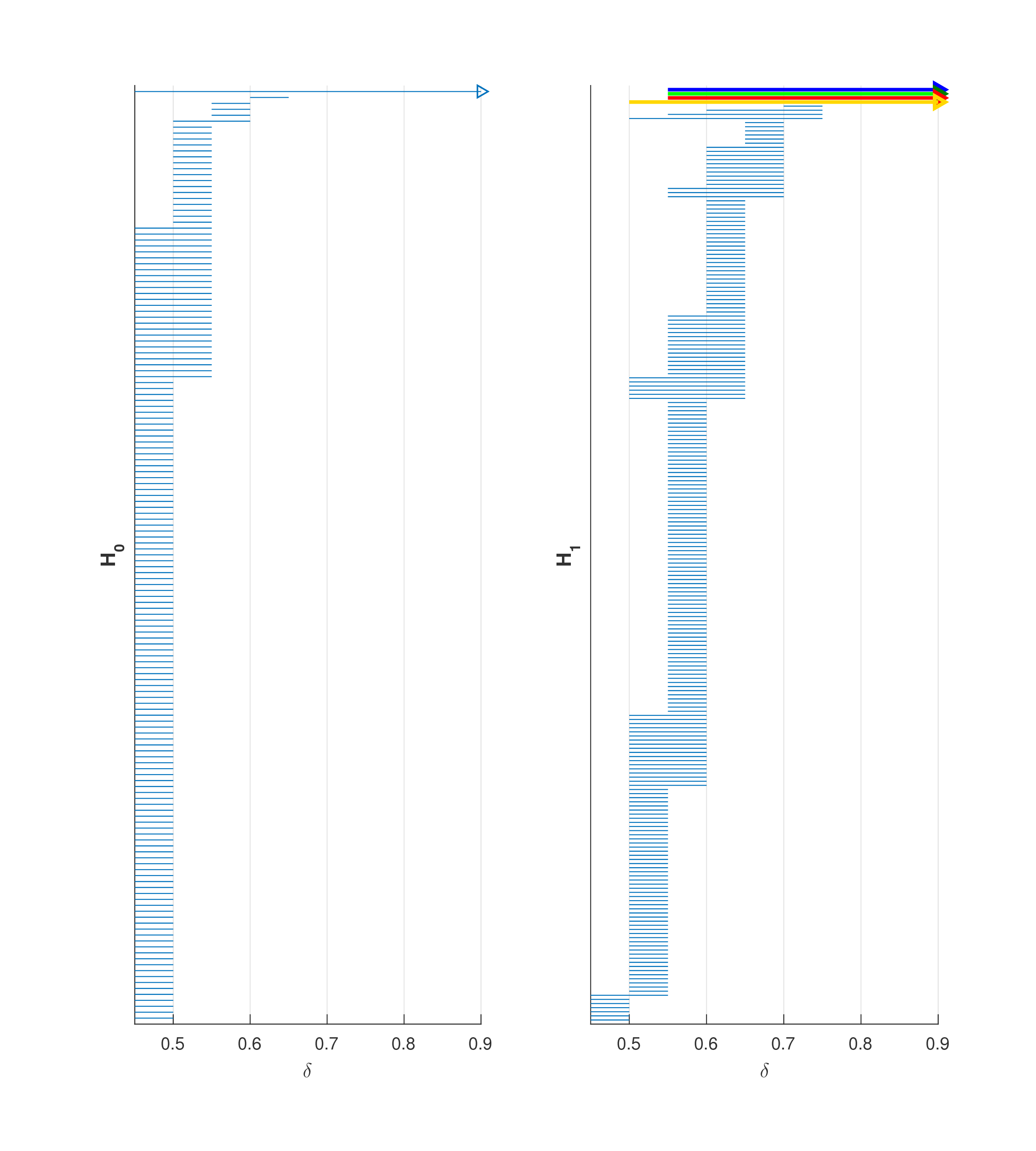}} 
			
		\end{tabular}
		\caption{Barcode diagram for each domain, generated from the filtration described in \autoref{subsec:Thresholding the Density Function to Generate the Map}, with $\delta_{cls}$ computed for each case as $\delta_{cls}=0.75$.}
		\label{fig:barcode}
	\end{figure*}

	
	
	\begin{figure*}[th]
		
		\begin{tabular}{ccccc}
			\centering	
			\hspace{-3mm}		
			
			\subfigure[PAO=0\% \label{fig:no obstacle map}]{\includegraphics[width=.18\linewidth, height=.16\linewidth]{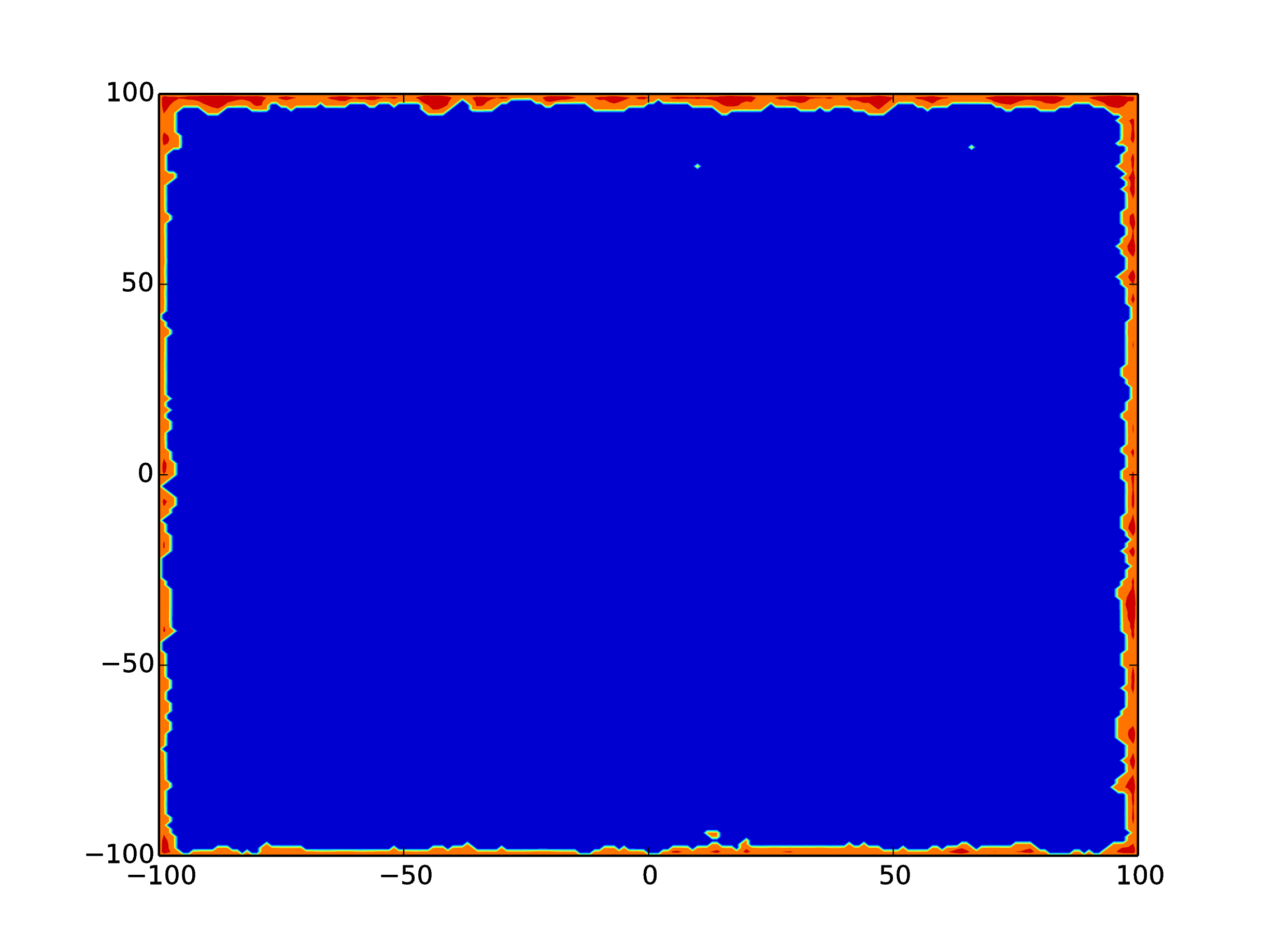}} 
			&
			
			\subfigure[PAO=6\% \label{fig:one obstacle map}]{\includegraphics[width=.18\linewidth, height=.16\linewidth]{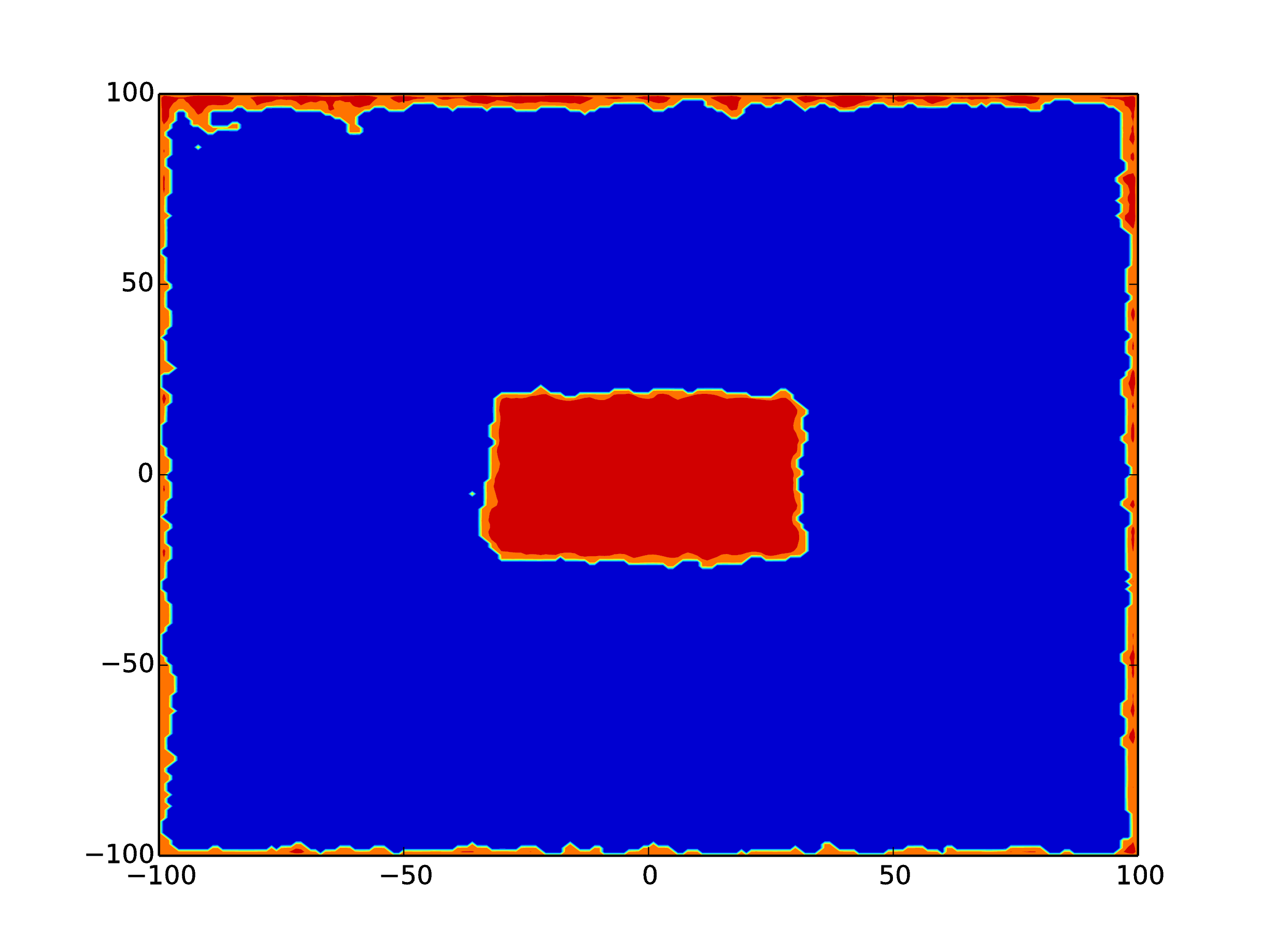}} 
			&

			\subfigure[PAO=9\% \label{fig:two obstacle map}]{\includegraphics[width=.18\linewidth, height=.16\linewidth]{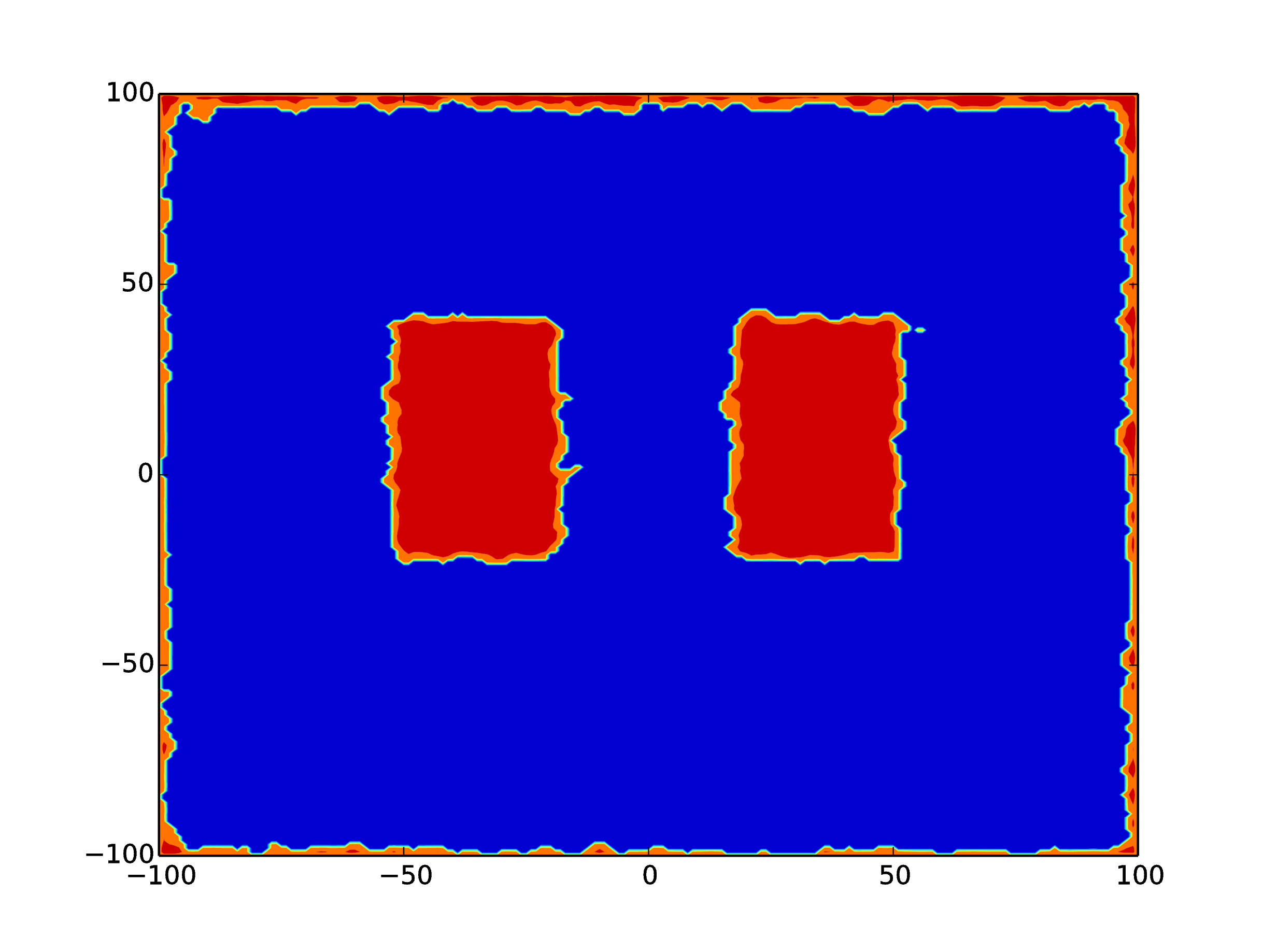}}
			&

			\subfigure[PAO=23.27\% \label{fig:three obstacle map}]{\includegraphics[width=.18\linewidth, height=.16\linewidth]{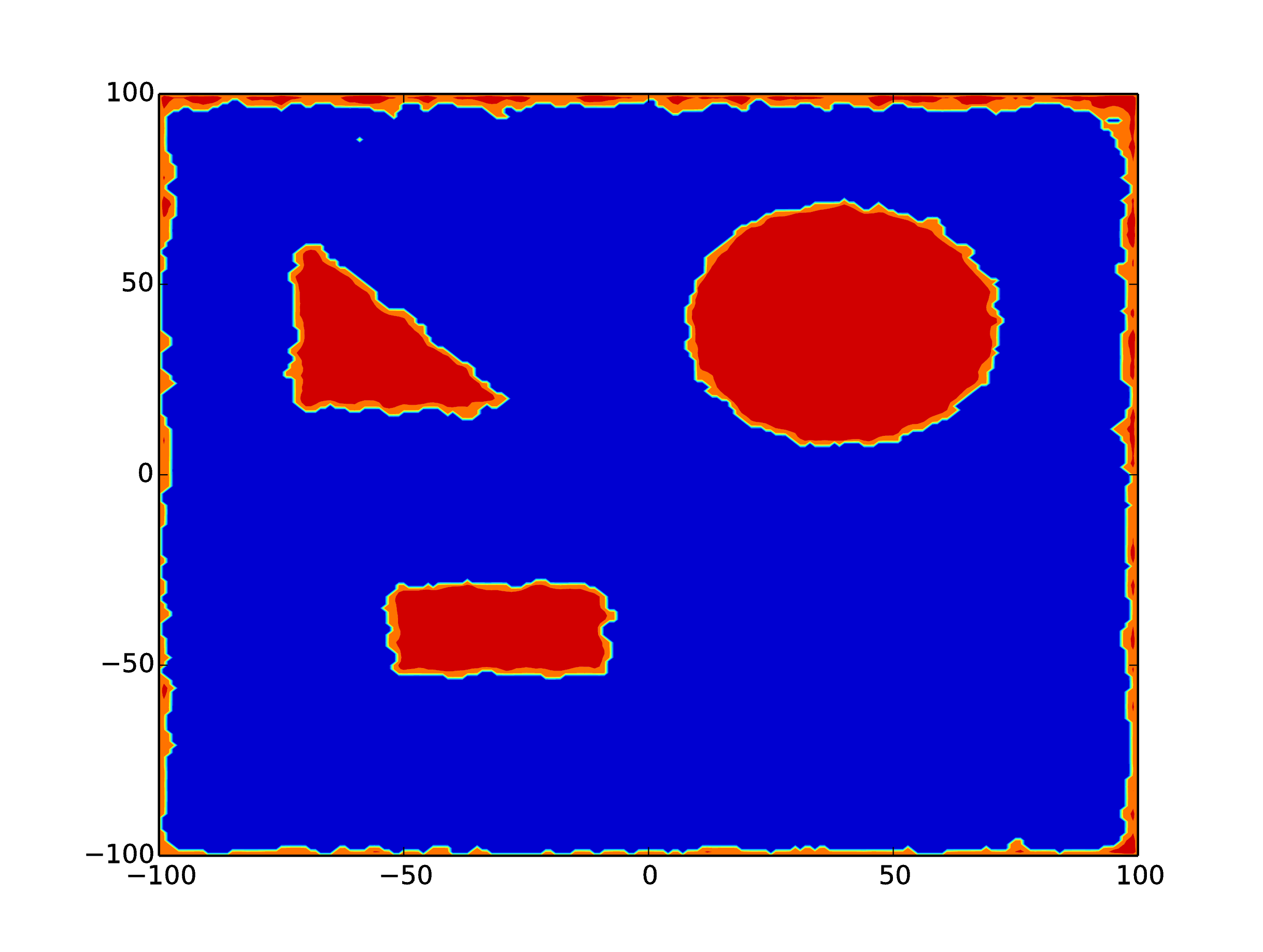}}
			&
			
			\subfigure[PAO=8\% \label{fig:four obstacle map}]{\includegraphics[width=.18\linewidth, height=.16\linewidth]{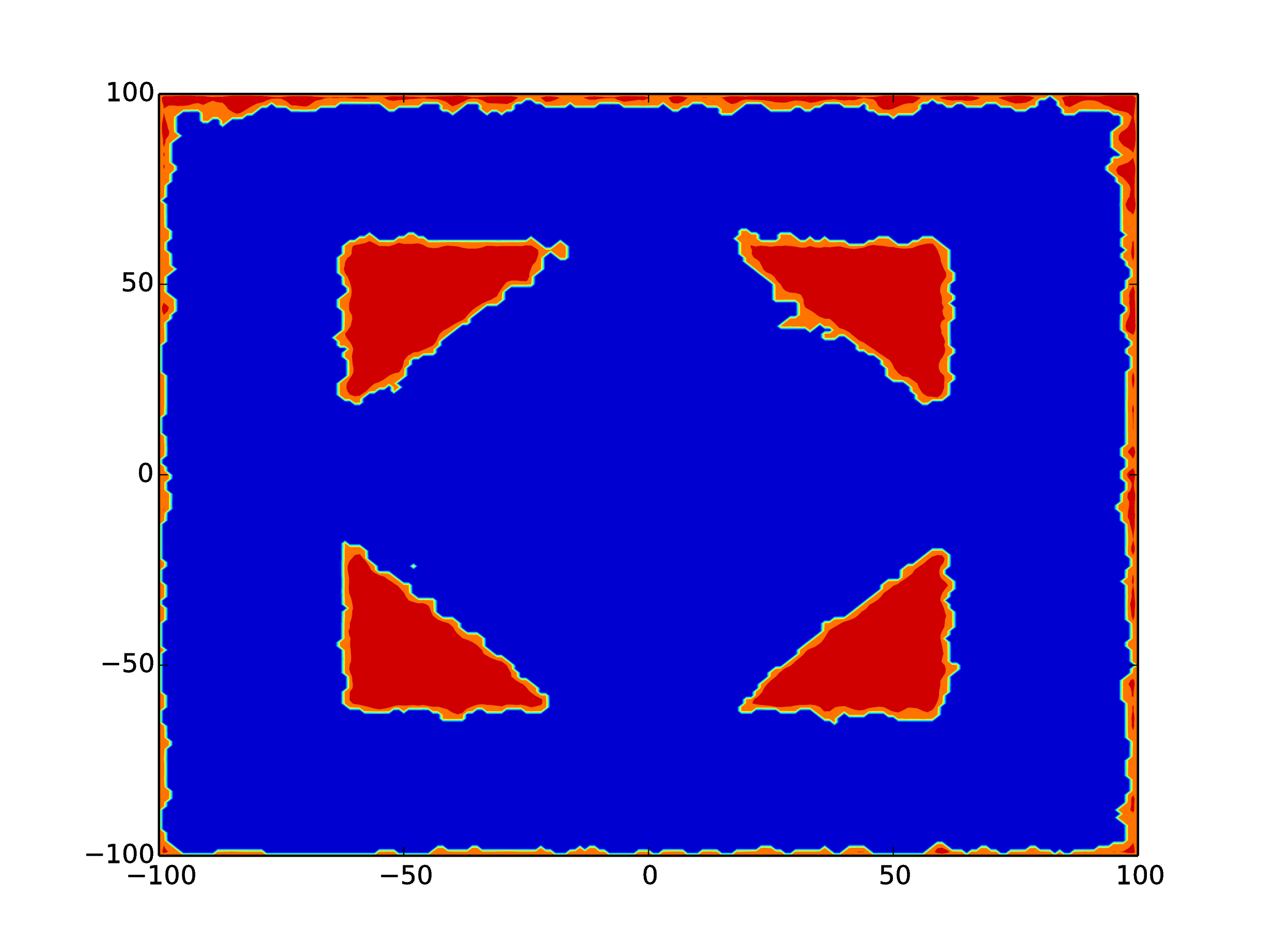}}
			
		\end{tabular}
		\caption{Contour plots of the thresholded map based on the thresholds computed using the TDA technique described in \autoref{subsec:Thresholding the Density Function to Generate the Map}.}
		\label{fig:maps}
	\end{figure*}
	
	
	
	\begin{figure*}[th]
		
		\begin{tabular}{ccccc}
			\centering	
			\hspace{-3mm}		
			
			\subfigure[PAO=0\% \label{fig:no obstacle error}]{\includegraphics[width=.18\linewidth, height=.16\linewidth]{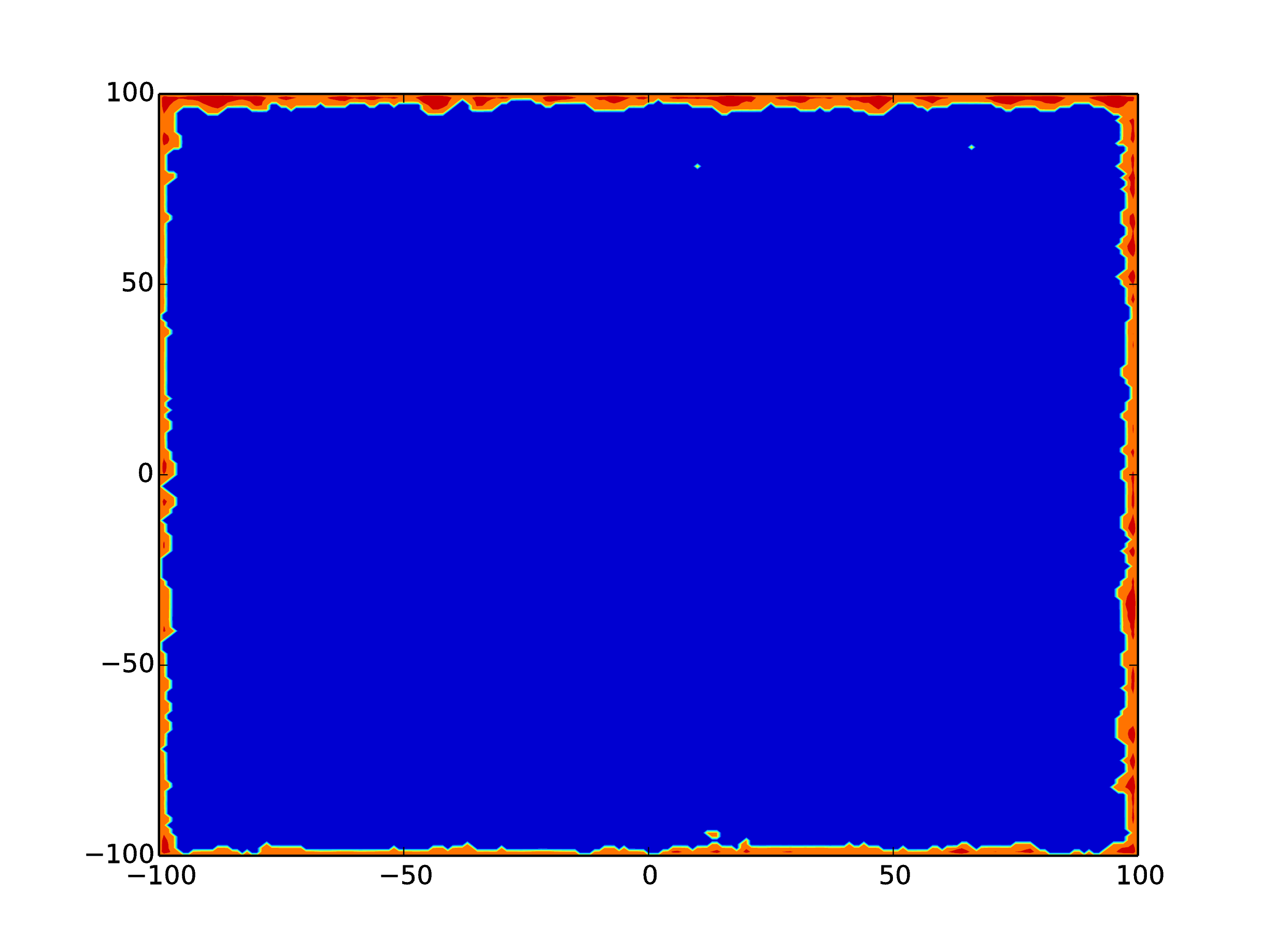}} 
			&
			
			\subfigure[PAO=6\% \label{fig:one obstacle error}]{\includegraphics[width=.18\linewidth, height=.16\linewidth]{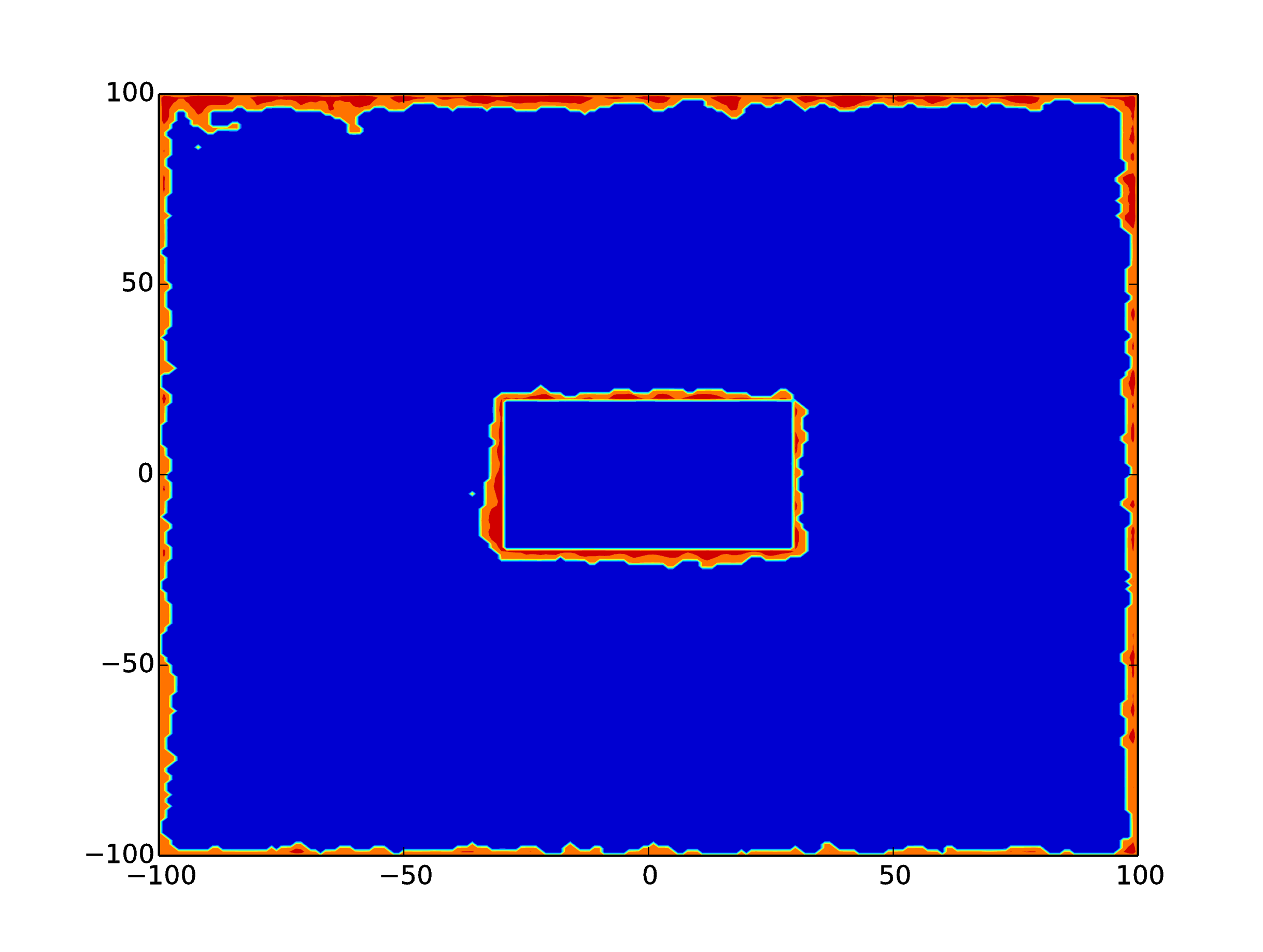}} 
			&

			\subfigure[PAO=9\% \label{fig:two obstacle error}]{\includegraphics[width=.18\linewidth, height=.16\linewidth]{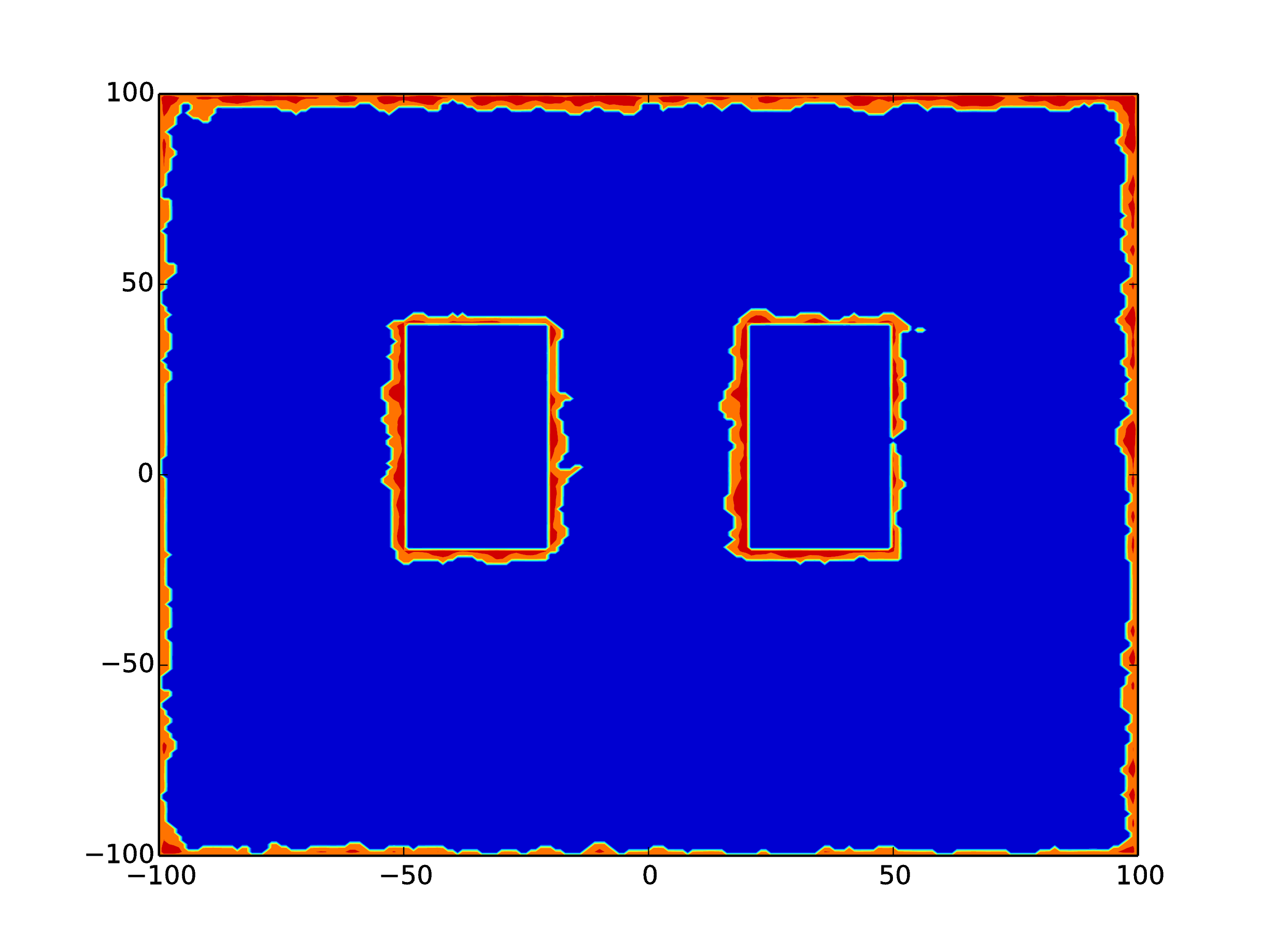}}
			&

			\subfigure[PAO=23.27\% \label{fig:three obstacle error}]{\includegraphics[width=.18\linewidth, height=.16\linewidth]{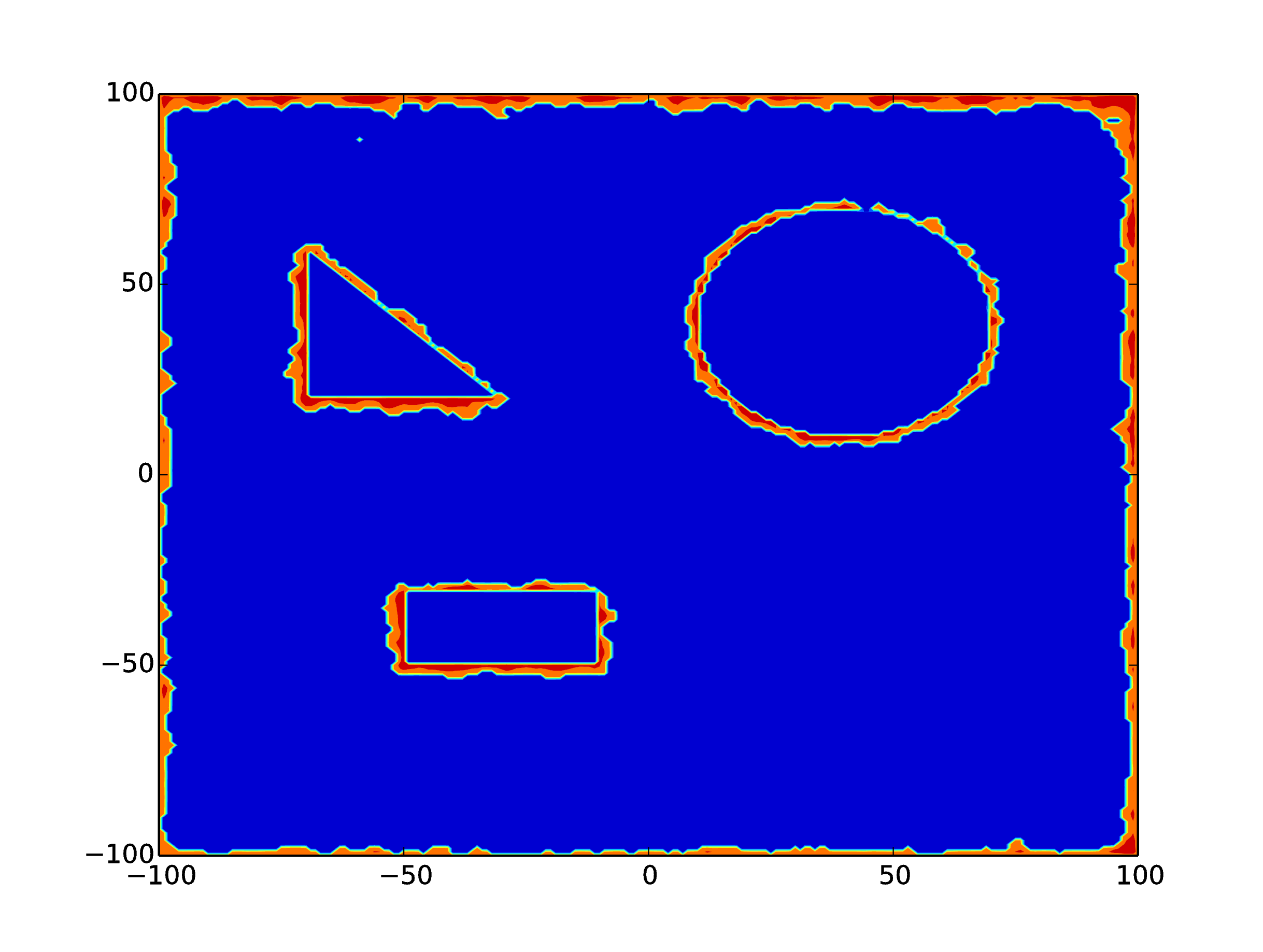}}
			&
			
			\subfigure[PAO=8\% \label{fig:four obstacle error}]{\includegraphics[width=.18\linewidth, height=.16\linewidth]{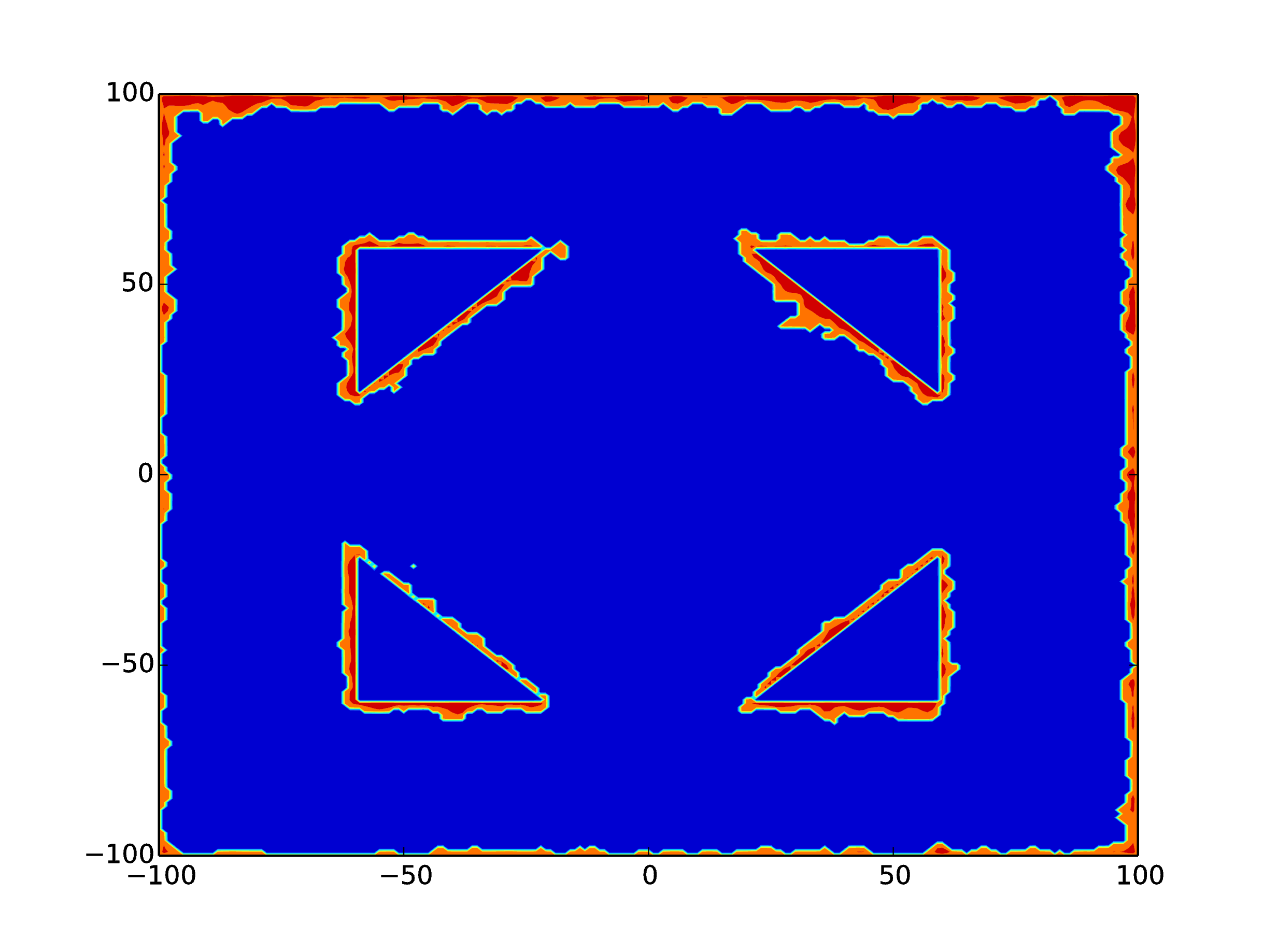}}
			
		\end{tabular}
		\caption{Contour plots of the absolute error between \autoref{fig:worlds} and \autoref{fig:maps}.}  
		\label{fig:error maps}
	\end{figure*}
	

	\begin{figure*}[th]
		
		\begin{tabular}{ccccc}
			\centering	
			\hspace{-3mm}

			\subfigure[Complex domain \label{fig:complex world}]{\includegraphics[width=.18\linewidth, height=.16\linewidth]{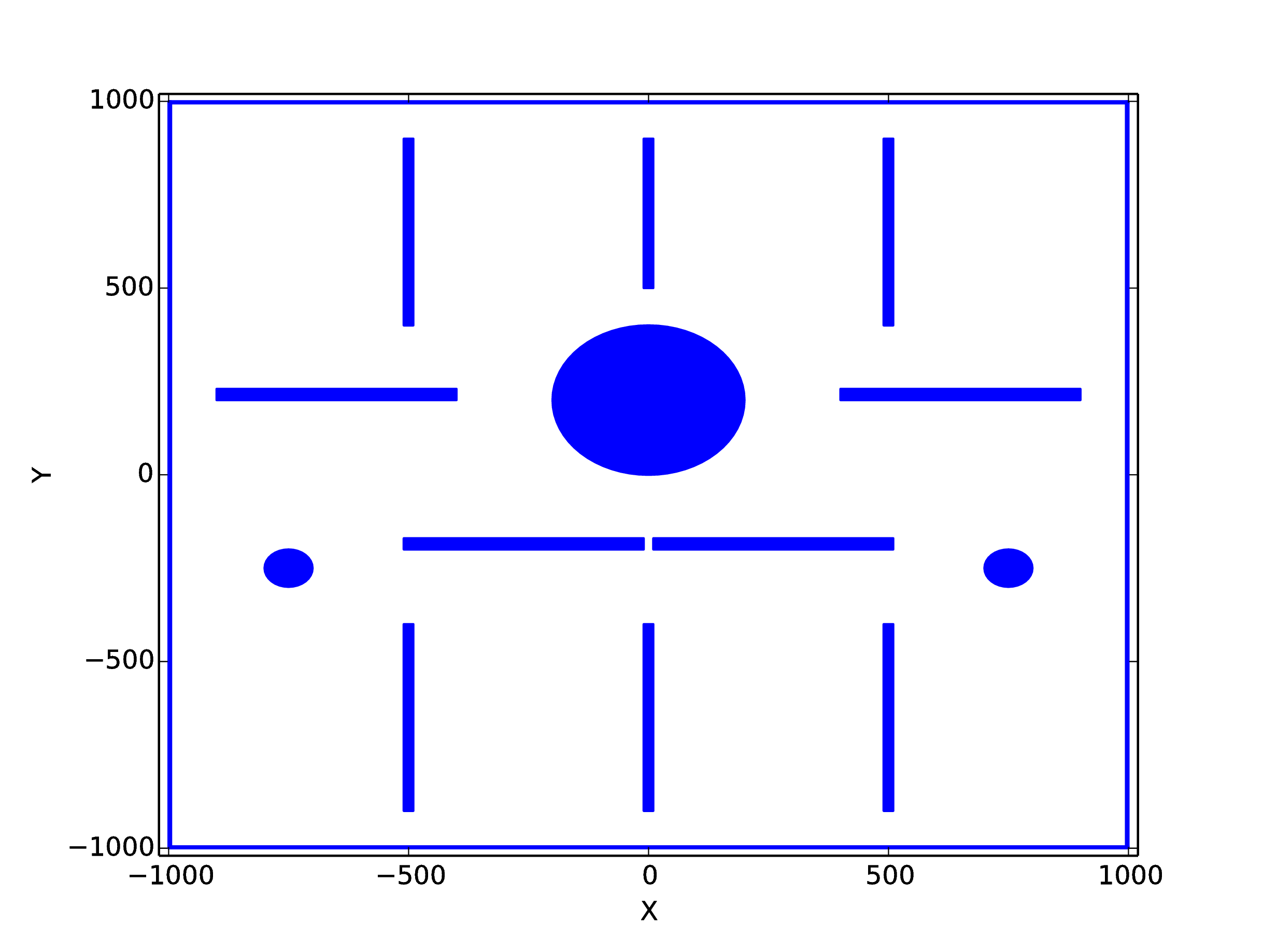}}
			&
			
			\subfigure[Computed $p^f_i$ \label{fig:complex computed}]{\includegraphics[width=.18\linewidth, height=.16\linewidth]{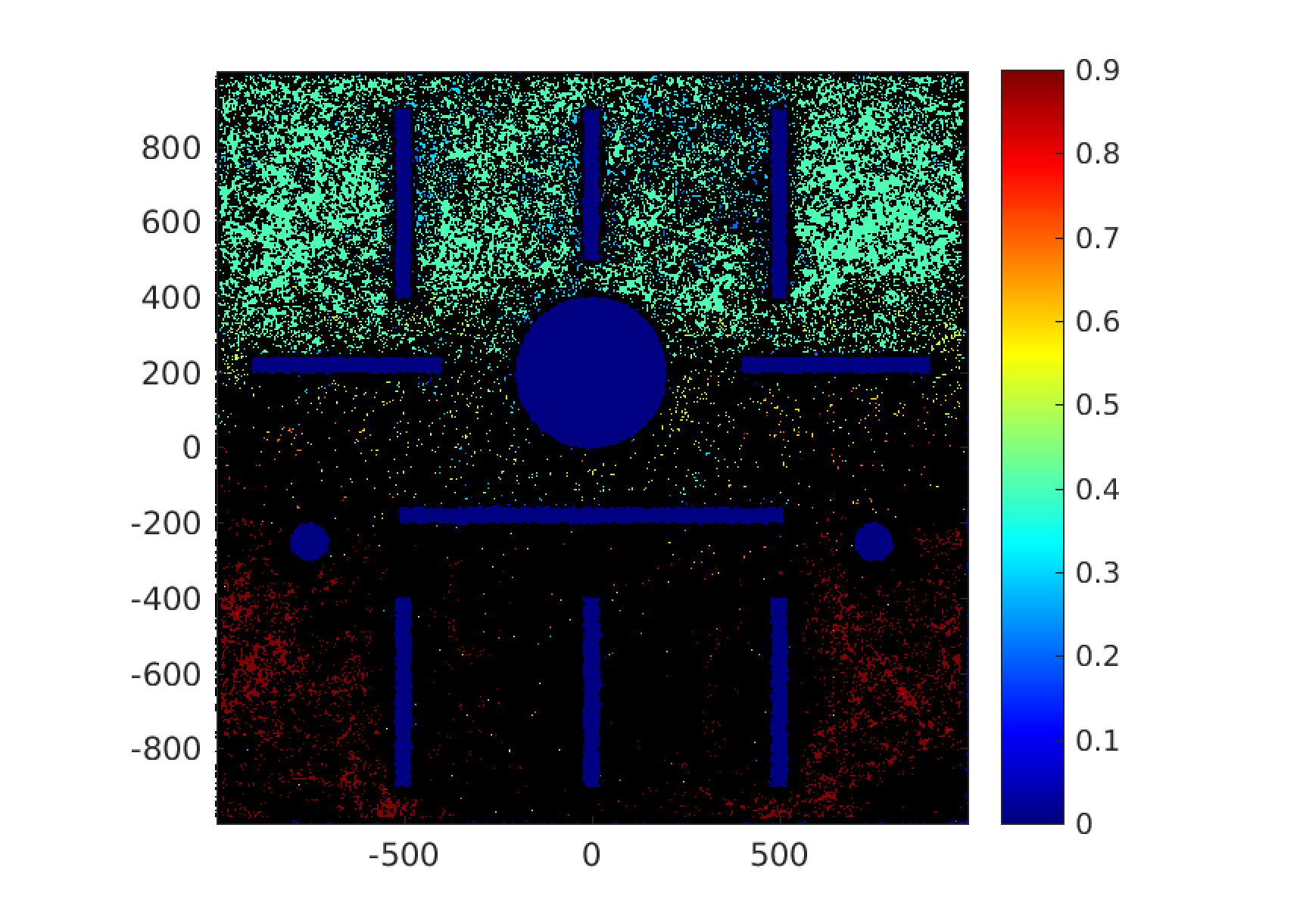}} 
			&
			
			\subfigure[Filtered $p^f_i$ \label{fig:complex filtered}]{\includegraphics[width=.18\linewidth, height=.16\linewidth]{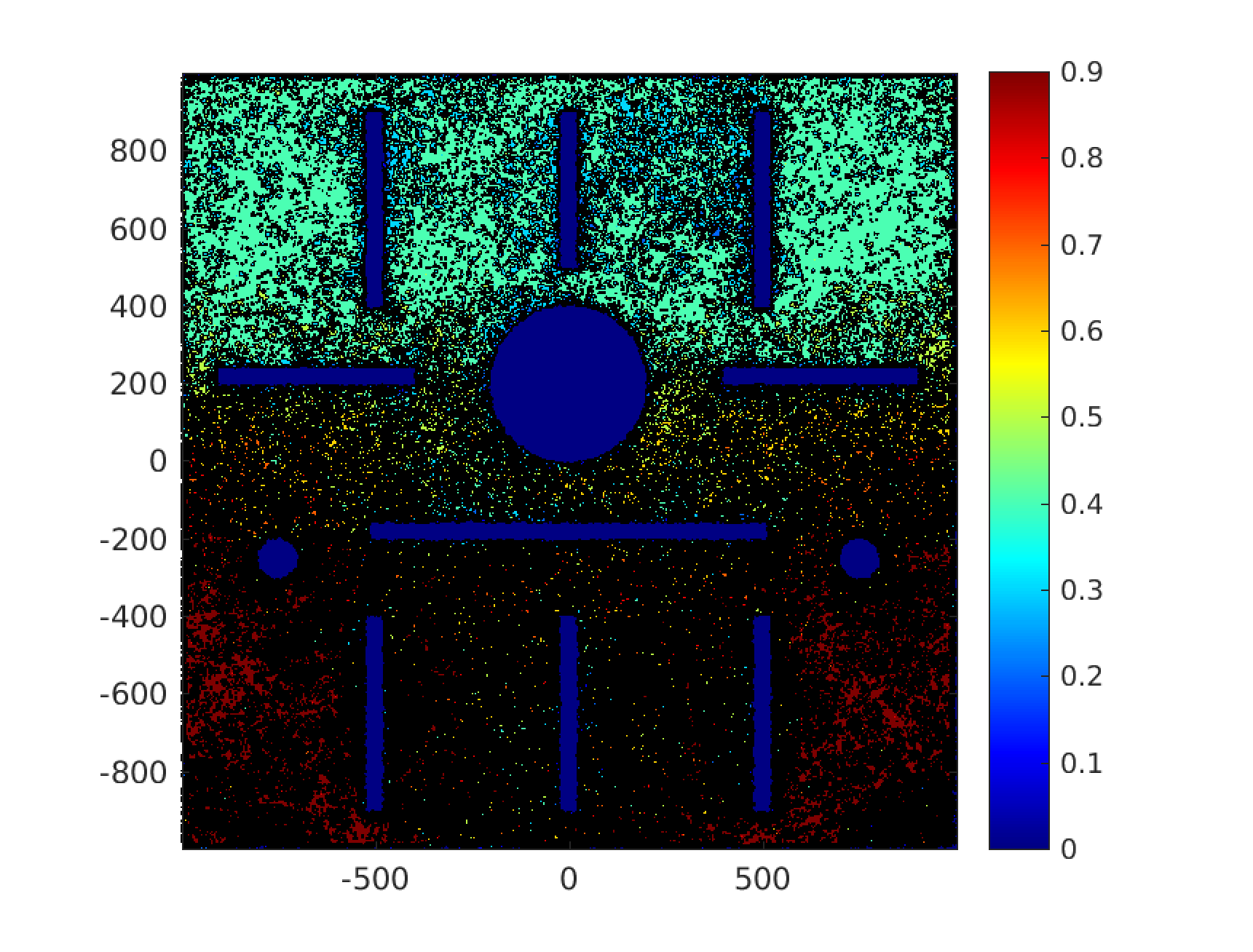}} 
			&		
			
			\subfigure[Map \label{fig:complex map}]{\includegraphics[width=.18\linewidth, height=.16\linewidth]{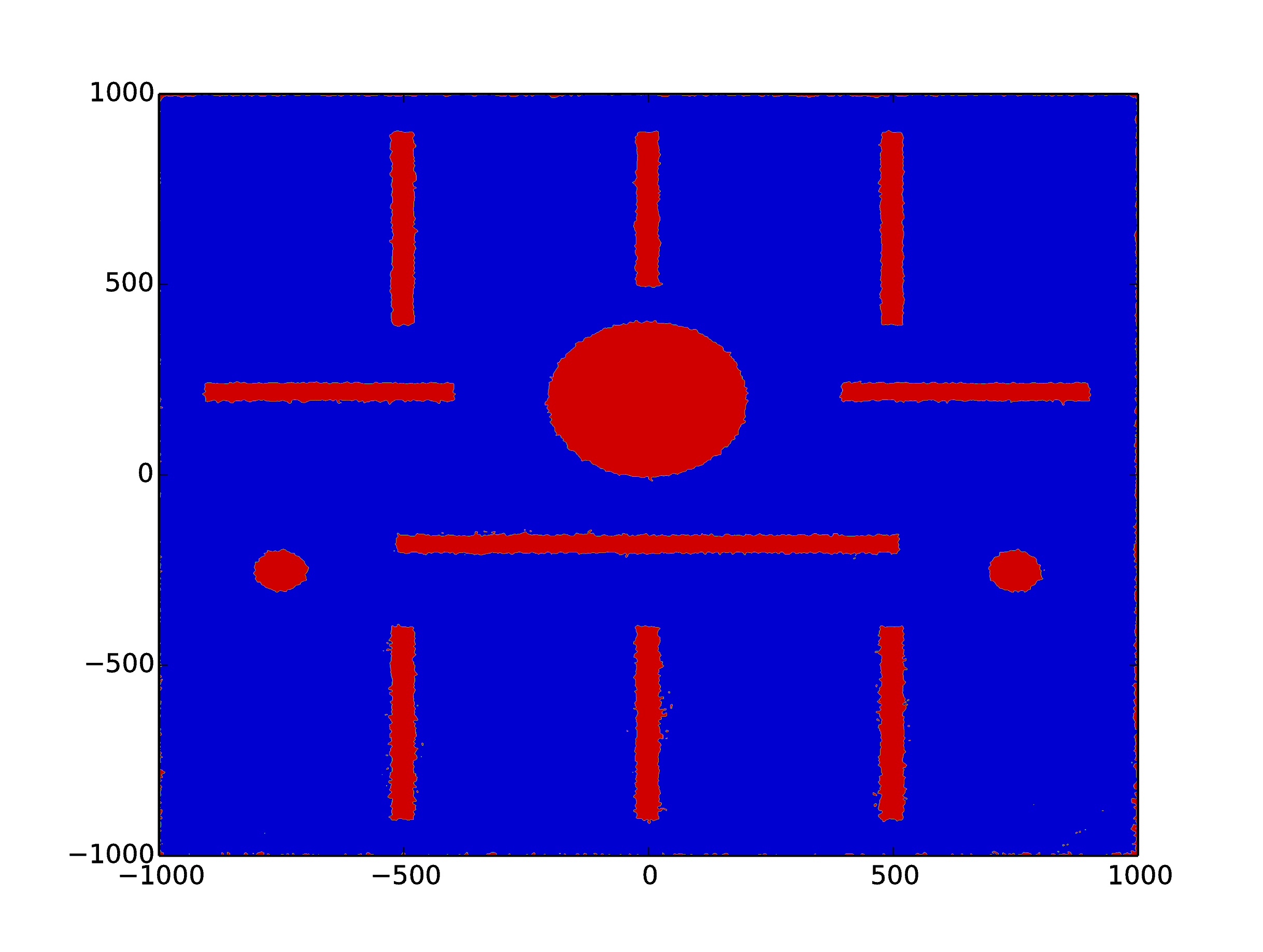}}
			&
			
			\subfigure[Absolute error \label{fig:complex  error}]{\includegraphics[width=.18\linewidth, height=.16\linewidth]{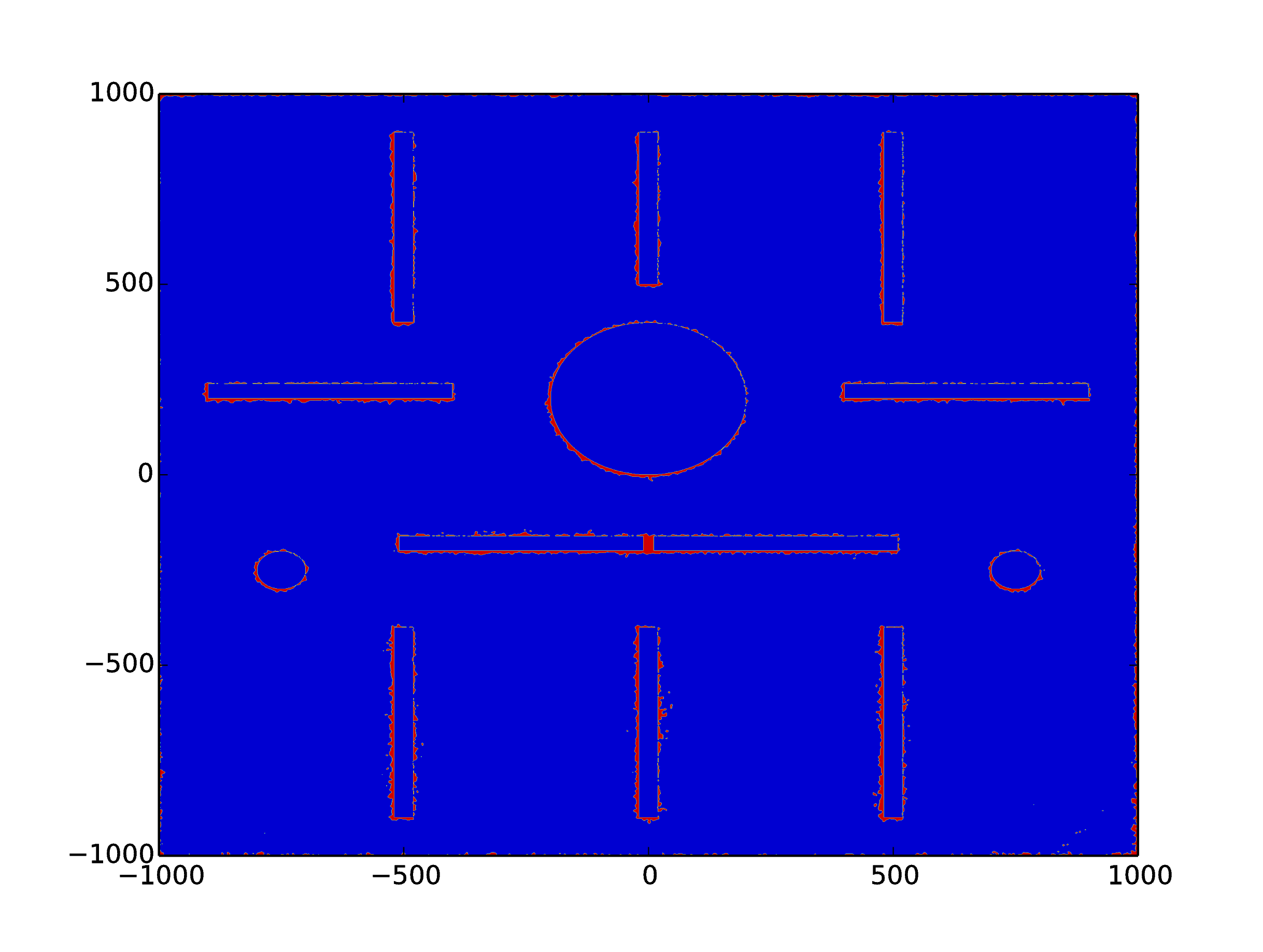}}
			
		\end{tabular}
		\caption{The outputs of the mapping procedure for a complex domain with PAO=5.78\%.}  
		\label{fig:complex results}
	\end{figure*}

	
	\begin{figure*}[th]
		
		\begin{tabular}{ccccc}
			\centering	
			\hspace{-3mm}

			\subfigure[Threshold from all domains combined \label{fig:Threshold combined robot}]{\includegraphics[width=.18\linewidth, height=.16\linewidth]{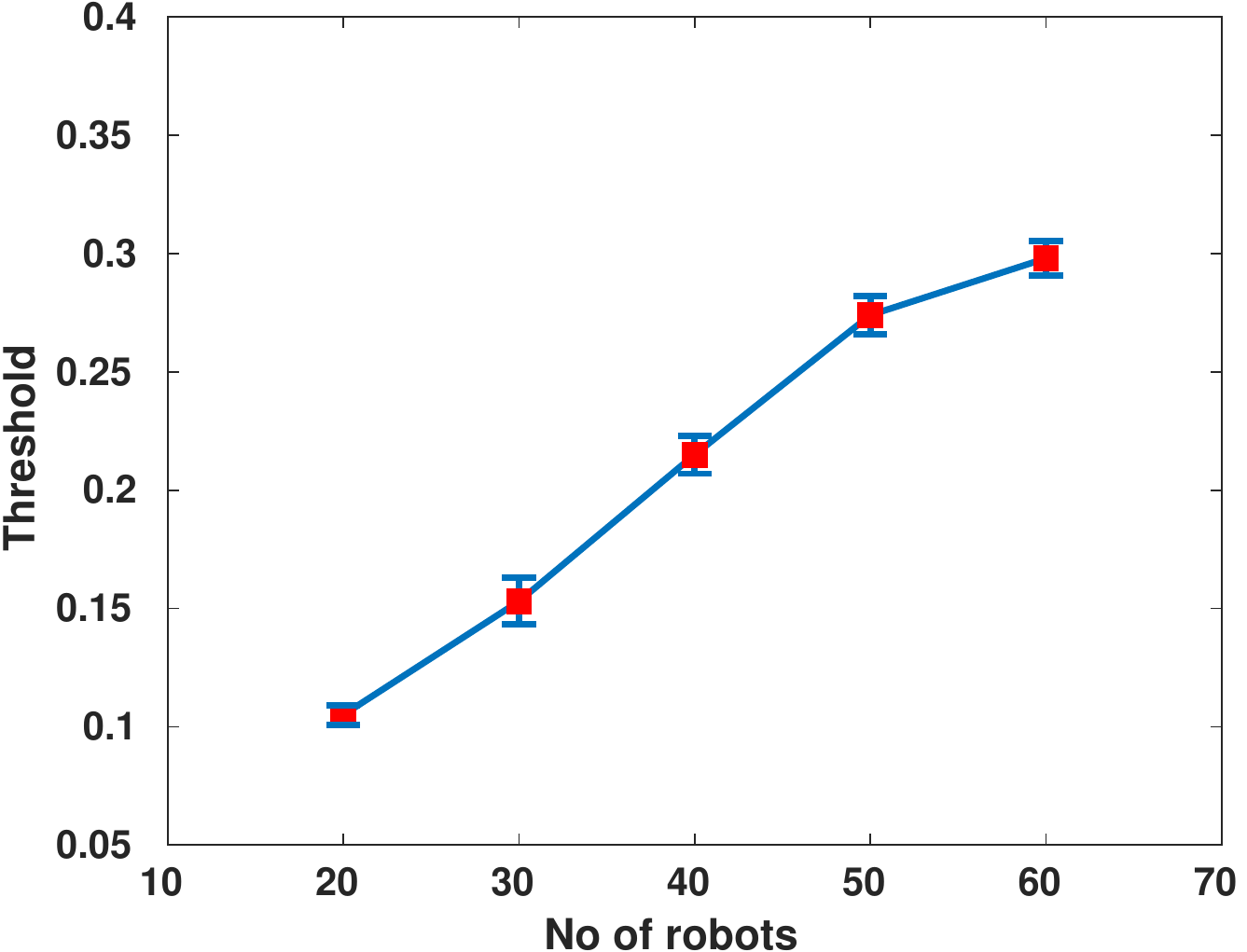}}
			&
			
			\subfigure[Mean threshold for each domain \label{fig:Threshold various domain robot}]{\includegraphics[width=.18\linewidth, height=.16\linewidth]{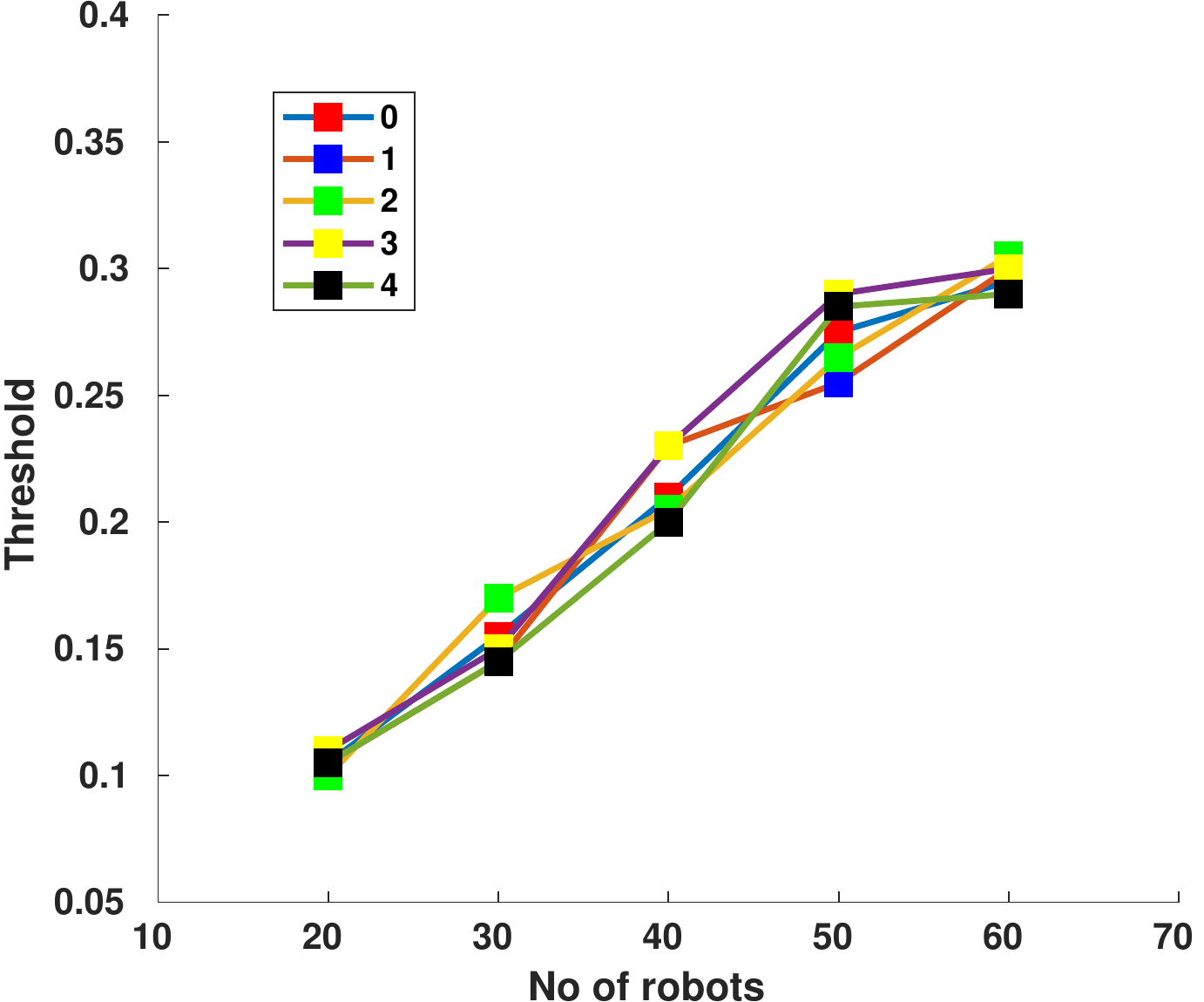}} 
			
			&		
			
			\subfigure[Map estimation error from all domains combined \label{fig:Estimation error combined robots}]{\includegraphics[width=.18\linewidth, height=.16\linewidth]{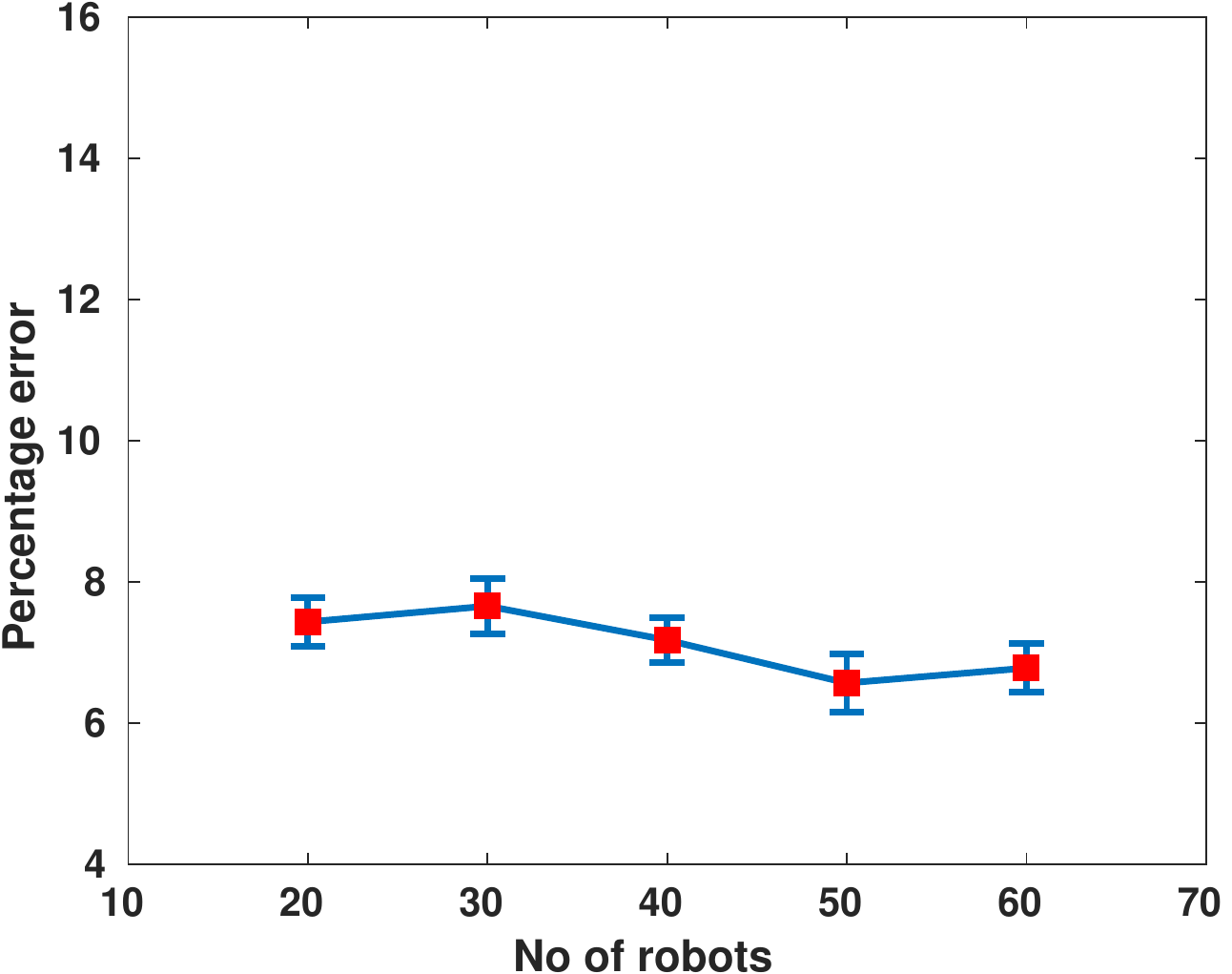}}
			&
			
			\subfigure[Mean map estimation error for each domain  \label{fig: error various domains robot}]{\includegraphics[width=.18\linewidth, height=.16\linewidth]{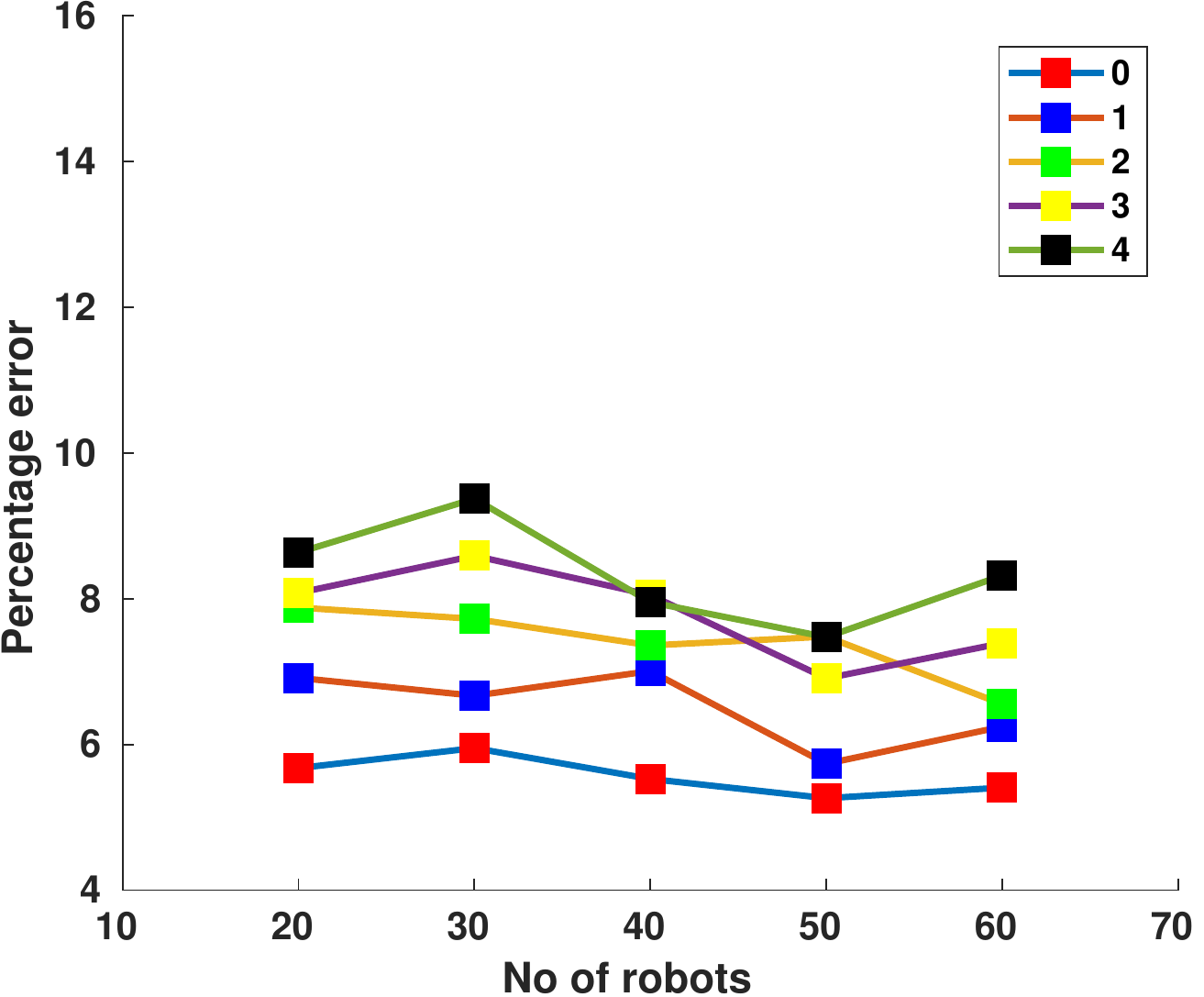}}
			
			&
			
			\subfigure[\% of trials that correctly characterize the domain \label{fig:success robots}]{\includegraphics[width=.18\linewidth, height=.16\linewidth]{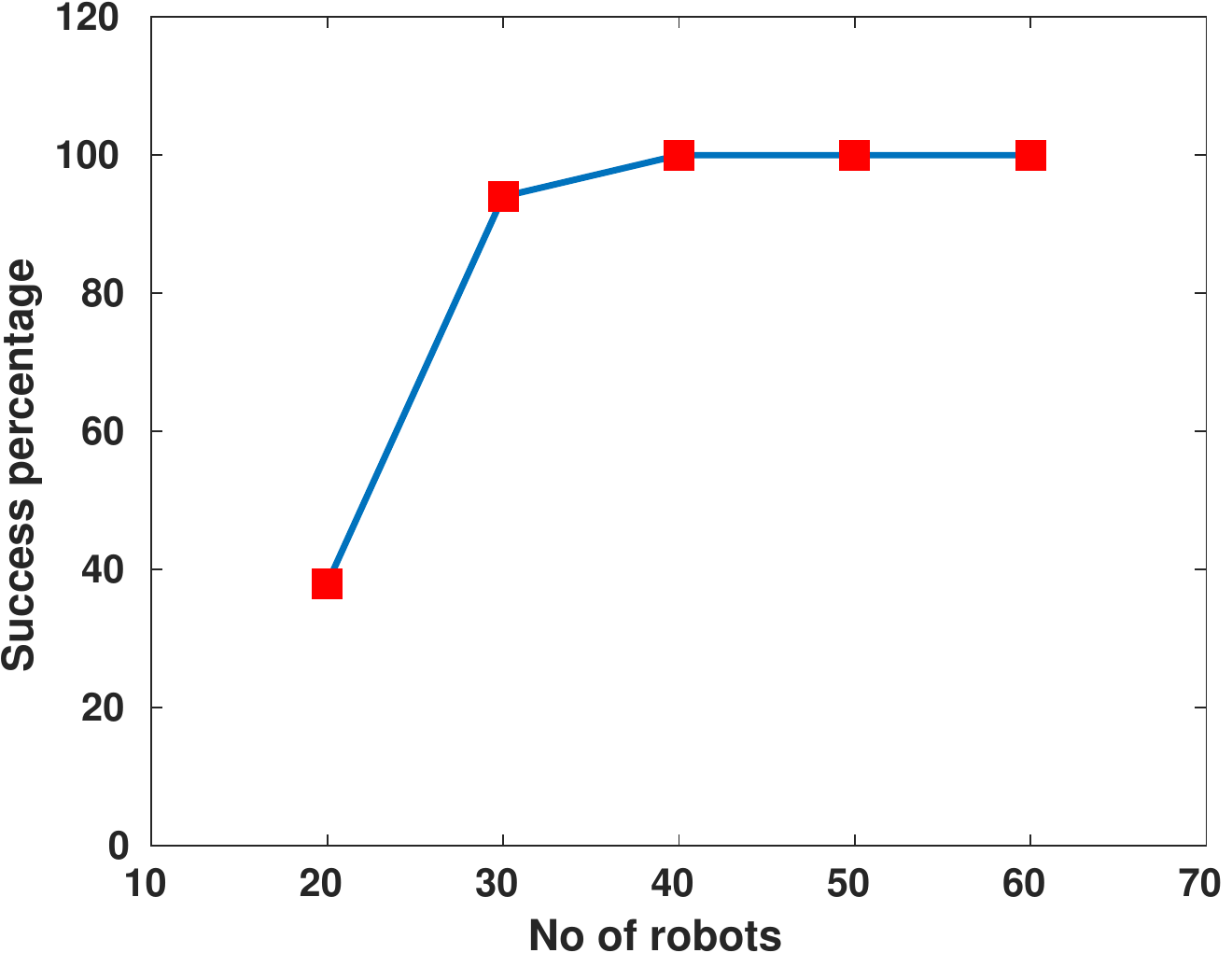}} 
			
		\end{tabular}
		\caption{Plots showing the effect of the number of robots $N$ on the threshold and map estimation error with $T~=~300s$.}  
		\label{fig:varying robots}
	\end{figure*}

	\begin{figure*}[th]
		
		\begin{tabular}{ccccc}
			\centering	
			\hspace{-3mm}

			\subfigure[Threshold from all domains combined \label{fig:Threshold combined time}]{\includegraphics[width=.18\linewidth, height=.16\linewidth]{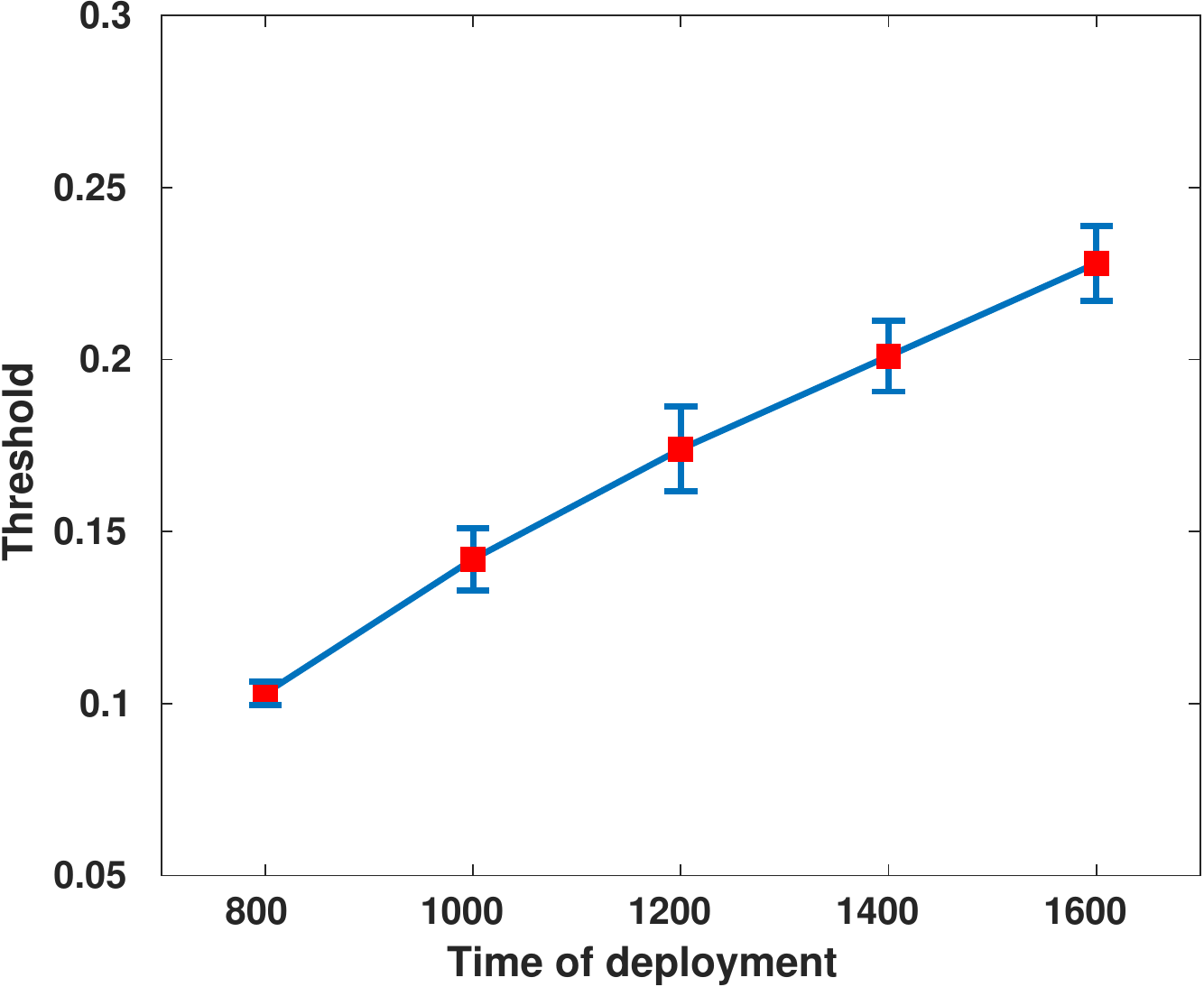}}
			&
			
			\subfigure[Mean threshold for each domain \label{fig:Threshold various domain time}]{\includegraphics[width=.18\linewidth, height=.16\linewidth]{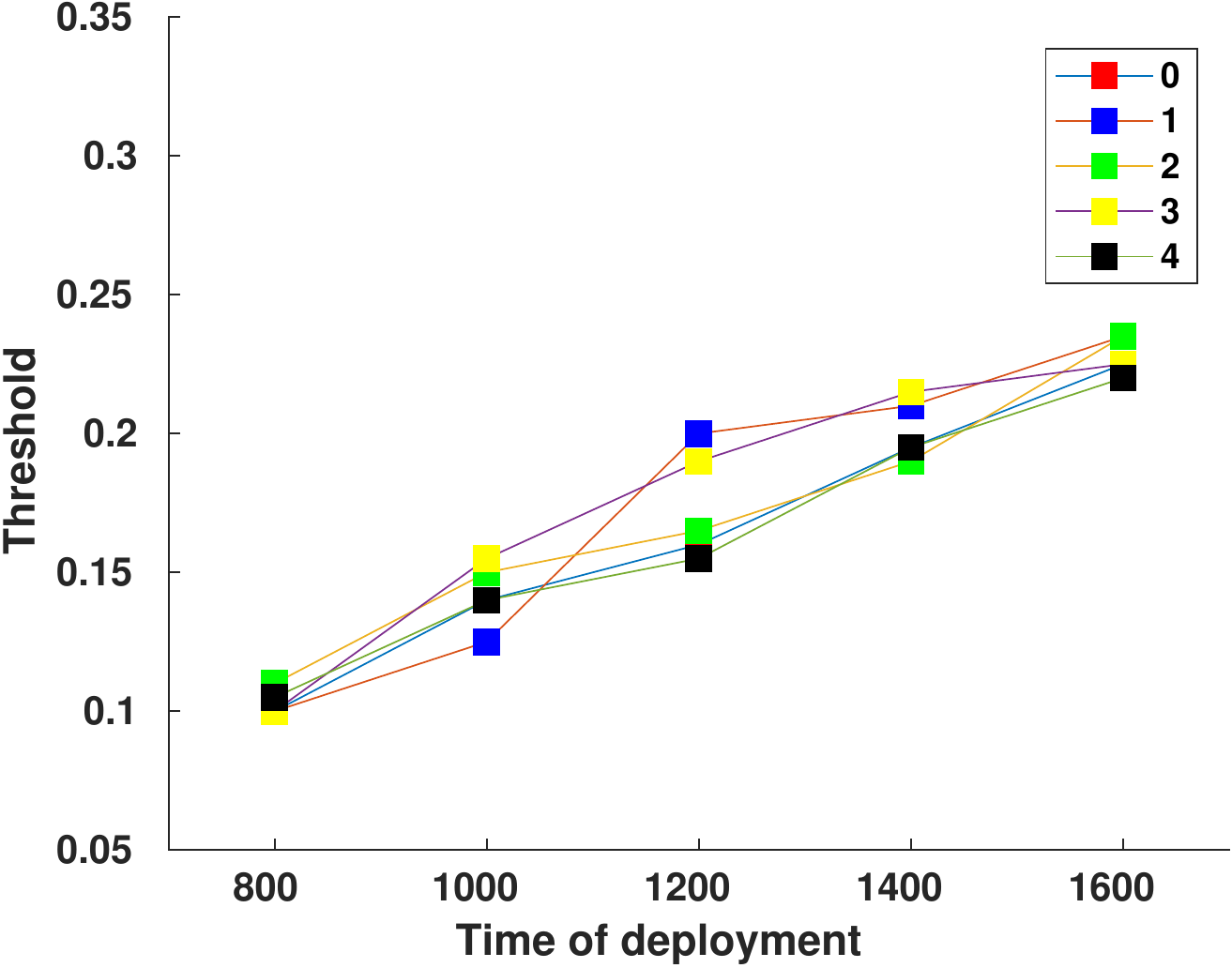}} 
			&		
			
			\subfigure[Map estimation error from all domains combined \label{fig:Estimation error combined time}]{\includegraphics[width=.18\linewidth, height=.16\linewidth]{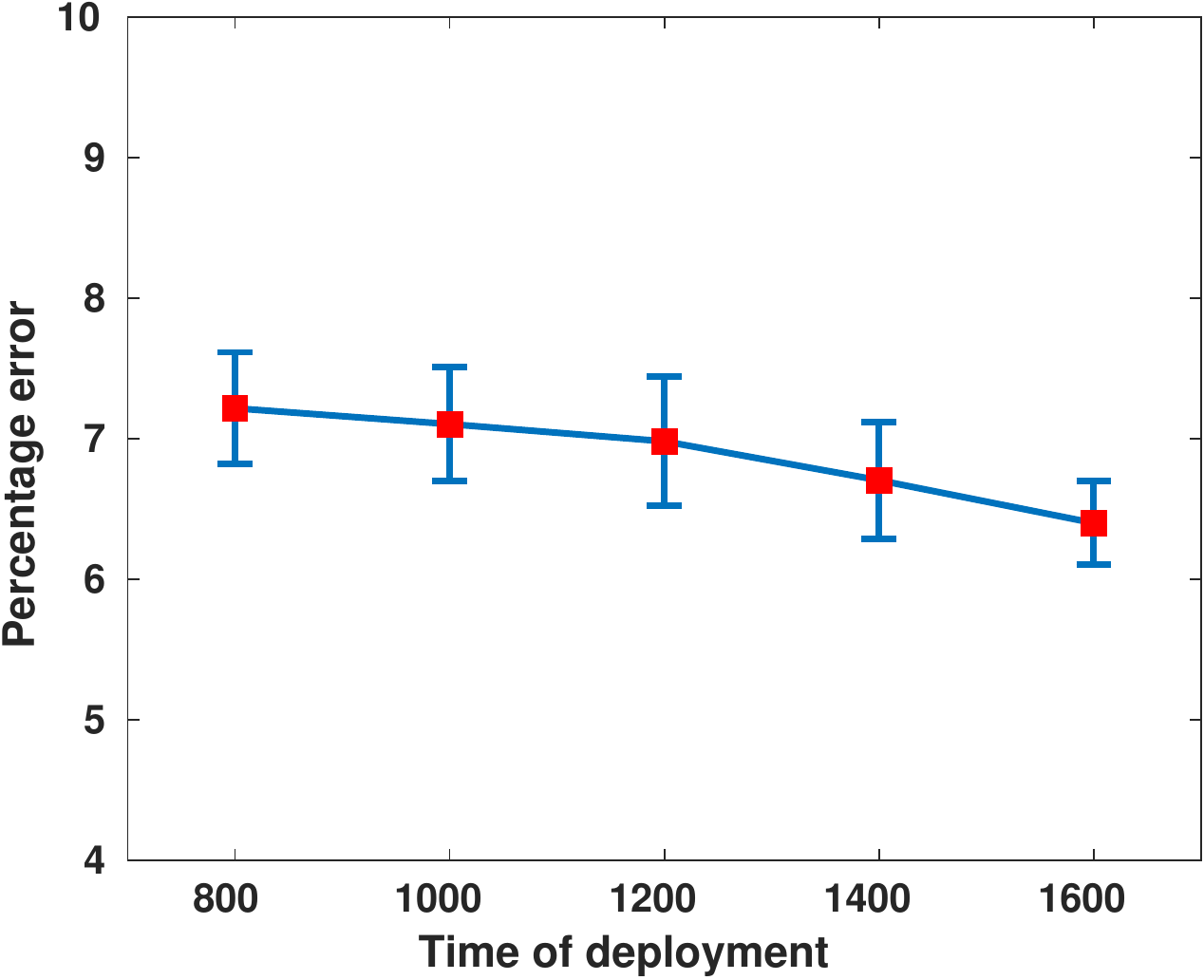}}
			&
			
			\subfigure[Mean map estimation error for each domain  \label{fig: error various domains time}]{\includegraphics[width=.18\linewidth, height=.16\linewidth]{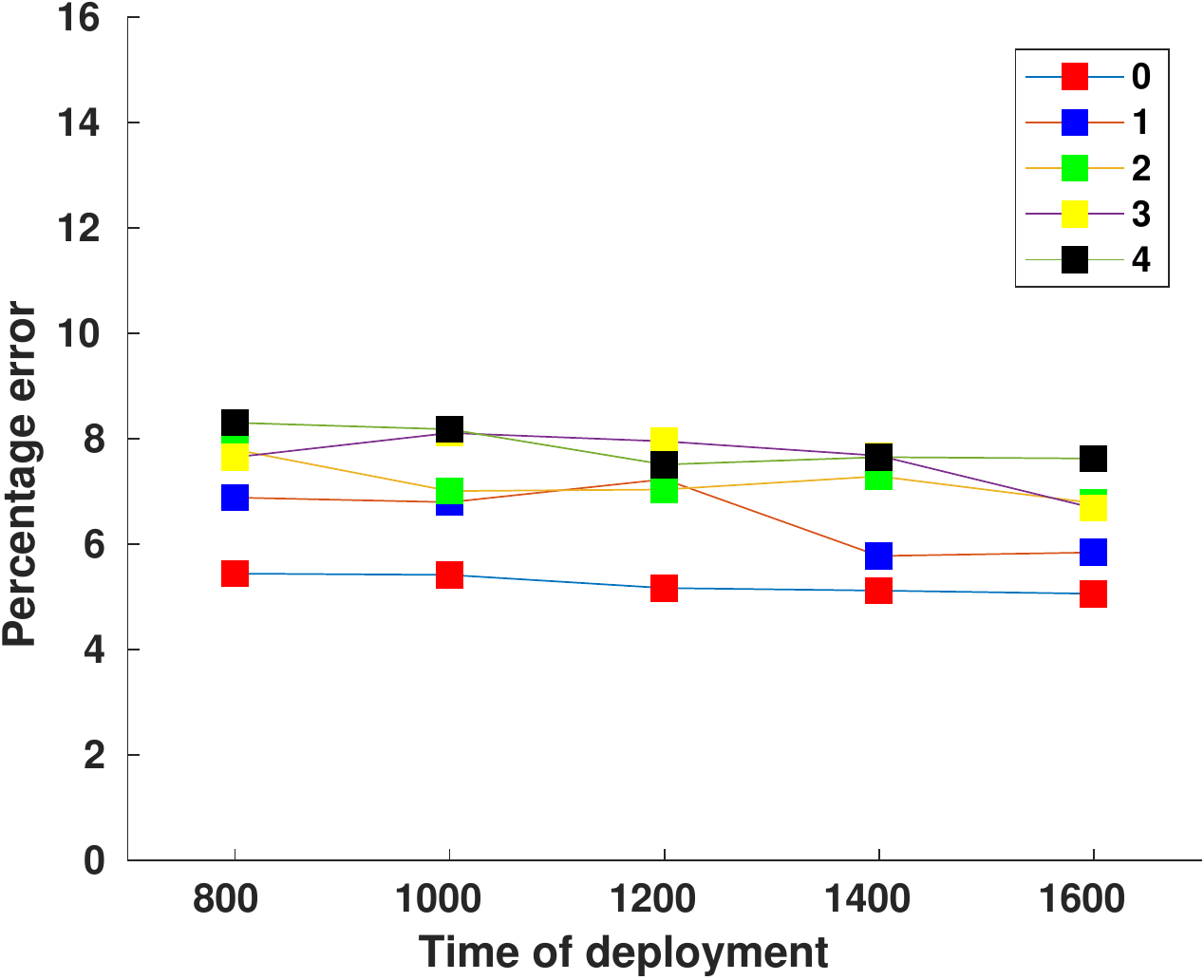}}
			
			&
			
			\subfigure[\% of trials that correctly characterize the domain  \label{fig:success time}]{\includegraphics[width=.18\linewidth, height=.16\linewidth]{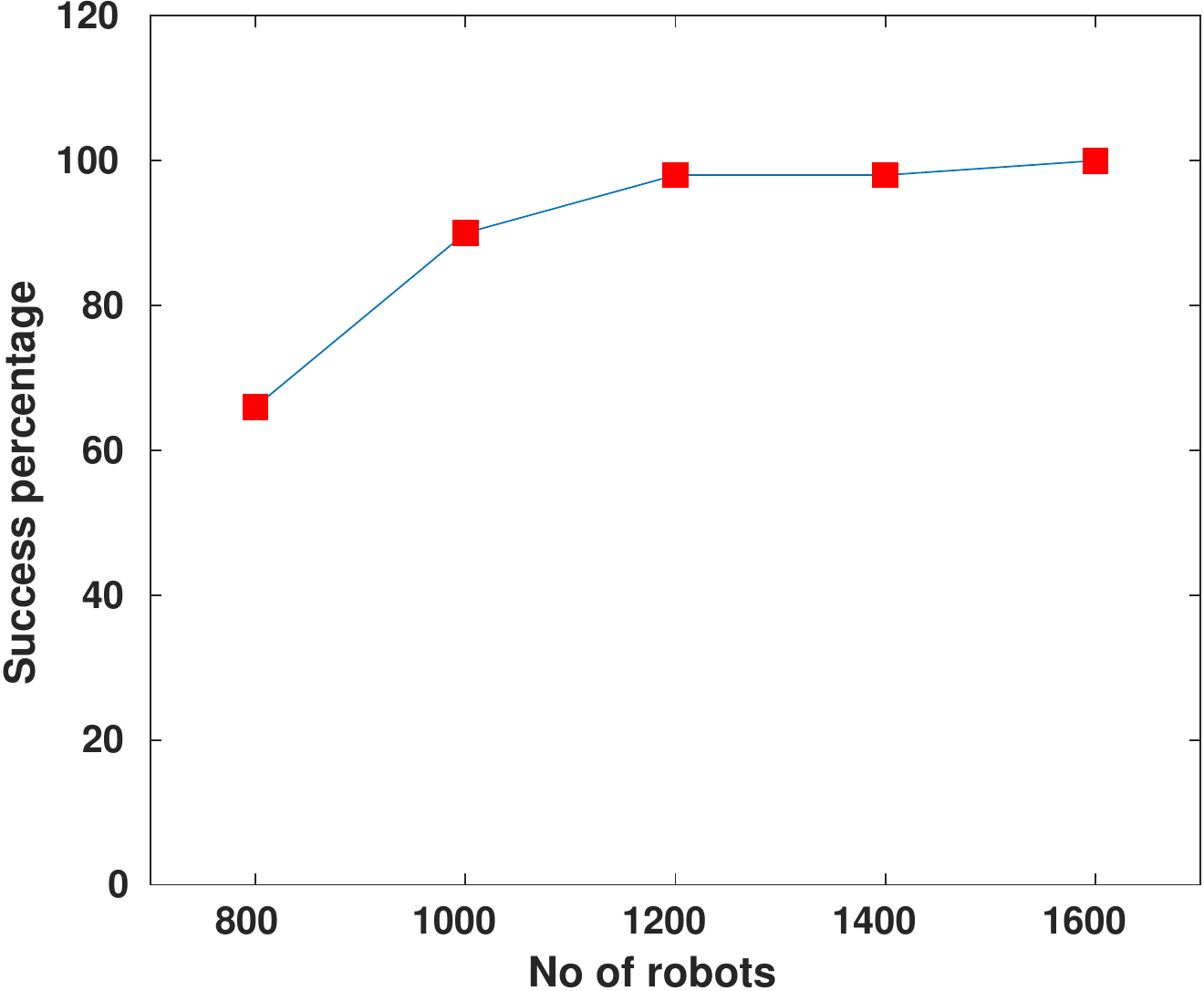}} 
			
		\end{tabular}
		\caption{Plots showing the effect of the deployment time $T$ on the threshold and map estimation error.}  
		\label{fig:varying time}
	\end{figure*}
	

	
	\begin{figure}[th]
		\begin{tabular}{cc}
			\centering	
			\hspace{-3mm}

			\subfigure[Threshold for various $N$ \label{fig:Threshold combined robot low}]{\includegraphics[width=.48\linewidth, height=.4\linewidth]{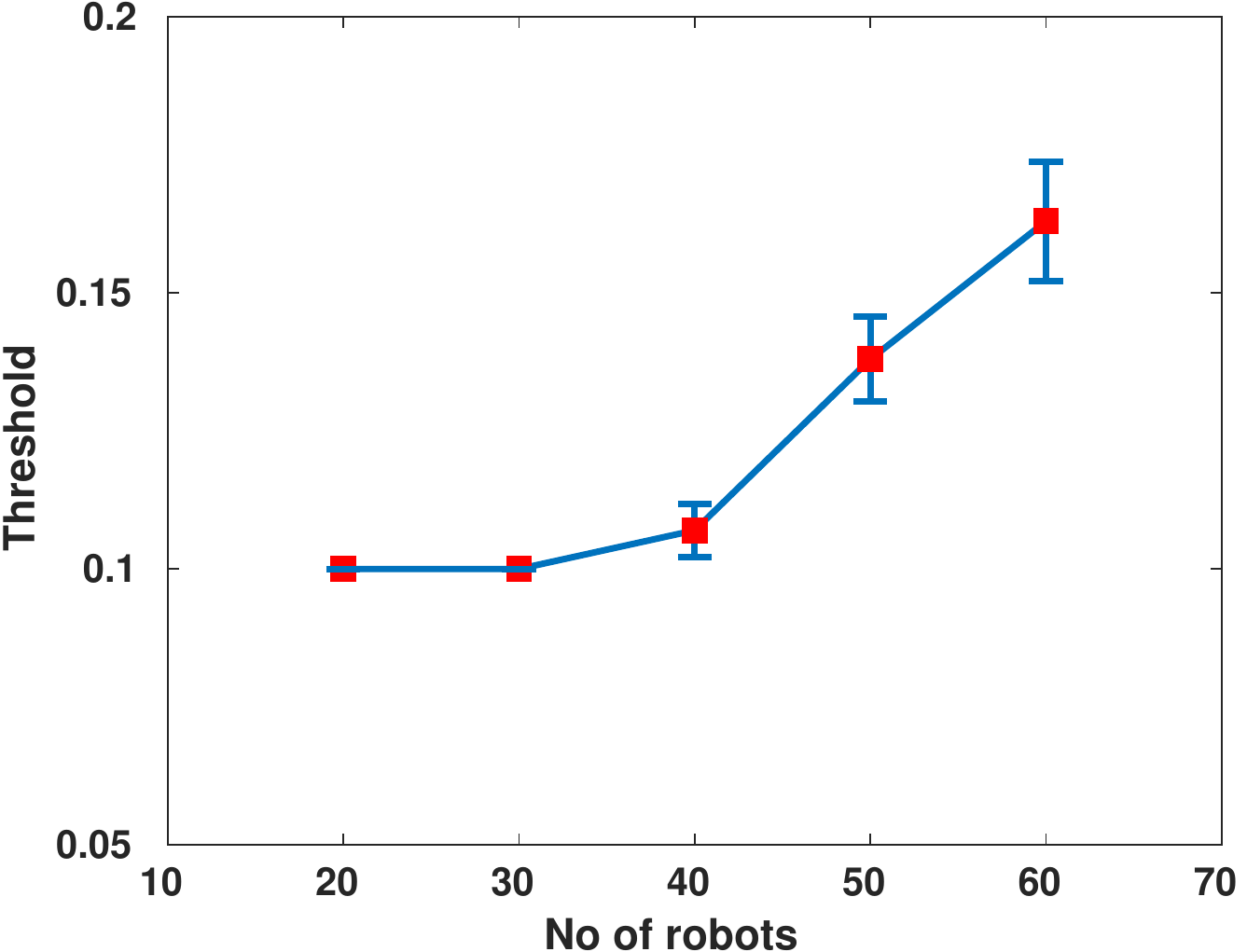}}
			&

			\subfigure[Estimation error for various $N$ \label{fig:Estimation error combined robots low}]{\includegraphics[width=.48\linewidth, height=.4\linewidth]{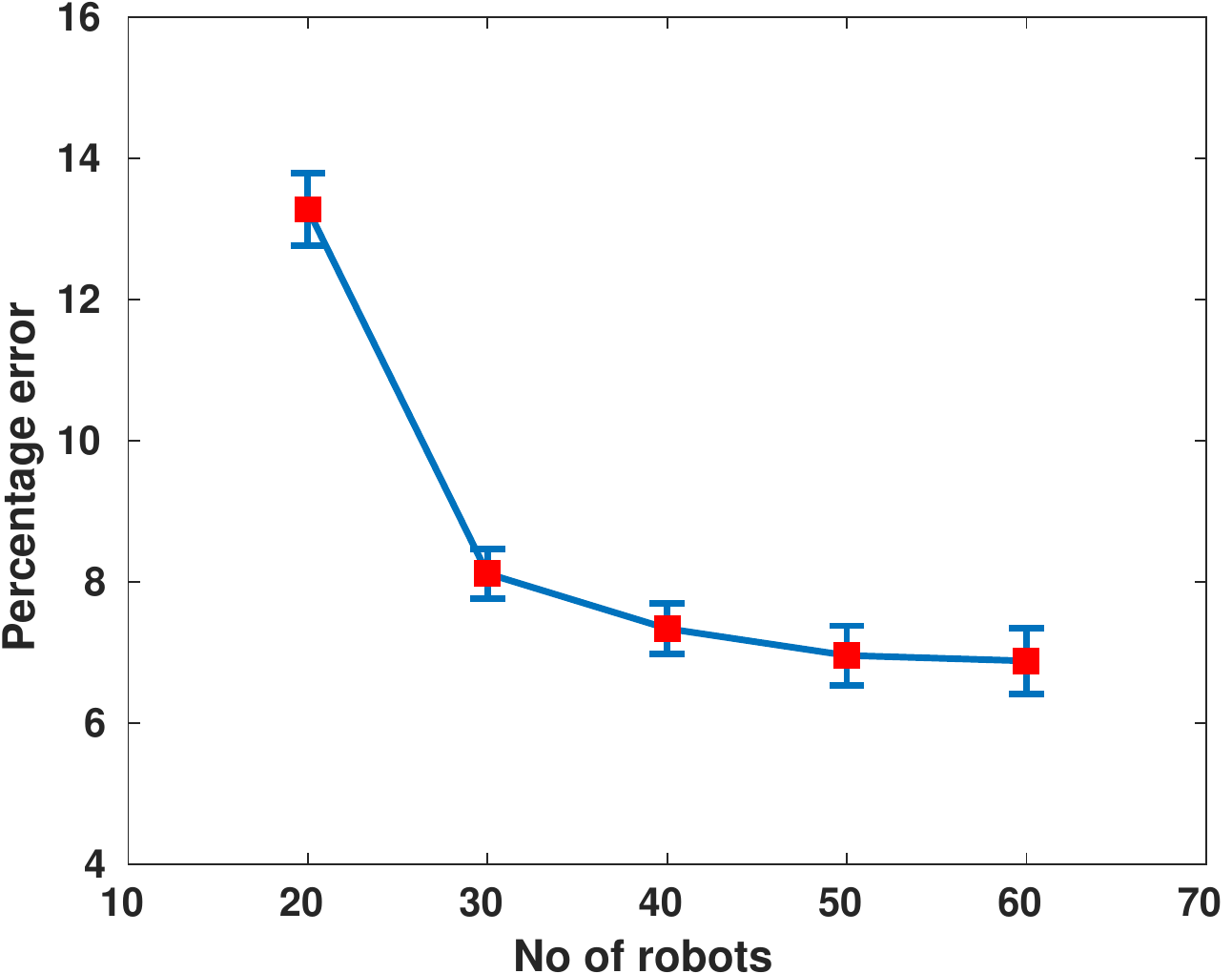}}

		\end{tabular}
		\caption{Plots showing the effect of the number of robots $N$ on the threshold and map estimation error with $T=160s$.}  
		\label{fig:varying robots low}
	\end{figure}
	

	\section{SIMULATION RESULTS}  
	\label{sec:simulation results}

	\begin{figure}[th]
		
		\begin{tabular}{cc}
			\centering	
			\hspace{-3mm}

			\subfigure[Threshold \label{fig:threshold 20 trails}]{\includegraphics[width=.48\linewidth, height=.4\linewidth]{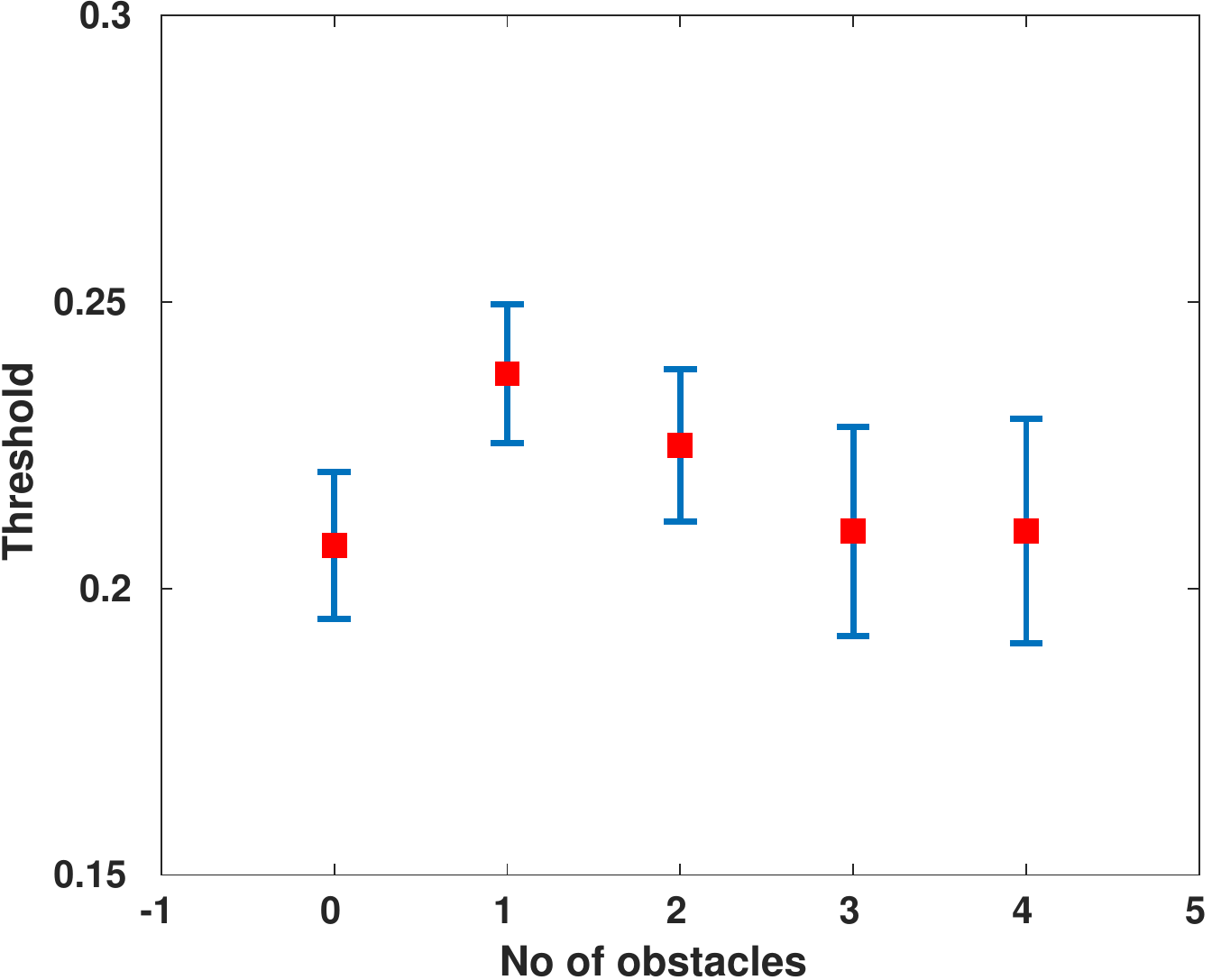}}
			&
			\subfigure[Map estimation error \label{fig:error 20 trails}]{\includegraphics[width=.48\linewidth, height=.4\linewidth]{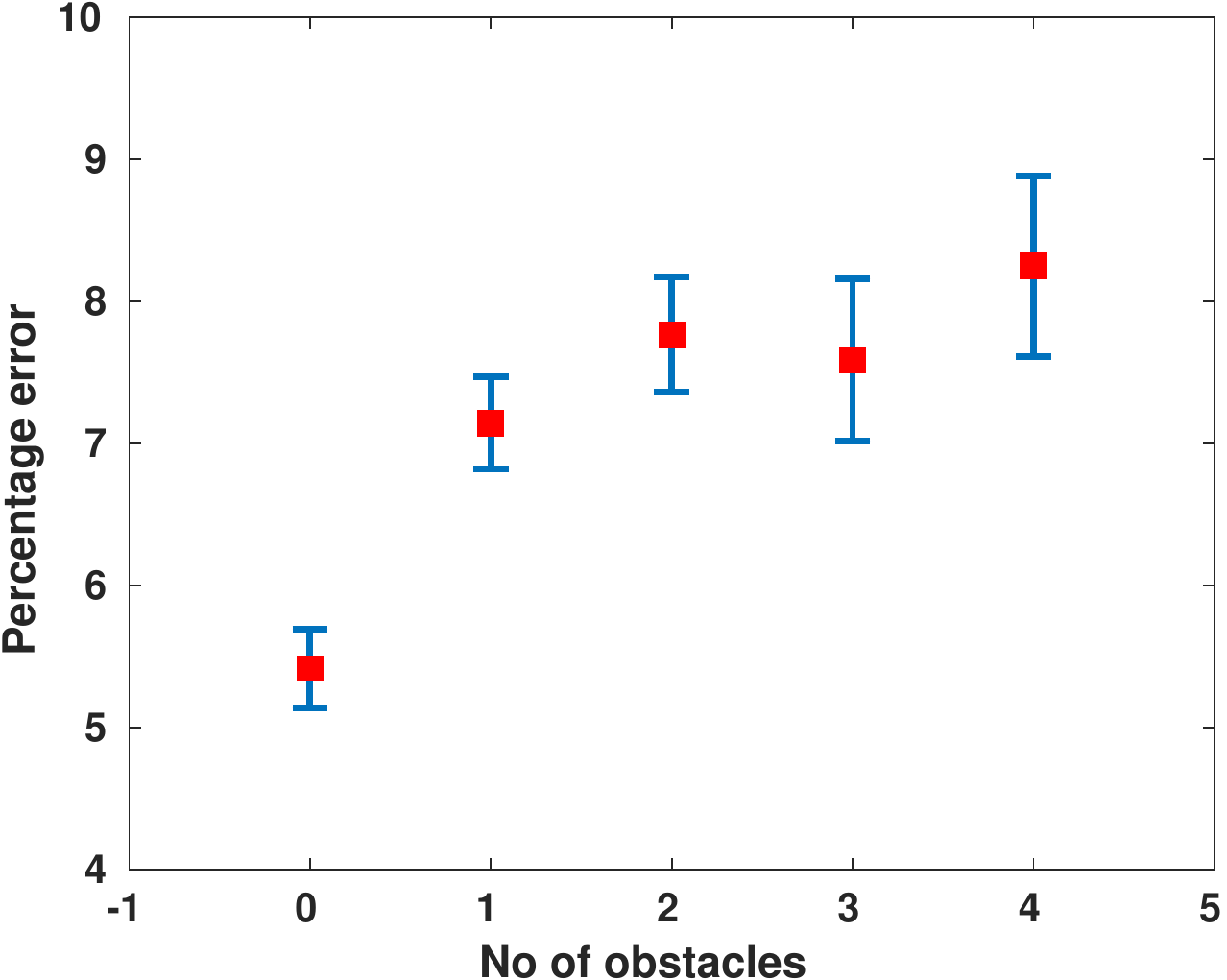}} 
			
		\end{tabular}
		\caption{Results from 20 simulations on each domain in \autoref{fig:worlds}.}  
		\label{fig:validation on domain}
	\end{figure}
	
	\begin{figure}[th]
		
		\begin{tabular}{cc}
			\centering	
			\hspace{-3mm}

			\subfigure[Topological map from \cite{Ramachandran_Wilson_Berman_RAL_17} \label{fig:complex top earlier}]{\includegraphics[width=.48\linewidth, height=.4\linewidth]{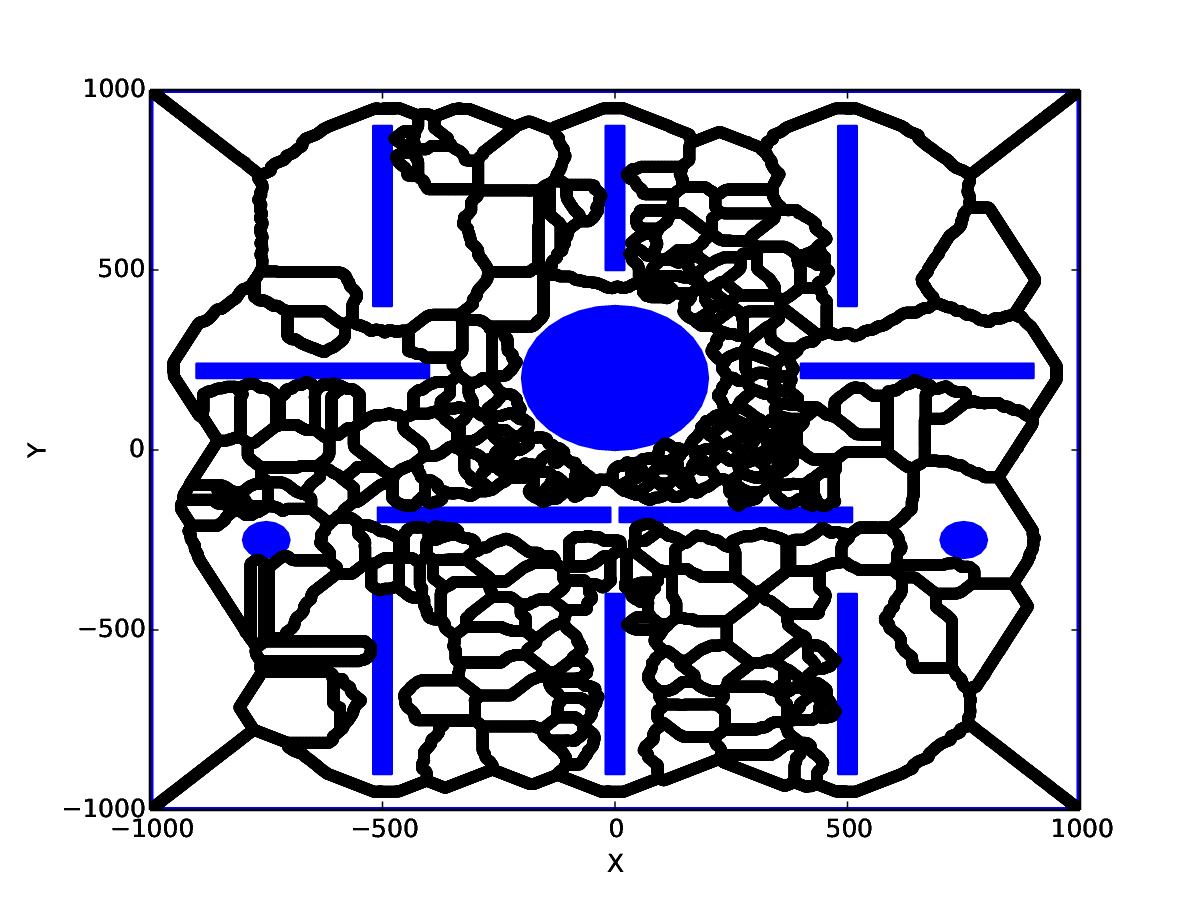}}
			&
			
			\subfigure[Topological map using current approach \label{fig:complex topological now}]{\includegraphics[width=.48\linewidth, height=.4\linewidth]{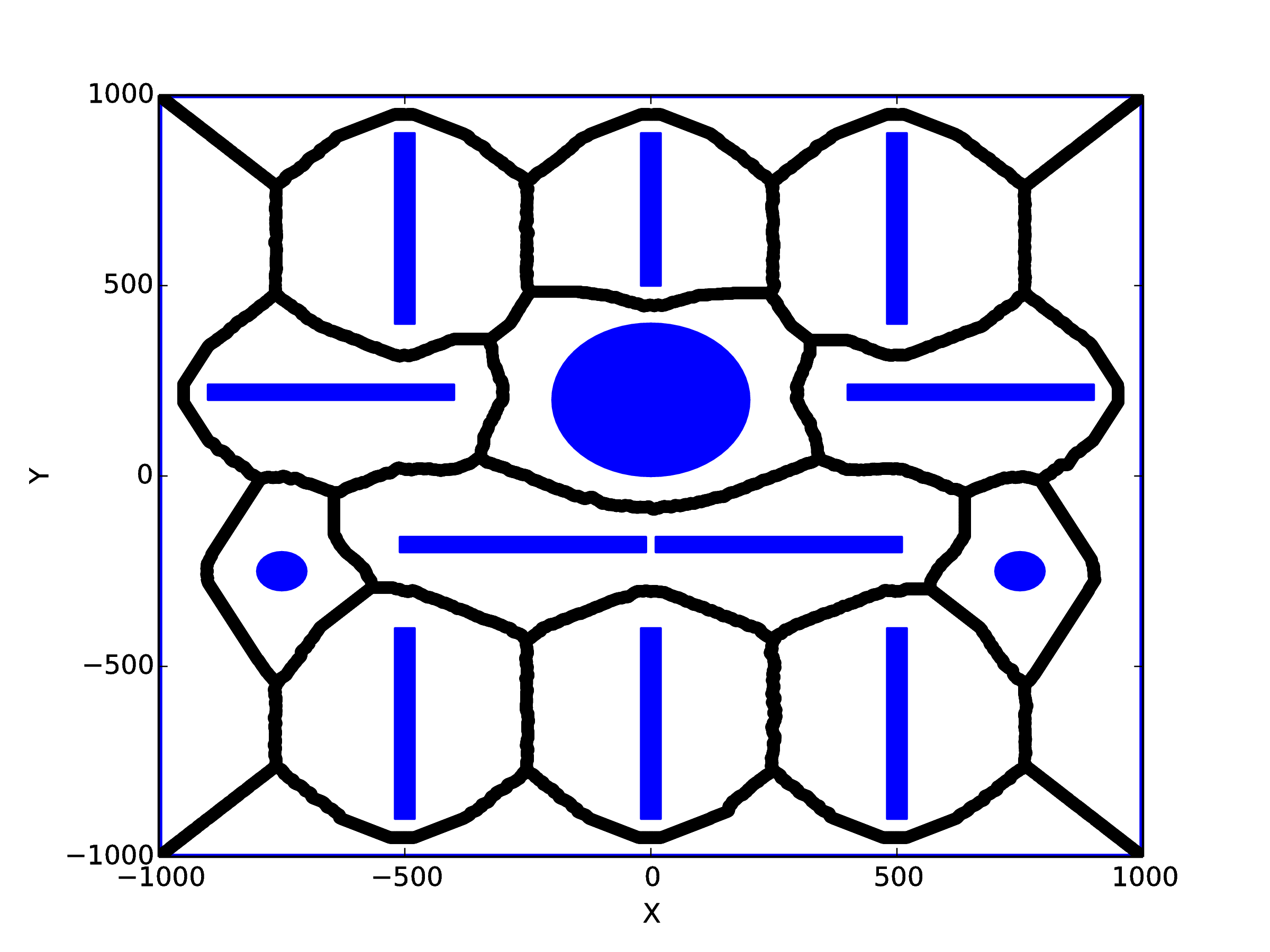}} 		
			
		\end{tabular}
		\caption{Topological maps generated for a complex domain.}  
		\label{fig:complex topological results}
	\end{figure}
	
	\begin{figure}[th]
		
		\begin{tabular}{cc}
			\centering	
			\hspace{-3mm}

			\subfigure[Actual domain\label{fig:domain 5}]{\includegraphics[width=.48\linewidth, height=.4\linewidth]{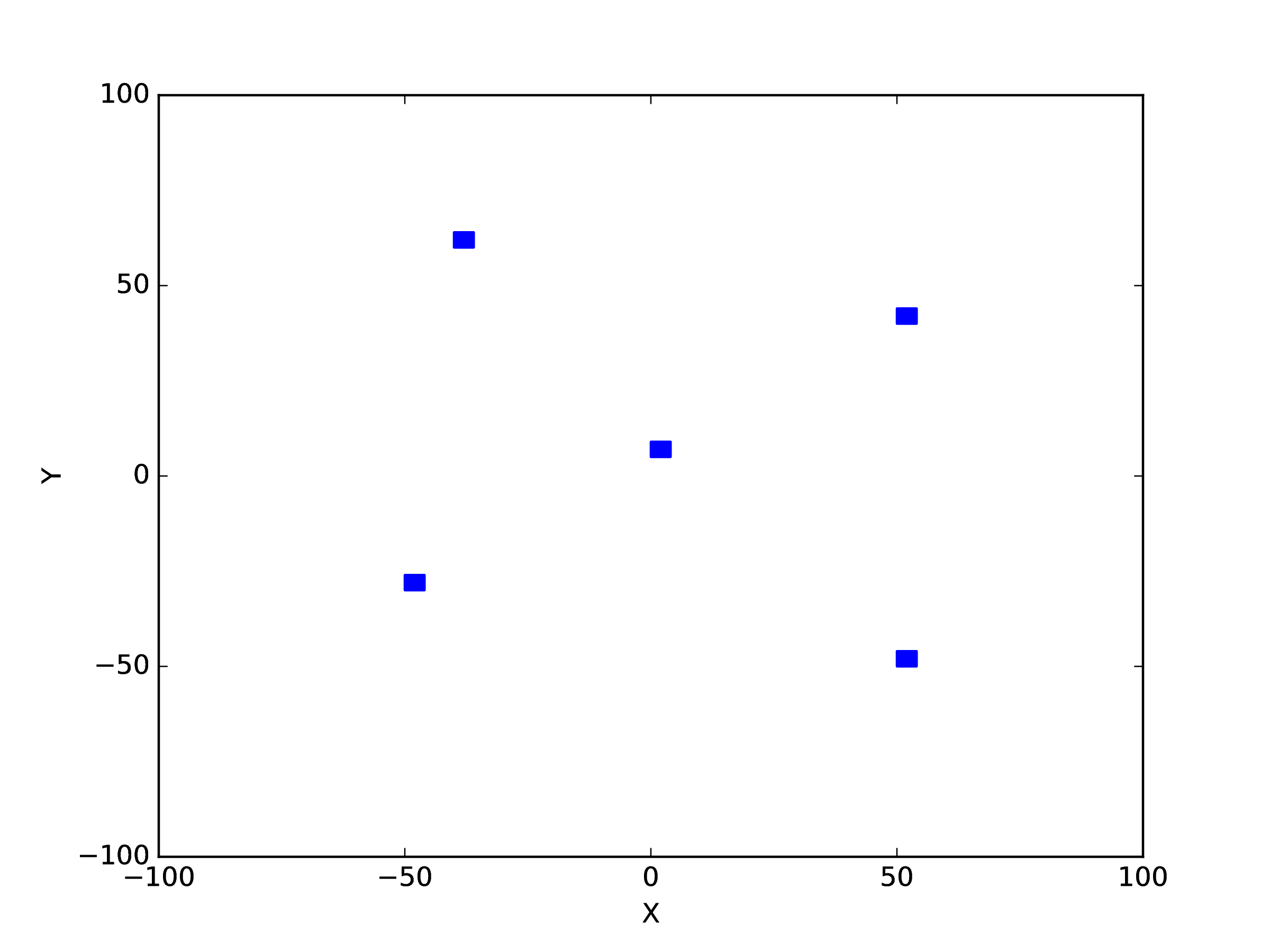}}
			&
			
			\subfigure[Absolute error\label{fig:absolute error 5}]{\includegraphics[width=.48\linewidth, height=.4\linewidth]{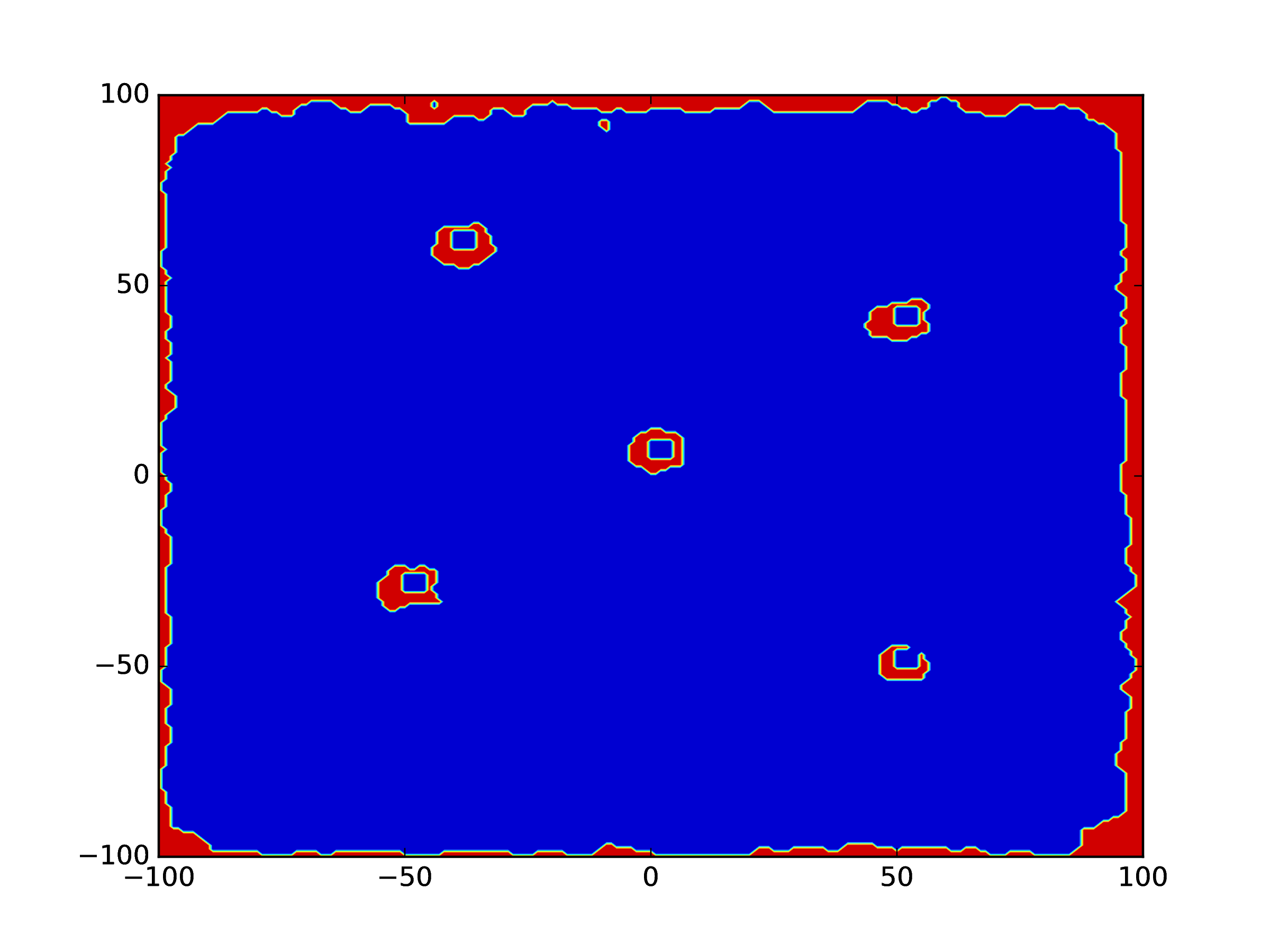}}

		\end{tabular}
		\caption{Simulation on a domain  with PAO=0.2\%.}  
		\label{fig:validation on domain with small figures}
	\end{figure}
    
	\begin{figure}[!th]
		
        \vspace{-3mm}
        
        \centering

				\includegraphics[width=.9\linewidth, height=.4\linewidth]{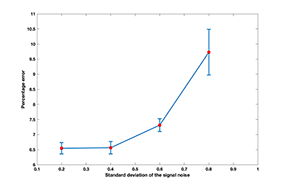}

		\caption{Effect of noise in the received signal on the map estimation error.}  
		\label{fig:validation on domain sigma}
	\end{figure}

	In this section, we validate the mapping procedure in \autoref{sec: procedure} by constructing metric maps of six simulated domains, each with two signal transmitters. Swarms of point robots in each domain were simulated in Python, and all other computations were performed in MATLAB. The robots have a sensing radius of 0.06 m and an average speed of $v = 0.2$ m/s. The simulations were initialized by placing the robots at random points near one of the domain boundaries. The robots follow the motion model \autoref{eqn:State_space_motion_model}, in which $\mathbf{W}(t)$ is a diagonal matrix with 0.1 on the diagonal. The robots employ a simple collision avoidance policy, for which $p_{th}$ = 0.2, by choosing a new random direction upon encountering the domain boundary, an obstacle, or another robot.
	
	For the $2$ m $\times$ $2$ m domains shown in \autoref{fig:worlds}, swarms of $N = 50$ robots were simulated over a deployment time of $T = 300$ s. The outputs at various stages of the mapping procedure for these domains are displayed in \autoref{fig:pdf computed}-\autoref{fig:error maps}: the contour plots of the computed $p^f_i$ (\autoref{fig:pdf computed}), the filtered $p^f_i$ (\autoref{fig:pdf Filtered}), the barcode diagrams (\autoref{fig:barcode}), 
	the thresholded maps (\autoref{fig:maps}), and the absolute error in the maps (\autoref{fig:error maps}). The caption below each sub-figure in \autoref{fig:worlds}-\autoref{fig:error maps} indicates the  percentage of the domain area that is covered by obstacles (PAO). These results show that the procedure generates an accurate metric map of each domain. To further evaluate the performance of our procedure, we ran 20 simulations on each domain in \autoref{fig:worlds} with the same parameters. 
	\autoref{fig:validation on domain} shows the average  threshold value $\gamma$ with its $95\%$ confidence interval and the mean absolute error (MAE) of the map estimation error with the corresponding $95\%$ confidence interval for each domain. We used MAE rather than root-mean-squared error, since each error contributes proportionally in MAE. The plots confirm the effectiveness of our approach, since the average MAE lies between 5\% to 8\% and the error bars are relatively small. 
	We also conducted simulations in a larger, more complex domain of size $20$ m $\times$  20 m, in which $N = 200$ robots were deployed for $T = 1200$ s. 
	These results are presented in \autoref{fig:complex results} and show that the procedure still generates an accurate map. 
	
	A topological map for this domain was also constructed using the technique described in our previous paper \cite{Ramachandran_Wilson_Berman_RAL_17} and compared with the one presented in that paper. From \autoref{fig:complex topological results}, we see that our current approach results in an improved  topological map. 
	
	Finally, in order to examine the effectiveness of our strategy on domains with small obstacles, we performed a simulation on a domain of size $2$ m $\times$ $2$ m with five square obstacles, each of size 4 cm $\times$ 4 cm, using $ 50 $ robots with a  deployment time of $ 300$ s. 
	The maximum standard deviation of the normal distribution associated with the robot's position was approximately $2.8$ cm in both the $x$ and $y$ directions. 
	The results of this simulation are shown in \autoref{fig:validation on domain with small figures}. 
    These results show that our technique is able to generate a reasonably accurate map of the domain, even when the uncertainty in the robots' position is comparable to the size of obstacles in the domain.

	We also studied the effect of the number of robots $N$  and the deployment time $T$ on the performance of the procedure. For each of the 5 domains in \autoref{fig:worlds}, we ran $10$ simulations each with $N \in \{20, 30, 40, 50, 60\}$ and $T = 300$ s (\autoref{fig:varying robots}), and $10$ simulations each with $T \in \{800, 1000, 1200, 1400, 1600\}$ s and $N = 40$ (\autoref{fig:varying time}). The legends in subfigures (b) and (d) of \hyperref[fig:varying robots]{Figures~\ref{fig:varying robots}}-\ref{fig:varying time} show the number of obstacles in the domain corresponding to each plot. In addition, the error bars in \hyperref[fig:varying robots]{Figures~\ref{fig:varying robots}}-\ref{fig:validation on domain} represent the $95\%$ confidence interval of the true value. 
	\autoref{fig:Threshold combined robot} and \autoref{fig:Threshold combined time} show that the resulting mean threshold $\gamma$ with its $95\%$ confidence interval, computed from the 50 simulations over all domains for each parameter set, increases with increasing $N$ and $T$, respectively. The corresponding plots of the mean $\gamma$ for each domain, \autoref{fig:Threshold various domain robot} and \ref{fig:Threshold various domain time}, exhibit the same trend. \autoref{fig:Estimation error combined robots} and \autoref{fig:Estimation error combined time} plot the mean  MAE of the map estimation error with its $95\%$ confidence interval versus $N$ and $T$, respectively, from the 50 simulations for each parameter set. The mean map error does not vary significantly with $N$, possibly because the deployment time $T = 300$ s is sufficient to thoroughly cover the domains, and it decreases with increasing $T$, as would be expected since more localization data is gathered during the deployment.
	To test this hypothesis, we reran the 50 simulations with 5  swarm sizes and $T = 160$ s (\autoref{fig:varying robots low}). \autoref{fig:Estimation error combined robots low} indeed shows that for this low $T$, the MAE of the map estimation error decreases as $N$ increases. The mean map error for each domain versus $N$ and $T$ are shown in \autoref{fig: error various domains robot} and \autoref{fig: error various domains time}, respectively.  
	Finally, \autoref{fig:success robots} and \autoref{fig:success time} show the dependence on $N$ and $T$ of the percentage of the 50 simulations for each parameter set in which the topological technique described in \autoref{subsec:Thresholding the Density Function to Generate the Map} successfully identifies the number of obstacles (topological features) in the five domains. As expected, the success rate increases with increasing $N$ and $T$.
    
   In addition, we investigated the effect of noise in the received signal 
    on the map estimation error. For each of the 5 domains in \autoref{fig:worlds}, we ran $50$ simulations each with the standard deviation of this noise set to $\{0.2,0.4,0.6, 0.8\}$ m and $N = 40$, $T = 300$ s. For each trial, $\mathbf{N}_S(t)$ in \autoref{eqn:output_eqn} was set to a diagonal matrix with 0.1 along the diagonal. \autoref{fig:validation on domain sigma} demonstrates that our method produces a fairly accurate map even when the noise in the received signal exceeds what is anticipated from the measurement model of the robots.  

	\section{CONCLUSIONS}
	\label{sec:conclusion}
	We have proposed a novel technique for automatically generating the metric map of an unknown environment as an occupancy grid map using data collected by a robotic swarm without global localization or inter-robot communication. This data combines the robots' odometry information with their noisy measurements of signals from transmitters outside the domain. 
	The approach was validated through simulations on domains with different numbers of obstacles and robots and different deployment times. We plan to validate our method with robot experiments  
	and investigate the effect of distance-varying sensor noise models on the map accuracy.
	
	



\end{document}